\pdfoutput=1
\documentclass[journal]{IEEEtran}
\usepackage[table]{xcolor}
\usepackage{times}  
\usepackage{helvet}  
\usepackage{courier}  
\usepackage{url}  
\usepackage{graphicx}  
\usepackage{bm}
\usepackage{amsfonts}
\usepackage{amssymb}
\usepackage{makecell}
\usepackage{amsmath}
\usepackage[bookmarks=false]{hyperref}   
\usepackage[noabbrev]{cleveref}
\usepackage[algoruled]{algorithm2e}
\usepackage{subfigure}
\usepackage{booktabs}
\usepackage{pifont}
\usepackage{multirow}
\usepackage{mathtools}

\newcommand{\xmark}{\ding{55}}
\usepackage{amsmath}

\newcommand{\tb}{\textbf}

\newcommand{\bx}{\mathbf{x}}
\newcommand{\bX}{\mathbf{X}}
\newcommand{\bz}{\mathbf{z}}

\newcommand{\bt}[1][]{\theta_{#1}}
\newcommand{\bphi}{\phi}

\newcommand{\beps}{\bm{\epsilon}}
\newcommand{\bmu}{\bm{\mu}}
\newcommand{\bsigma}{\bm{\sigma}}

\newcommand{\bh}{\mathbf{h}}

\newcommand{\by}{\mathbf{y}}
\newcommand{\br}{\mathbf{r}}
\newcommand{\bs}{\mathbf{s}}

\newcommand{\bW}{\mathbf{W}}
\newcommand{\bU}{\mathbf{U}}
\newcommand{\bb}{\mathbf{b}}

\newcommand{\pT}[1][]{p_{\bt[#1]}}
\newcommand{\qPhi}{q_{\bphi}}
\newcommand{\gPhi}{g_{\bphi}}
\newcommand{\fT}{f_{\bt}}

\newcommand{\bxhat}{\hat{\bx}}

\newcommand{\bbR}{\mathbb{R}}
\newcommand{\mcN}{\mathcal{N}}
\newcommand{\mcL}{\mathcal{L}}
\newcommand{\mcX}{\mathcal{X}}

\begin{document}
\title{Anomaly Detection of Time Series with Smoothness-Inducing Sequential Variational Auto-Encoder}
\author{Longyuan~Li,
        Junchi~Yan,~\IEEEmembership{Member,~IEEE,},
        Haiyang Wang,
        and~Yaohui~Jin~\IEEEmembership{Member,~IEEE,}
\thanks{L.Li, H.Wang and Y. Jin are with State Key Lab of Advanced Optical Communication System and Network, and MoE Key Lab of Artificial Intelligence, AI
Institute, Shanghai Jiao Tong University, Shanghai, 200240, P.R. China.

J. Yan is with Department of Computer Science and Engineering, and MoE Key Lab of Artificial Intelligence, AI Institute, Shanghai Jiao Tong University, Shanghai, 200240, P.R. China.

Email: \{jeffli, yanjunchi, 0103050180, jinyh\}@sjtu.edu.cn

Junchi Yan and Yaohui Jin are the corresponding authors. \protect}
}

\markboth{Journal of \LaTeX\ Class Files,~Vol.~14, No.~8, August~2018}%
{Shell \MakeLowercase{\textit{et al.}}: Bare Demo of IEEEtran.cls for IEEE Journals}

\maketitle
\begin{abstract}
Deep generative models have demonstrated their effectiveness in learning latent representation and modeling complex dependencies of time series. In this paper, we present a Smoothness-Inducing Sequential Variational Auto-Encoder (SISVAE) model for robust estimation and anomaly detection of multi-dimensional time series. Our model is based on Variational Auto-Encoder (VAE), and its backbone is fulfilled by a Recurrent Neural Network to capture latent temporal structures of time series for both generative model and inference model. Specifically, our model parameterizes mean and variance for each time-stamp with flexible neural networks, resulting in a non-stationary model that can work without the assumption of constant noise as commonly made by existing Markov models. However, such a flexibility may cause the model fragile to anomalies. To achieve robust density estimation which can also benefit detection tasks, we propose a smoothness-inducing prior over possible estimations. The proposed prior works as a regularizer that places penalty at non-smooth reconstructions. Our model is learned efficiently with a novel stochastic gradient variational Bayes estimator. In particular, we study two decision criteria for anomaly detection: reconstruction probability and reconstruction error. We show the effectiveness of our model on both synthetic datasets and public real-world benchmarks.
\end{abstract}
\begin{IEEEkeywords}
Anomaly Detection, Time Series, Deep Generative Model, Variational Auto-Encoder, Recurrent Neural Network
\end{IEEEkeywords}
\IEEEpeerreviewmaketitle

\section{Introduction}\label{sec:intro}
Time series anomaly detection is an important and challenging problem, and it has been studied over decades across wide application domains, including intelligence transportation \cite{li2009temporal}, health care \cite{hauskrecht2013outlier}, KPI monitoring \cite{xu2018unsupervised}, web intrusion detection \cite{jyothsna2011review}, environmental monitoring \cite{hill2010anomaly}, and fault diagnosis \cite{YanIETEL11}, malware detection \cite{kim2018zero} etc. Such a broad application is also reflected in the diversity of formulations and models to time series anomaly detection. In contrast to anomaly detection on static data, one widely accepted idea is that the temporal continuity plays a key role in all formulations, and unusual changes in the time series data are detected and used to model anomalies. For multi-dimensional correlated time series data, the correlations make the problem further complex, which highlights the need for robust and flexible models.

Consider a matrix $\bX\in \mathbb{R}^{M\times T}$ for a time series corpus that consists of $M$ correlated streams with $T$ time steps for each stream. On one hand, there are generally three scenarios of time series anomaly detection depending on the availability of training data and anomaly labels: supervised, semi-supervised and unsupervised \cite{chandola2009anomaly}.  Anomaly detection tasks without \textit{clean}\footnote{We use  `clean' to describe time series data contain only normal patterns} data or anomaly labels also referred to as unsupervised detection. Unsupervised methods have their advantages that no anomaly labels is needed, which is cost-saving, and they can enjoy better flexibility when the anomaly pattern drifts over time. On the other hand, time series anomaly detection can be in general divided into two settings: i) subsequence or whole sequence level anomaly whereby a subsequence $\bx_{m,t_1:t_2}$ is labeled as an anomaly; ii) point level anomaly for which a measurement  $x_{m,t}$ at time $t$ in sequence $m$ is treated as an anomaly. The former scenario is usually more challenging than the latter. For more comprehensive  surveys, readers are referred to~\cite{aggarwal2015outlier,gupta2014outlier,chandola2012anomaly}.

Technique and methodology to address the above two settings are rather different, and we focus on the popular point-wise anomaly detection for multi-dimensional time series in this paper, which has received wide attention in literature~\cite{tsay2000outliers,basu2007automatic,hill2010anomaly}. In some cases, sequence or subsequence level anomaly detection can be generalized from point level model. In addition, many practical applications require point level detection results as abnormal signal can quickly disappear in the time series especially when the sampling rate is low. We leave it for future work for generalizing our model to more macro-level anomaly detection tasks. On the other hand, as a lack of labeled anomalies necessitates the use of unsupervised approaches, we focus on unsupervised detection model for its practical applicability. This also aligns with the rising trend of adopting unsupervised learning algorithms for anomaly detection such as one-class SVM~\cite{erfani2016high}, GMM~\cite{laxhammar2009anomaly}, tDCGAN~\cite{kim2018zero}, VAE~\cite{xu2018unsupervised}, and VRNN~\cite{chung2015recurrent}.

Different from general time series modeling aiming to fit with the data curve, anomaly detection often need to recover `normal' patterns robustly in the presence of anomalies. This is challenging since the model is optimized to fit all data points. Previous works \cite{locatello2019challenging,zhao2018bias} show that inductive bias is crucial for generalization when learning deep generative models. Motivated by previous works, we seek to take a Bayesian approach. Specifically, we devise a temporal smoothness prior which can be coherently incorporated in the Sequential Variational Auto-Encoder learning procedure, under the well-established probabilistic framework. As such, the learning of latent space can be more focused on the `normal' pattern rather than blindly minimizing the reconstruction error between the raw signal and the generated one by the latent variables. Based on the learned generation model, the anomalies can be detected via a certain criterion by comparison of the reconstructed signal and the raw data point.

\begin{figure*}[t!]
    \centering
    \subfigure[multi-dimensional time series with anomalies]
    {\includegraphics[scale=0.23]
    {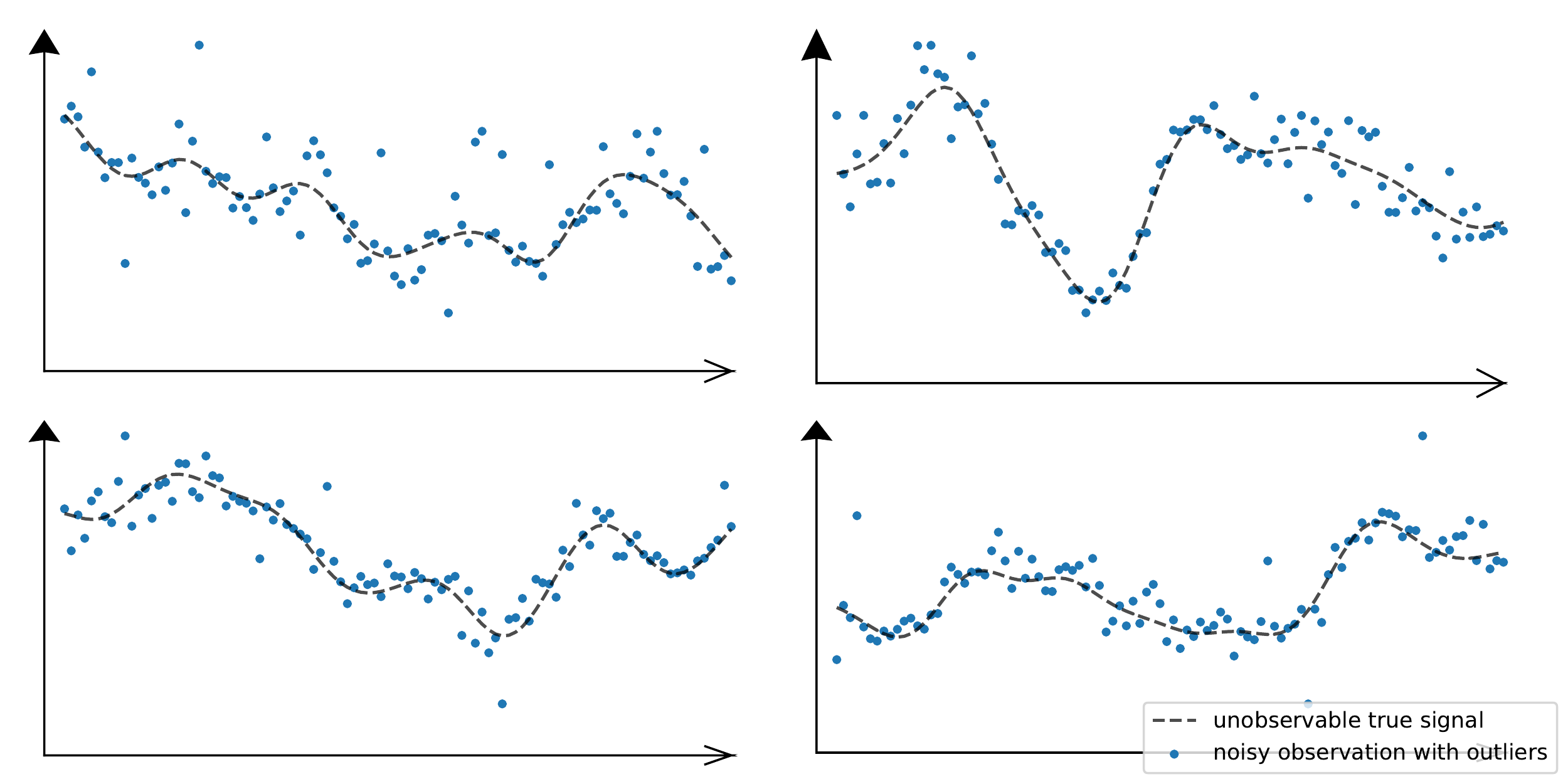}\label{fig:syn_frac}}
    \subfigure[density estimation (shaded area)]
    {\includegraphics[scale=0.23]{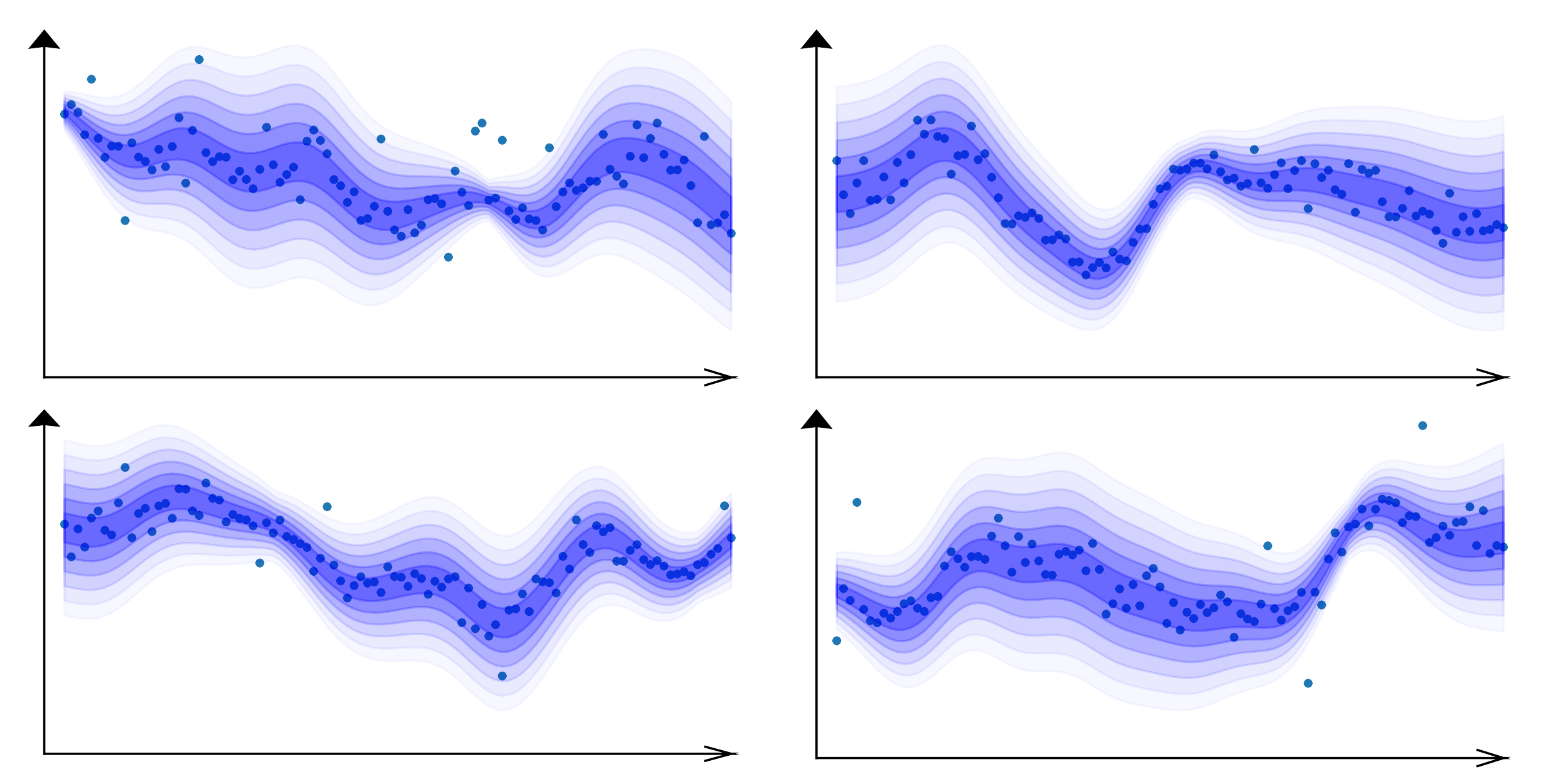}\label{fig:pr_regularization}}
    \subfigure[anomaly detection (with red circle)]
    {\includegraphics[scale=0.23]{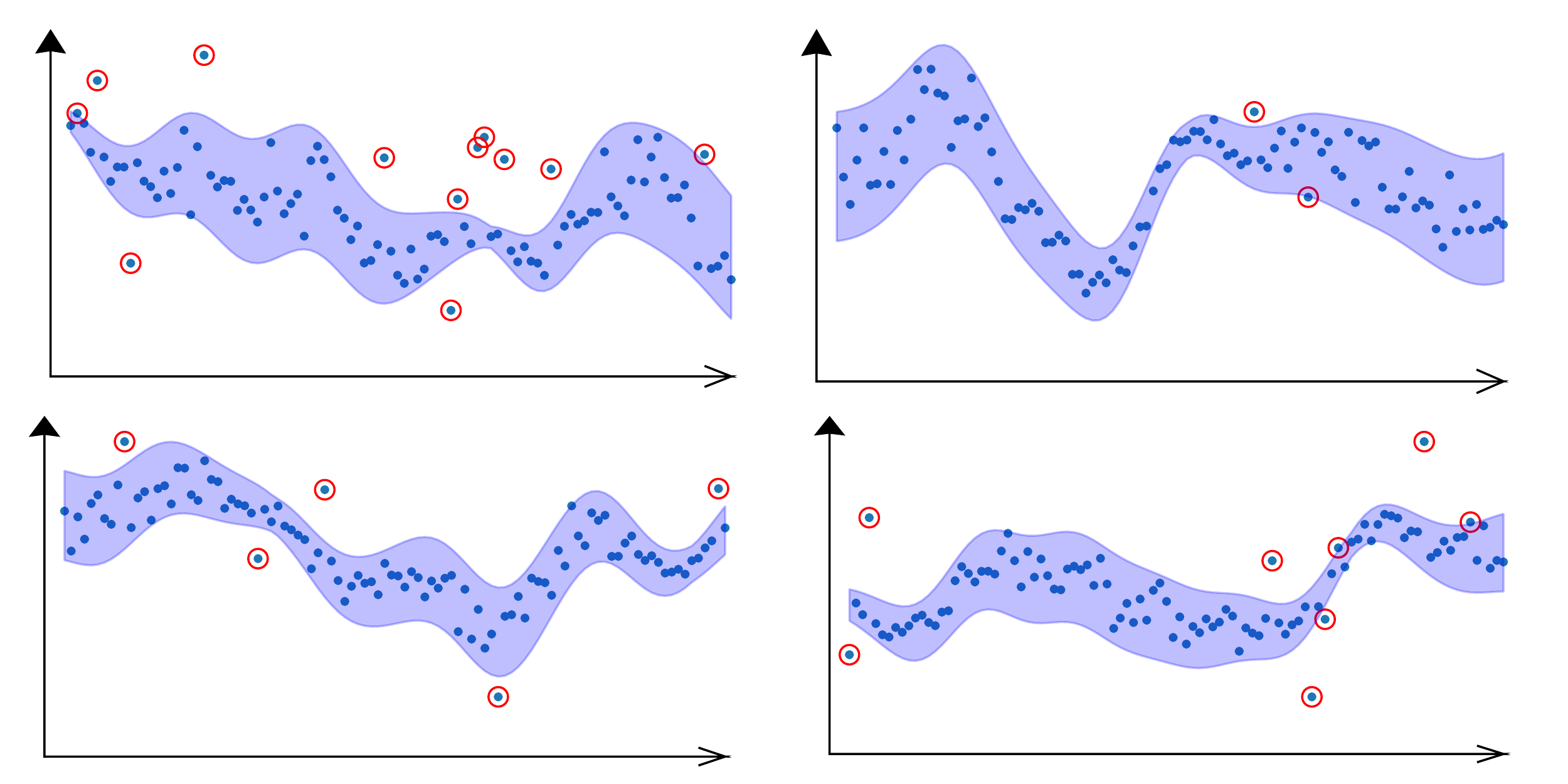}\label{fig:convergence}}
    \label{fig:overall}
    \caption{Procedure illustration of the proposed model: (a) Input multi-dimensional (here 4-dim) time series for anomaly detection. (b) Time-varying probability density function estimation using the proposed model -- shaded region denote variance levels. (c) Anomalies (red circle) detection by thresholding.}
\end{figure*}


\section{Related Work}\label{sec:related}
This paper is devoted to unsupervised point level anomaly detection for multi-dimensional time series. We review mostly related work by roughly dividing them into two groups: deterministic models and probabilistic methods.

\textbf{Deterministic models.} From the optimization perspective, classical approaches often introduce additional regularization term to account for \emph{temporal smoothness}. The extra regularization enables trade-off between \emph{fitness} and \emph{smoothness}, making the model less sensitive to anomalies associated with the training data~\cite{shang2015enhancing,cai2015fast}. Representative works include time series de-trending~\cite{enders2008applied}, and its extension for multi-dimensional correlated time series~\cite{zhou2018non}. These methods tend to learn the general trend of time series. However, they often pay less attention to the stochastic nature of time series, as such the modeling capabilities are degraded when the data are noisy and anomaly-contaminated.

\textbf{Probabilistic models.} In these methods, the core idea is to learn the marginal likelihood $\pT(\bx)$ of the data generation process. Then the anomalies are detected by thresholding reconstruction probability, using the techniques e.g. Hidden Markov Models~\cite{cai2015facets,gornitz2015hidden} and Matrix Factorizations~\cite{li2009dynammo,xiong2011direct}. However, generative models are often computationally restricted to a simple linear model with conjugate probability distributions, due to the intractable integral of marginal likelihood $\pT(\bx) =\int \pT(\bz)\pT(\bx|\bz) d\bz$. Moreover, limited by the potential intractability, the noise is often assumed constant, making them unsuitable for non-stationary time series. 

Recently, the work~\cite{kingma2013auto} proposes the Stochastic Gradient Variational Bayes (SGVB) estimator and Auto-encoded Variational Bayes (AEVB) algorithm. One popular technique under this framework is Variational Auto-Encoder (VAE). By using neural network for parameterizing a flexible inference model $\qPhi(\bz|\bx)$ to approximate posterior $\pT(\bz|\bx)$, along with the reparameterization trick, the lower bound $\mathcal{L}$ of marginal likelihood is differentiable. In consequence, the model can be effectively trained. To a certain extent, the AEVB algorithm liberates the limitations when devising complex probabilistic generative models, especially for deep generative models.

One step further, by taking advantage of the AEVB algorithm, recent studies have introduced deep generative models for anomaly detection. The work \cite{an2015variational} uses Variational Auto-Encoder (VAE) \cite{kingma2013auto} for anomaly detection on non-sequential datasets. However, vanilla VAE makes i.i.d assumption among data points $\bx=\{\bx^{(i)} \}_{i=1}^{N}$, which is unrealistic for real-world time series data. To solve this problem, the very recent approach \cite{xu2018unsupervised} further extends vanilla VAE by preprocessing the raw data with sliding windows, and the windows are assumed to obey i.i.d. Although this approach mitigates the artifact, while the temporal structure is still not well modeled. Alternatively, researchers seek to use a deterministic Recurrent Neural Network (RNN)~\cite{Elman1990RNN} to produce hidden states, then the latent variable is conditioned on the RNN hidden states and the previous latent variables, such that the temporal structure is captured. The resulting models include VRNN~\cite{chung2015recurrent}, STORN~\cite{solch2016variational}, which can be viewed as sequential versions of VAE. However, these models essentially hardly consider the presence of anomalies in their objective functions.

In this paper, we propose a Smoothness-Inducing Sequential Variational Auto-Encoder (SISVAE) model for learning multi-dimensional correlated time series data. In particular, we are focused on point level anomaly detection which can be naturally handled by our model. 

The technical highlights of the paper are as follows:

i) We devise a Bayesian technique to incorporate the smoothness prior to the learning of deep generative model for multi-dimensional time series anomaly detection. Such a design to our best knowledge, has not been studied in literature and it inherits merits for the efficiency of the classical optimization model and the uncertainty modeling capability by the deep generative models. Specifically, the proposed variational smoothness regularizer encourages both smooth mean and variance transitions over time. The resulting objective, with the network can be learned by standard backpropagation.

ii) Different from Markov models that depend on constant noise assumption, our model parameterizes mean and variance individually for each time-stamp using neural networks. This enables the dynamic anomaly detection thresholds adaption according to estimated data noise. This feature is important as the variance of time series may vary over time.

iii) We perform extensive empirical evaluations on both  synthetic datasets and several real-world benchmarks, showing that our model consistently outperforms the state-of-the-art competing models. We also study two decision criteria for performing anomaly detection tasks on a trained SISVAE model: reconstruction probability and reconstruction error.
\begin{table*}[h!]
  \begin{center}
  \caption{Glossary for the terms used in the paper and their interpretation.}
    \label{tab:table1}
    \begin{tabular}{|c|c|l|} 
    \hline
      \textbf{Terms} &\textbf{Symbol} &\textbf{Meaning}\\
      \hline
      latent variable model & $\pT(\bx,\bz)=\pT(\bx|\bz)p(\bz)$ & \makecell[l]{Description of the data generating process involving unobserved random variable $\bz$. }\\ 
      \hline
      \makecell[c]{AEVB algorithm} & $\min_{\bphi} D_{KL}\big[ \qPhi(\bz|\bx)|\pT(\bz|\bx)\big]$ & \makecell[l]{Use a pair of \textit{encoder} and \textit{decoder} to learn a complex deep latent variable model.}\\
      \hline
      generative model & $\pT(\bx|\bz)$ & \makecell[l]{A probabilistic mapping from latent variable $\bz$ to observation $\bx$, also called \textit{decoder}.} \\
      \hline
      inference model & $\qPhi(\bz|\bx)$& \makecell[l]{An approximation to the intractable true posterior $\pT(\bz|\bx)$, also called \textit{encoder}.}\\ 
      \hline
      \makecell[c]{reparameterization \\trick} &\makecell[c]{$\bz=\gPhi(\beps,\bx)$,\\$\beps\sim p(\beps)$} & \makecell[l]{Using a deterministic variable to express random variable $\bz=\gPhi(\beps,\bx)$, where $\beps$ is\\ an auxiliary variable with independent marginal $p(\beps)$.}\\
      \hline
      latent state & $\bz_t$ & \makecell[l]{latent variable associate with observation $\bx_t$.}\\
      \hline
      \makecell[c]{latent temporal \\dependencies} &$p(\bz_{1:T})$ & \makecell[l]{Latent variables are not independent but exhibits some kinds of structure, such as 1st-order\\ Markov chain.}\\
      \hline
      trending prior & $p(\bz_t)$ & \makecell[l]{A time-varying prior distribution for latent state.}\\
      \hline
      non-linear transitions &$p(\bz_t|\bz_{<t})$ &\makecell[l]{Non-linear relationship between latent space and past latent states. It is linear for Kalman \\filter and Hidden Markov Models.}\\
      \hline
      \makecell[c]{non-linear\\ emission probability} & $p(\bx_t|\bz_t)$& \makecell[l]{Nonlinear relationship between latent space and observation. It is linear for Kalman filter \\and Hidden Markov Models.}\\
      \hline
      smoothness prior & $p(\bm{f})$ & \makecell[l]{A generic contextual constraint on the true signal based on human knowledge\\ is the smoothness.}\\
      \hline
      non-stationary noise & $\beps_t$ & \makecell[l]{The noise model varies over time, in many currently used models, only the mean of\\ Gaussian is modeled with dynamics, while the variance is often set to be constant}\\
      \hline
      anomaly score & $\bm{A}$ & \makecell[l]{Score assigned by the model for each observation, higher score indicates higher probability\\ of anomaly.}\\
      \hline
    \end{tabular}
  \end{center}
\end{table*}
\section{Preliminaries}
\label{sec:prem}
Before devotion to the technical details of our proposed model, some preliminaries are described as background to facilitate the presentation of our main method.
\subsection{Problem Formulation}\label{sec:problem_form}
Due to different reasons such as faulty sensors or extreme events, the real-world time series data are often contaminated with anomalies. Consider we have time series data $\bx_{1:T} = (\bx_1, \bx_t,\dots,\bx_t)$ contaminated with anomalies, and each $\bx_t\in \bbR^{M}$ (indicating $T$-dimensional time series) is modeled by:
\begin{equation}\label{eq:problem_form}
    \bx_t = \bm{f}(t) + \beps_t + \bm{I}_{t}\odot\bm{e}_t
\end{equation}
where $\bm{f}(t)$ is the true signal, $\beps_t$ is non-stationary noise of observations, $\bm{I}$ is a $M\times T$ binary matrix, which is the indicator matrix of anomalies. The value $\bm{I}_{m,t} = 1$ when there is an anomaly at observation $\bx_{m,t}$, and $\bm{e}_t$ is the anomaly vector, and $\odot$ is element-wise multiplication. Our goal is to recover the true signal $\bm{f}(t)$ and detect anomaly $\bm{I}$ from $\bx_{1:T}$.


\subsection{Smoothness Prior Modeling}\label{subsec:smooth}
For the unsupervised learning problem, to recover true signal $\bm{f}(t)$ from anomaly-contaminated data, it is important to impose inductive bias or prior knowledge to the model as suggested in \cite{locatello2019challenging,zhao2018bias}. For time series modeling, we first introduce smoothness prior modeling of time series as described in \cite{kitagawa2012smoothness}. Consider the time series $y_{1:T}$ (for convenience, we deal with the uni-variate case), it is assumed to consist of the sum of a function $f$ and observation noise $\epsilon$:
\begin{equation}
    y_t = f(t)+\epsilon_t
\end{equation}
where $f\in\mathcal{F}$ is an unknown smooth function, and $\epsilon\sim\mcN(0,\sigma^2 )$. The problem is to estimate $f$ from the observations. A common approach is to use a parametric model such as polynomial regression or neural network to approximate $f$. The quality of estimation is dependent upon the appropriateness of the assumed model class. The idea is to balance a trade-off of goodness-of-fit to the data, and a smoothness criterion. The solution is to minimize the objective with an appropriately chosen smoothness trade-off parameter $\lambda$:
\begin{equation}\label{eq:smooth}
    \min_{f}\sum\left\{ (y_t - f(t))^2 + \lambda (\nabla^k f(t)))^2 \right\}
\end{equation}
where $\nabla^k$ is a $k$-th order differential constraint on the solution $f$, with $\nabla f(t) = f_t - f_{t-1}$, $\nabla^2 = \nabla(\nabla (f(t)))$, etc.

The solution can be solved from a Bayesian perspective: consider a homoscedastic (variance is constant over time) time series $y_t\sim \mcN \mathcal(f(t),\sigma^2)$ with constant variance $\sigma^2$. The solution becomes maximizing:
\begin{small}
\begin{equation}
    \mcL(f) = \exp \left(-\frac{1}{2\sigma^2}\sum_{t=1}^{T}(y_t-f(t))^2 \right)\exp \left(
    -\frac{\lambda}{2\sigma^2}\sum_{t=1}^{T}(\nabla^k f(t))^2 \right)
\end{equation}
\end{small}

Its Bayesian interpretation can be written by:
\begin{equation}
    \pi(f|y,\lambda,\sigma^2,k) \propto p(y|\sigma^2,f)\pi(f|\lambda,\sigma^2,k)
\end{equation}
where $\pi(f|\lambda,\sigma^2,k)$ is the prior distribution of $f$, and $p(y|\sigma^2,f)$ is data distribution conditioned on $\sigma^2$ and $\pi(f|y,\lambda,\sigma^2,k)$ is the posterior of $f$.
\subsection{Deep Generative Models and SGVB Estimator}
Consider a dataset $\bx=\{\bx^{(i)} \}_{i=1}^{N}$ consisting of $N$ i.i.d data points $\bx^{(i)}$. From a latent variable modeling perspective, we assume the data are generated by some random processes that capture the variations in the observed variables $\bx$. Thus we have the integral of marginal likelihood:
\begin{equation}
\pT(\bx) = \int \pT(\bz) \pT(\bx|\bz) d\bz
\end{equation}
where $\bt$ is generative model parameters, and $\pT(\bz)$ is the prior distribution over latent variable $\bz$, and $\pT(\bx|\bz)$ is the conditional distribution that generates data $\bx$. Variational Auto-Encoder (VAE) \cite{kingma2013auto} models the conditional probability $\pT(\bx|\bz)$ with a neural network as a flexible approximator. However, introducing a complex non-linear mapping for the generation process from $\bz$ to $\bx$ results in intractable inference of posterior $\pT(\bz|\bx)$. Hence VAE uses a variational approximation $\qPhi(\bz|\bx)$ of the posterior, i.e. the evidence lower bound (ELBO) of the marginal likelihood as written by:
\begin{small}
\begin{equation}\label{eq:vae_elbo}
\mathcal{L}(\bt,\bphi;\bx)=-D_{KL}\left(\qPhi(\bz|\bx)\rVert \pT(\bz)\right)+\mathbb{E}_{\qPhi(\bz|\bx)}[\log \pT(\bx|\bz)]
\end{equation}
\end{small}
where the first R.H.S. term is the Kullback-Leibler divergence between two probability distributions $Q$ and $P$, and $\bphi$ denotes variational parameters. The generative model and inference model are trained jointly by maximizing the ELBO w.r.t their parameters. The KL-divergence $D_{KL}(\qPhi(\bz|\bx)\rVert\pT(\bz))$ of Eq.~\ref{eq:vae_elbo} can often be integrated analytically. However, the gradient of the second term is problematic because latent variable $\bz \sim \qPhi(\bz|\bx)$ is  stochastic which blocks backpropagation. The Stochastic Gradient Variational Bayes (SGVB)~\cite{kingma2014stochastic} expresses the random variable $\bz$ as a deterministic variable $\bz=\gPhi(\beps,\bx)$, where $\beps$ is an auxiliary variable with independent marginal distribution $p(\beps)$. Then we have the SGVB estimator $\tilde{\mathcal{L}}(\bt,\bphi;\bx)\simeq \mathcal{L}(\bt,\bphi;\bx)$ of the lower bound:
\begin{align}\label{eq:sgvb2}
\begin{split}
\tilde{\mathcal{L}}(\bt,\bphi;\bx)
= -D_{KL}\left(\qPhi(\bz|\bx)\rVert \pT)\right)
+ \log \pT(\bx|\bz)\\
\text{where \quad} \bz=\gPhi(\beps,\bx)
\text{\quad and \quad} \beps\sim p(\beps)
\end{split}
\end{align}

Learning is performed by optimizing the SVGB estimator with backpropagation. VAE~\cite{kingma2013auto} uses centered isotropic multivariate Gaussian $\pT(\bz)=\mathcal{N}(\bz;\bm{0},\bm{I})$ as prior of latent variable $\bz$. The variational approximate posterior $\qPhi(\bz|\bx)$ is a multivariate Gaussian with diagonal covariance matrix: $\mathcal{N}(\bmu,\text{diag}(\bsigma^2))$, and its mean $\bmu$ and variance $\bsigma^2$ are parameterized by a neural network. For $\pT(\bx|\bz)$, a Bernoulli distribution for binary data or multivariate Gaussian distribution (as adopted in our model) for continuous data is often used.

\section{The Proposed Model}
We present the proposed Smoothness-Inducing Sequential Variational Auto-Encoder (SISVAE) model for multi-dimensional time series point anomaly detection. We first describe how the Sequential VAE can be employed for time series density estimation, which involves a generative model and inference model to learn the latent variable for data points. Then we devise a smooth prior under the Bayesian framework which leads to a smoothness-induced sequential VAE model. We show the model can be used to detect the anomaly either by absolute reconstruction loss or reconstruction probability.
\label{sec:proposed}
\subsection{Sequential VAE as Density Estimator of Time Series}
\begin{figure*}
    \centering
    \subfigure[inference model $\qPhi(\bz|\bx)$]
    {\includegraphics[scale=0.6]{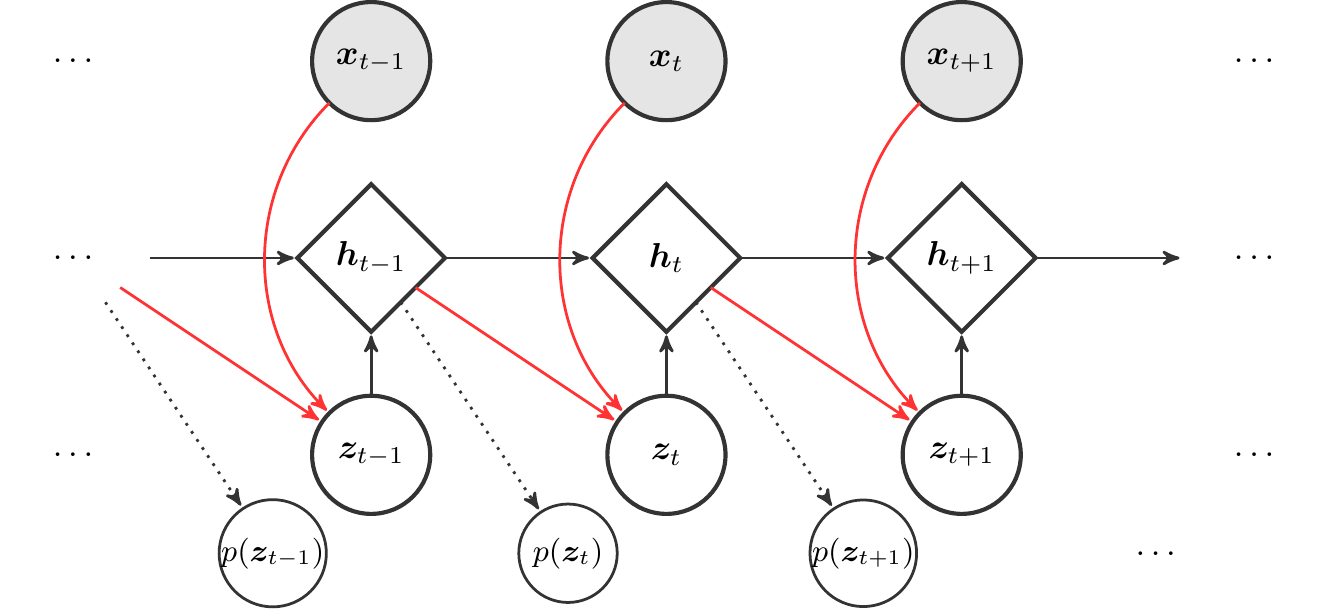}}
    \subfigure[generative model $\pT(\bx|\bz)$]
    {\includegraphics[scale=0.6]{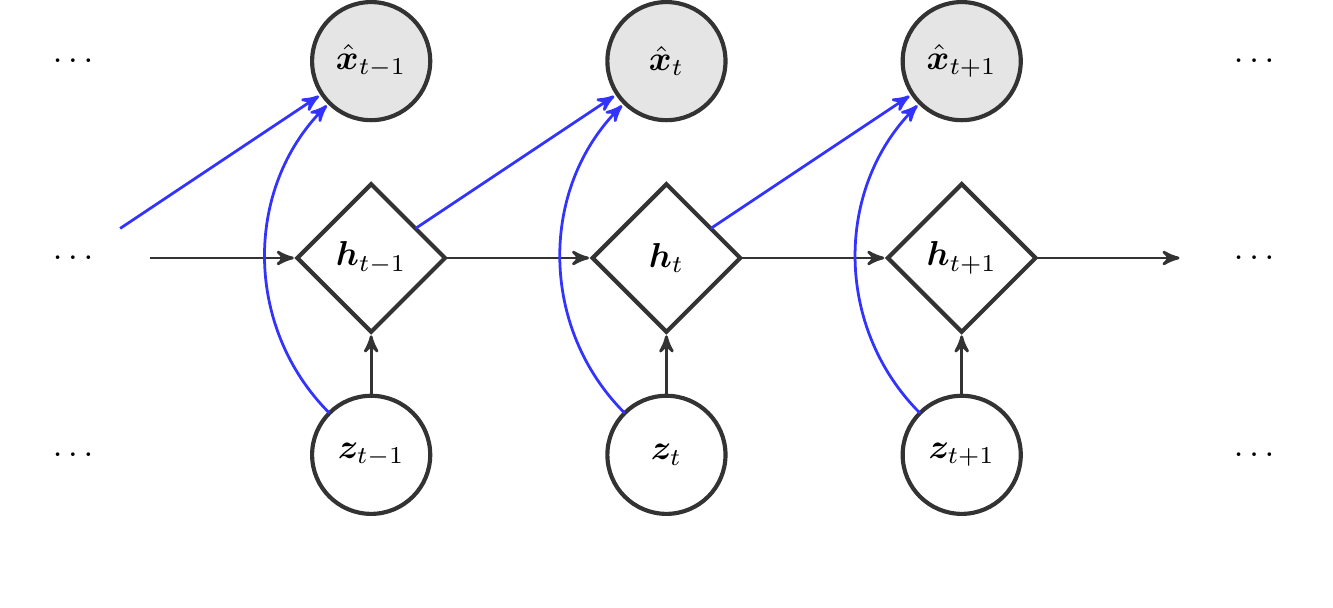}}
    
    \caption{Graphical illustration of the proposed SISVAE model. Circle nodes are random variables, and diamond node are deterministic variables. Gray nodes are observable, and white nodes are unobservable. Black solid lines are the backbone recurrent neural network, as shown in Eq. \ref{eq:gru}. a) Network structure of inference model $\qPhi(\bz|\bx)$, red solid lines are the inference step as shown in Eq. \ref{eq:approx_posterior}, and black dotted lines are the transition prior as shown in Eq.\ref{eq:prior}. b) Network structure of generative model $\pT(\bx|\bz)$, blue solid lines are generation step in Eq. \ref{eq:likeli}.}
    \label{fig:my_label}
\end{figure*}
For time series modeling, many prior publications have extended VAE in a sequential manner~\cite{fabius2014variational,bayer2014learning,chung2015recurrent,fraccaro2016sequential}. The main idea is to use a deterministic Recurrent Neural Network (RNN) as a backbone, represented by the sequence of RNN hidden states. At each timestamp, the RNN hidden state is conditioned on the previous observation $\bx_{t-1}$ and stochastic latent variable $\bz_{t-1}$, such that the model is able to capture sequential features from time series data.

Let $\bx_{1:T} = (\bx_1, \bx_t,\dots,\bx_t)$ denote a multidimensional time series, such as observation data collected by multiple sensors of T time slots. Our goal is to learn the data distribution of the training time series $p(\bx_{1:T})$. The model is composed of two parts, the generative model and the inference model.

\subsubsection{Generative model}
We specify the generative model $\pT(\bx_{1:T}|\bz_{1:T})$ of the following factorization:
\begin{equation}\label{eq:joint}
    \pT(\bx_{\le T},\bz_{\le T}) =
    \prod_{t=1}^{T}
    \pT(\bx_t|\bz_{\le t}, \bx_{<t})
    \pT(\bz_t|\bx_{<t},\bz_{<t})
\end{equation}

The first factor of R.H.S of Eq.~\ref{eq:joint} $\pT(\bx_t|\bz_{\le t}, \bx_{<t})$ is the conditional likelihood of a data point given all previous data and latent states. We notate $\bxhat_t$ as the random variable for the reconstructed data distribution of observation $\bx_t$, we have:
\begin{equation}\label{eq:likeli}
\begin{split}
    p(\bxhat|\bz_t)\sim\mathcal{N}\left(\bmu_{x,t},\text{diag}(\bsigma_{x,t}^2)\right),\\
    \text{where }[\bmu_{x,t},\bsigma_{x,t}]=
    \varphi_{\bt}^{\text{dec}}\left(\varphi_{\bt}^{\bz}(\bz_t),\bh_{t-1}\right)
\end{split}
\end{equation}
where $\bmu_{x,t}$ and $\bsigma_{x,t}$ denote the sufficient statistic parameters of the generated data distribution, and $\bt$ is the parameter set of the mapping function $\varphi^{\text{dec}}$. In this multivariate Gaussian case, they denote mean and variance respectively. The hidden states are updated with the recurrence equation using gated recurrent unit. The second factor $\pT(\bz_t|\bx_{<t},\bz_{<t})$ is the prior of latent state, depending on previous RNN hidden state $\bh_{t-1}$:
\begin{equation}\label{eq:prior}
\begin{split}
    \bz_t\sim\mathcal{N}\left(\bmu_{0,t},\text{diag}(\bsigma_{0,t}^2)\right),\\
    \quad\text{where }[\bmu_{0,t},\bsigma_{0,t}]=\varphi_{\bt}^{\text{prior}}(\bh_{t-1})
\end{split}
\end{equation}
where $\bmu_{0,t}$ and $\bsigma_{0,t}$ are sufficient statistic parameters of the prior. By introducing temporal structure into the prior distribution of the latent variable, the representation power is enhanced. The hidden states are updated with the recurrence equation using gated recurrent unit (GRU)~\cite{cho2014learning}:
\begin{equation}\label{eq:gru}
    \bh_t=\fT\left(\varphi_{\bt}^{\bz}(\bz_t),\bh_{t-1}\right) 
\end{equation}
The GRU transition equations of our recurrence model are:
\begin{equation}\label{eq:transition}
    \begin{split}
        \by_t &= [\varphi_{\bt}^{\bz}(\bz_t)]\\
        \br_t &= \sigma(\bW_r\by_t+\bU_r \bh_{t-1}+\bb_r)\\
        \bs_t &= \sigma(\bW_s\by_t+\bU_s \bh_{t-1}+\bb_r)\\
        \tilde{\bh}_t &= \tanh(\bW_h+\br_t(\bU_h\bh_{t-1})+\bb_h)\\
        \bh_t &= (1-\bs_t)\tilde{\bh_t}+\bs_t \bh_{t-1}
    \end{split}
\end{equation}
where $\bW_*,\bU_*,\bb_*$ are model parameters of GRU in SISVAE model, and $\sigma$ is the sigmoid function. In \cref{eq:prior,eq:likeli,eq:gru}, $\varphi_{\bt}^{\text{prior}}$ and $\varphi_{\bt}^{\text{dec}}$ are realized by networks. $\varphi_{\bt}^{\bx}(\bx_t)$ and $\varphi_{\bt}^{\bz}(\bz_t)$ are feature extractors of observed data $\bx_t$ and latent variable $\bz_t$ respectively, which are also realized by networks.

\subsubsection{Inference model}
We use variational inference to learn an approximate posterior over latent variables conditioned on the given data, which means we need to construct an approximating distribution $\qPhi$ parameterized by $\bphi$. We train the generative model using Auto-Encoding Variable Bayes (AEVB) algorithm~\cite{kingma2013auto}:
\begin{equation}\label{eq:aevb}
\max_{\bt,\bphi}\mathbb{E}_{p(\bx_{1:T})}
\bigg[\mathbb{E}_{\qPhi}\bigg[\log\frac{\pT(\bx_{1:T},\bz_{1:T})}{\qPhi(\bz_{1:T}|\bx_{1:T})}\bigg]\bigg]
\end{equation}

We construct a $\qPhi$ distribution of the following factorization:
\begin{equation}\label{eq:q_factor}
    \qPhi(\bz_{\le T}|\bx_{\le T})=\prod_{t=1}^{T}\qPhi(\bz_t|\bx_{\le t},\bz_{<t})
\end{equation}

The approximate posterior of latent variable $\bz_t$ depends on $\bx_t$ and $\bh_{t-1}$:
\begin{equation}\label{eq:approx_posterior}
\begin{split}
    p(\bz_t|\bx_t)\sim\mathcal{N}\left(\bmu_{z,t},\text{diag}(\bsigma_{z,t}^2)\right),\\
    \text{where }[\bmu_{z,t},\bsigma_{z,t}]=
    \varphi_{\bphi}^{\text{enc}}\left(\varphi_{\bt}^{\bx}(\bx_t),\bh_{t-1}\right)
\end{split}
\end{equation}
where $\bmu_{z,t}$, $\bsigma_{z,t}$ denote the mean and variance of the approximate posterior, and $\varphi_{\bphi}^{\text{enc}}$ is approximated by a neural network.

\subsection{Anomaly Detection with Sequential VAE}

As discussed before, we seek to devise suitable inductive bias for anomaly detection of time series in the sequential VAE framework. To this end, we begin with the factorization of the existing loss function. Specifically, we substitute the numerator and  denominator in Eq.~\ref{eq:aevb} with Eq.~\ref{eq:joint} and Eq.~\ref{eq:q_factor} respectively, the objective of Sequential VAE becomes:
\begin{equation}\label{eq:objective}
\begin{split}
    \tilde{\mathcal{L}}_{\text{SAE}}
    =&\sum_{t=1}^{T}\bigg\{\underbrace{\-D_{KL}(\qPhi(\bz_t|\bx_{\le t},\bz_{<t})\rVert \pT(\bz_t|\bx_{< t},\bz_{<t}))}_{\text{inference loss}} \\
    &+\underbrace{\log \pT(\bx_t|\bz_{\le t},\bx_{<t})}_{\text{reconstruction loss}}\bigg\}
\end{split}
\end{equation}

The first term is the loss of inference model, which can be viewed as a regularizer, encouraging the model to learn disentangled feature by putting a prior over latent variable \cite{higgins2017beta}. The second term is the reconstruction loss of the generative model. Suppose $x_{m,t} = \bm{f}_m(t) + \epsilon_{m,t} + \bm{e}_{m,t}$ is an anomaly, and its associated reconstruction distribution is $\hat{x}_{m,t}=\mcN(\mu_{m,t}, \sigma_{m,t})$. The generative model tries to reconstruct the contaminated $x_{m,t}$, rather than the true signal $\bm{f}_m(t)$, otherwise the reconstruct loss, which is the negative log likelihood $-\pT(x_{m,t}|\hat{x}_{m,t})$ will be high. To minimize the objective loss function from the data with anomalies, the model will either learn a biased mean $\mu_{m,t}$, or an over-estimated variance $\sigma_{m,t}$. As a result, the Sequential VAE can be vulnerable to anomalies.

To address account for such a bias caused by the anomalies, one natural idea is to induce smoothness prior to the generative model. Recall in Section \ref{subsec:smooth}, the smoothness criterion refers to the accumulative difference of signal $\sum_{t=1}^{T}\nabla f_t$. One might be tempted to simply add such a smoothness regularization to the objective. However, since the Sequential VAE is under probabilistic framework, such a forced combination is not mathematically coherent. However, the classical approach treat observation as deterministic variable, however, in sequential VAE, we treat each observation $\bx_t$ as a random variable, and we estimate the probability distribution for each observation $p(\bx_t)\sim\mcN(\bmu_t,\text{diag}(\bsigma)_t^2)$, such that the classical regularizer failed to work in probabilistic framework, which calls for more comprehensive method.

We extend the smoothness regularization to probabilistic framework, by following the common assumption that the probability density function over time would vary smoothly over time. Among all possible reconstructed series $\bxhat_{1:T}\in \mathcal{X}$, the true signal should have a smooth transition of distribution. To construct a valid regularizer, we need to measure the smoothness of a time-varying probability distribution.

For two consecutive data points of a single time series: $x_{m,t-1}$ and $x_{m,t}$, associated with reconstructions $\pT(\hat{x}_{m,t-1})$ and $\pT(\hat{x}_{m,t})$. Let $d(\cdot,\cdot)$ to be a distance metric between two distributions, we propose accumulative transition cost of time-varying probability distributions:
\begin{equation}
    \mcL_{smooth}=\sum_{t=1}^{T}\sum_{m=1}^{M} d\left(\pT(\hat{x}_{m,t-1}),\pT(\hat{x}_{m,t})\right)
\end{equation}

In this paper, we choose KL-divergence as the distance metric between distributions. For Isotropic Gaussian case in this paper, the transition cost of two Normal distributions is:
\begin{equation}
\begin{split}
    &D_{KL}\left(\mcN(\mu_{m,t-1},\sigma_{m,t-1})| \mcN(\mu_{m,t},\sigma_{m,t})\right) \\
    = &\log\frac{\sigma_{m,t}}{\sigma_{m,t-1}} + \frac{\sigma_{m,t-1}^2+(\mu_{m,t-1}-\mu_{m,t})^2}{2\sigma_{m,t}^2} - \frac{1}{2}
\end{split}
\end{equation}
\begin{algorithm}[tb!]
\label{alg:training}
\caption{Auto-Encoded Variational Bayes algorithm for learning Smoothness Inducing Sequential Variational Auto-Encoder for Time Series (SISVAE).}
\KwIn{Time series matrix $\bX\in \mathbb{R}^{M\times T}$}
\KwIn{$\lambda$ for smoothness, $W$ and $s$ for sliding windows}
Split time series into chunks $\mathcal{D}=\{
\mcX_d\}_{1:D}, \mcX\in\bbR^{M\times W}$\\
Randomly initialize $\bt$, $\bphi$\; Initialize $\bh_0=\bm{0}$\;
\While{not converged}{
Sample a mini-batch of $P$ chunks from dataset $\mathcal{D}$;\\
\For{each chunk in parallel}
{ 
\For{$t=1$ to $W$}
{$\bx_t=\mcX_{:,t}$\;
$\beps\sim \mathcal{N}(0,\bm{I})$//random sampling\;
$[\bmu_{z,t},\bsigma_{z,t}]=
    \varphi_{\bphi}^{\text{enc}}(\varphi_{\bt}^{\bx}(\bx_t),\bh_{t-1})$//encoding\;
$\bz_t=\bmu_{z,t}+\bsigma_{z,t}\odot\beps$ // reparam trick\;
$[\bmu_{0,t},\bsigma_{0,t}]=\varphi_{\bt}^{\text{prior}}(\bh_{t-1})$//prior\;
$[\bmu_{x,t},\bsigma_{x,t}]=
    \varphi_{\bt}^{\text{dec}}(\varphi_{\bt}^{\bz}(\bz_t),\bh_{t-1})$//decoding\;
$\bh_t=\fT(\varphi_{\bt}^{\bz}(\bz_t),\bh_{t-1})$//recurrence\;
}}
Compute gradient $\nabla_{\bt}\mathcal{L}$ and $\nabla_{\bphi}\mathcal{L}$ with $\bz_{\le W}$\;
Update parameters $\bt$ and $\bphi$;
}
\Return{$\bt$, $\bphi$}
\end{algorithm}

The final objective involves: 1) encoding loss for inference model; 2) expected reconstruction error for generative model; 3) the proposed variational smoothness reguarlizer. We apply stochastic gradient variational Bayes (SGVB) estimator to approximate the expected reconstruction error $\mathbb{E}_{\qPhi(\bz_{\le T}|\bx_{\le T})}[\log \pT(\bx_t|\bz_{\le t},\bx_{<t})]$. The learning objective is:
\begin{small}
\begin{equation}\label{eq:sisvae_obj}
\begin{split}
    \mcL_{SISVAE}
    =&\underbrace{\sum_{t=1}^{T}\Big\{-D_{KL}(\qPhi(\bz_t|\bx_{\le t},\bz_{<t})\rVert \pT(\bz_t|\bx_{< t},\bz_{<t}))}_{\text{inference loss}}\\
    &+\underbrace{\log \pT(\bx_t|\bz_{\le t},\bx_{<t})}_{\text{reconstruction loss}}\\
    &+\underbrace{\lambda\ D_{KL}(\pT(\bxhat_{t-1}|\bz_{\le t-1},\bx_{<t-1})\rVert \pT(\bxhat_t|\bz_{\le t},\bx_{<t}))}_{\text{smoothness loss}}\Big\}
\end{split}
\end{equation}
\end{small}
where $\bz_t=\bmu_{z,t}+\bsigma_{z,t}\odot\beps$ and $\beps\sim \mathcal{N}(0,\bm{I})$, and $\lambda$ is a smoothness regularization hyper-parameter. We set $\pT(\bxhat_{t}|\bz_{\le t-1},\bx_{<t-1})=\pT(\bxhat_t|\bz_{\le t},\bx_{<t})$ when $t=1$. 
\subsection{Training}

\subsubsection{Preparing dataset}
Ideally, we could train SISVAE with a whole time series as input, and the model outputs the whole reconstruction. However, in many real applications, the time series is typically very long. The objective function Eq. \ref{eq:sisvae_obj} of our model is summed sequentially over time, such that computing the gradient of it requires back propagation over time \cite{werbos1990backpropagation}. Researches \cite{pascanu2013difficulty} observe that RNNs are hard to train when sequence length is over 200, the model may face the problem of vanishing gradients and exploding gradients, which is also undesirable for anomaly detection tasks.

To solve this problem, we divide the long time series into short chunks, using sliding windows technique. We apply a sliding windows to the time series, which slides over multiple time series synchronously. The sliding windows is controlled by two parameters: window size $W$ and step size $s$. Specifically, for each dataset $\bX \in \mathbb{R}^{M\times T}$, we have:
\begin{equation}
\begin{aligned}
    \mcX_1 &\coloneqq \bX_{M,1:1+W}\\
    \mcX_2 &\coloneqq \bX_{M,1+s:1+s+W}\\
    &\quad\vdots\\
    \mcX_d &\coloneqq \bX_{M,1+(d-1)*s:1+(d-1)*s+W}\\
\end{aligned}
\end{equation}
where $\mcX \in \bbR^{M\times W}$, each $\mcX$ is a short time series chunk with length W, putting together, we have:
\begin{equation}
    \mathcal{D} = \{\mcX_{1}, \dots,\mcX_{d}, \dots, \mcX_{D}\}
\end{equation}
where $\mathcal{D}$ is the whole dataset. Each item is a short time series chunk. The learning procedure is depicted in Algorithm \ref{alg:training}.

\begin{figure}[tb!]
    \centering
    \includegraphics[width=0.4\textwidth]{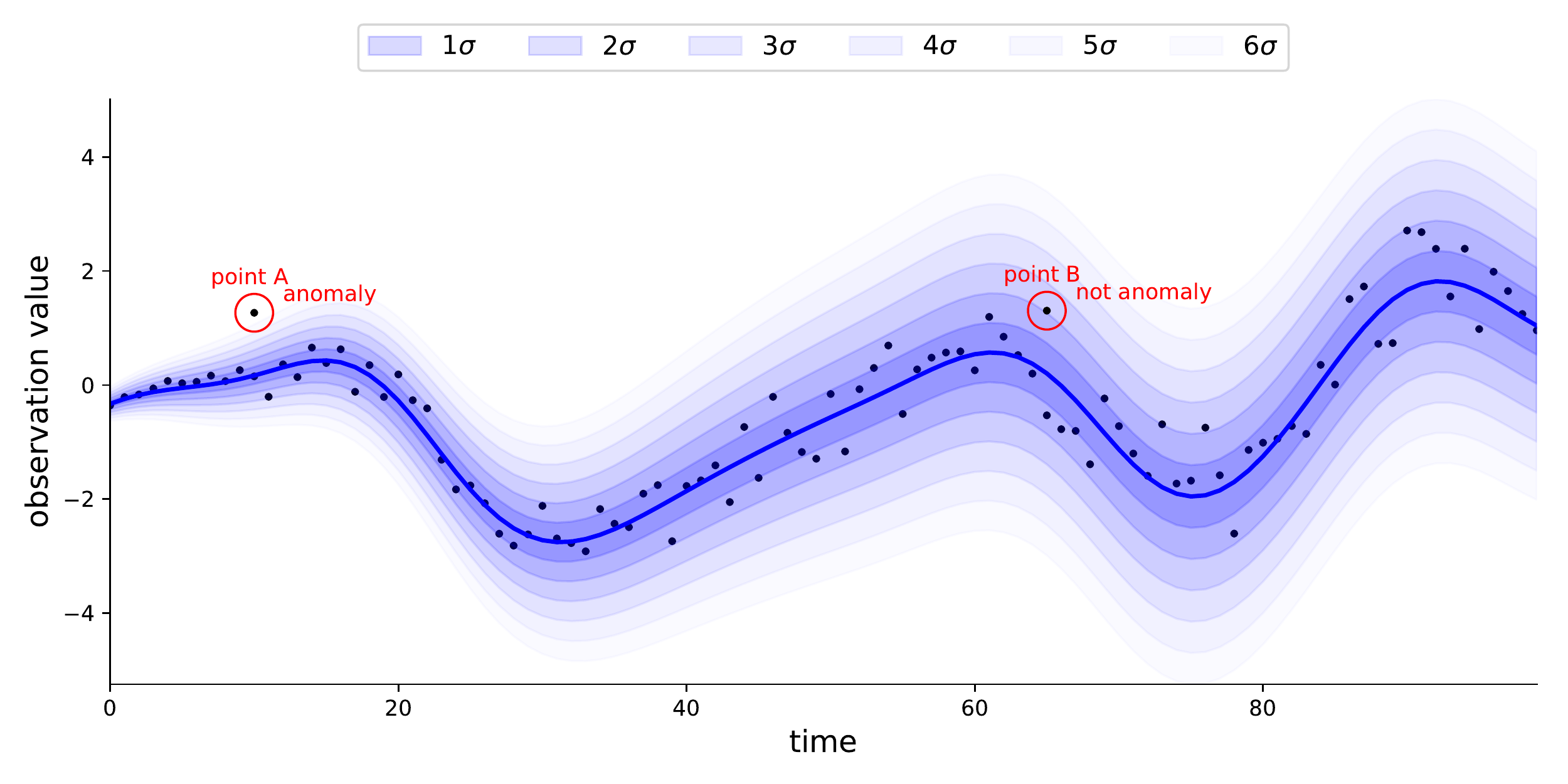}
    \caption{When the variance of data varies over time, absolute reconstruction error may fail to detect some anomalies.}
    \label{fig:score_example}
\end{figure}

\begin{table}[!tb]
\centering
\caption{Summary of different time series anomaly detection models. Note the triangle for Donut denotes the objective robustness is incorporated indirectly by considering the labels of anomalies in the objective while our method fulfills this purpose by using the (smooth) trending prior.}
\resizebox{0.48\textwidth}{!}{
\begin{tabular}{@{}lcc|cccc@{}}
\toprule
& \multicolumn{2}{c|}{Traditional Models} & \multicolumn{3}{c}{Deep Latent Variable Models} \\
        & ARMA         & LDS        & Donut     & STORN     & SISVAE-p     \\ \midrule
latent temporal dependencies          & --            & \checkmark          & \xmark         & \checkmark         & \checkmark                  \\
non-linear emission probability               & --           & \xmark          & \checkmark         & \checkmark         & \checkmark                   \\
non-linear latent transitions            & --            & \xmark          & \xmark         & \checkmark         & \checkmark                   \\
trending prior                   & --            & \checkmark          & \xmark         & \xmark         & \checkmark                   \\
multi-dimensional                 & \xmark            & \checkmark          & \xmark         & \checkmark         & \checkmark                    \\
non-stationary variance               & \xmark            & \xmark          & \checkmark         & \checkmark         & \checkmark                  \\
long sequence            & \xmark            & \xmark          & \checkmark         & \xmark         & \checkmark                  \\
objective robustness             & --            & \xmark          & $\nabla$       & \xmark         & \checkmark                   \\ \bottomrule
\end{tabular}}
 \label{tab:overview}
\end{table}


\subsection{Computing Anomaly Scores}
The next step is to compute anomaly score for each data point $\bx_{m,t}$. For a trained model, input a chunk of time series $\mcX = \{ \bx_{t} \}_{1:W}$, each $\bx$ is a $M$-dimensional vector, the model first encode it into $\bz_{1:W}$, then decode to a sequence of random variables $\hat\mcX=\{ \mcN(\bmu_t, \text{diag}(\bsigma_t))\}_{1:W}$. The whole encoding and decoding process is called \textit{probabilistic reconstruction}.

Recall that in Eq. \ref{eq:problem_form} of Section \ref{sec:problem_form}, we use $\bm{I}$ to indicate the indicator matrix of anomaly labels. In this step, we use $\bm{A}$ to indicate the matrix of anomaly scores.

One natural choice is to use absolute reconstruction error, like anomaly detection with deterministic auto-encoders \cite{zhou2017anomaly}. The anomaly score $e_{m,t}$ for observation $x_{m,t}$ is computed as:
\begin{equation}
    e_{m,t} = \lVert x_{m,t} - \mu_{m,t} \rVert
\end{equation}

where $\mu_{m,t}$ is the $m$-th dimension of reconstructed mean $\bmu_t$. Intuitively, a large reconstruction error indicates anomaly behavior of time series, which is a common assumption for many reconstruction based anomaly detection algorithms. The underlying assumption of absolute reconstruction error is that the variance is stationary, which means that the variance is constant over time. However, in many real-world data, the variance is non-stationary, and it varies over time. We illustrate this using an example as shown in Fig. \ref{fig:score_example}. We generate some time series data with Gaussian process as true signal, denoted as the real blue line centering scatters, then we add some Gaussian noise $\mcN(0, \sigma_t)$, where $\sigma_t$ increases over time. The black dots is the noisy observation, and blue transparent regions increasing level of noise. Then, we add two points, point A and point B, they both deviate from the true signal with distance 1. We see that, point A is more likely to be an anomaly point, while point B is more likely to be a normal point that located at some noisy period. In this case, if we use absolute reconstruction error, we may fail to detect anomaly point A, or otherwise report normal point B as an anomaly.

Due to the non-stationary nature of real-world time series, we need a robust method for computing anomaly scores. Previous VAE-based anomaly detection methods have introduced reconstruction probability based anomaly score, here we devise a reconstruction probability for sequential Auto-encoders. Specifically, the probability density of input $\bx$ is:
\begin{equation}
    \pT(\bx) = \mathbb{E}_{\qPhi(\bz|\bx)}[\pT(\bx|\bz)]
\end{equation}

Draw $L$ samples from $\qPhi(\bz|\bx)$, and the anomaly score is negative of average of $L$ probability. For numerical stability, we often work with $\log$ probability, and we have 
\begin{equation}
    \log \pT(\bx) = \frac{1}{L}\sum_{l=1}^{L}\pT(\bx_t|\bmu_{x}^{(l)},\bsigma_{x}^{(l)})
\end{equation}

However, for Sequential VAE, this can be more challenging. Because $\bz$ is not independent, but has sequential dependencies. We choose to adopt Sequential Monte Carlo (SMC)~\cite{doucet2001introduction}, which is widely used in state-space models, such as Bayesian filters~\cite{doucet2000sequential}, and dynamical systems~\cite{liu1998sequential}.

\begin{algorithm}[tb!]
\label{alg:smc}
\caption{Sequential Monte Carlo (SMC) for computing anomaly score and detecting anomalies.}
\KwIn{Trained model $\bt$, $\bphi$, threshold $\alpha$, time series chunk $\mcX\in \bbR^{M\times W}$, \# of SMC iterations $L$.}
\KwOut{Anomaly score $\bm{A}$, detected anomaly $\bm{I}_\alpha$}
Initialize $\bm{A} = \bm{0}$ ;\
\For{$l=1$ to L}{
\For{$t=1$ to $W$}
{$\bx_t=\mcX_{:,t}$\;
$\beps\sim \mathcal{N}(0,\bm{I})$//random sampling\;
$[\bmu_{z,t},\bsigma_{z,t}]=
    \varphi_{\bphi}^{\text{enc}}(\varphi_{\bt}^{\bx}(\bx_t),\bh_{t-1})$//encoding\;
$\bz_t=\bmu_{z,t}+\bsigma_{z,t}\odot\beps$ // reparam trick\;
$[\bmu_{x,t},\bsigma_{x,t}]=
    \varphi_{\bt}^{\text{dec}}(\varphi_{\bt}^{\bz}(\bz_t),\bh_{t-1})$//decoding\;
$\bh_t=\fT(\varphi_{\bt}^{\bz}(\bz_t),\bh_{t-1})$//recurrence\;
$\bm{A}_{:,t} \mathrel{+}= -\frac{1}{L} \log p(\bx_t|\bmu_{x,t},\bsigma_{x,t})$;
}
}
$\bm{I}_\alpha = \bm{A} > \alpha$; // element-wise comparison\;
\Return{$\bm{A}$, $\bm{I}_\alpha$};
\end{algorithm}

We conclude the SMC for computing anomaly scores and detecting anomalies in Algorithm \ref{alg:smc}. For the full multivariate time series matrix, $\bm{X}\in \bbR^{M\times T}$, we slice them into non-overlapped chunks, and reconstruct each of them individually. 

\section{Experiments}
\label{sec:experiments}
We present the experiment for robust modeling and anomaly detection for multi-dimensional correlated time series, compared with peer VAE methods. Recall that in this paper we deal with point level anomaly detection without supervision.

\subsection{Protocols and Settings}\label{sec:setting}
\textbf{Performance Metrics.} For time series data $\bX\in\bbR^{M\times T}$, the model outputs anomaly score matrix $\bm{A}\in\bbR^{M\times T}$, and each item $A_{m,t}$ is the likelihood of data point $x_{m,t}$ being an anomaly. For detection, we set a threshold $\alpha$ to obtain the final anomalies label indicator matrix $\bm{I}_\alpha \coloneqq \bm{A} > \alpha$. Then we compute precision, recall, true positive rate (TPR), false positive rate (FPR) at time series point level.



Trade-off is needed as the detection characteristics vary by different threshold $\alpha$. When threshold is tight, the precision will be higher and recall will be lower.  We compute area under receiver operating characteristic (AUROC), area under precision recall curve (AUPRC) and the best F1-score. AUROC quantifies the FPR and TPR with different $\alpha$, and AUPRC quantifies the precision and recall  with different $\alpha$. Note that all three metrics are independent of specific threshold.

There are inherent differences between AUROC and AUPRC. For anomaly detection tasks where class (anomaly or normal data points) distributions are heavily imbalances, the AUROC metric is less insensitive to false positives than the AUPRC metric. As discussed in \cite{davis2006relationship,saito2015precision}, the AUPRC score is more discriminative than AUROC score. Previous work may choose one of the metrics to evaluate the performance. To comprehensively evaluate the model's performance, we report both AUROC and AUPRC scores, but we mainly focus on analysis AUPRC in this paper.


\textbf{Peer methods.}
Recent deep learning based methods are compared, including Donut and STORN.
\begin{itemize}
    \item \textbf{Donut~\cite{xu2018unsupervised}.} The model proposed very recently is based on VAE, which takes a fixed length sliding windows of an uni-variate time series as input. To capture `normal patterns' of data, the modified ELBO technique is devised to leverage generative model $\pT(\bx|\bz)$ and latent regularization $\pT(\bz)$ of VAE, such that the model is encouraged to learn only normal patterns.

    \item \textbf{STORN~\cite{solch2016variational}.} As a sequential version of VAE, the model is based on stochastic recurrent neural network (STORN). Different from SISVAE, the sequences of latent variables $\bz_t$ are generated independently. The model uses lower bound output as anomaly score.
    
\end{itemize}

We also include three traditional statistical models.
\begin{itemize}
    \item \textbf{History Average (HA).}  We use absolute deviation from history average value as anomaly score. We include this model in comparison to show how a very simple model perform within different datasets.
    
    \item \textbf{Auto-regressive Moving Average (ARMA)~\cite{galeano2006outlier}.} ARMA is a classic time series analysis model \cite{hamilton1994time}. We use absolute fitting residuals as anomaly score.
    
    \item \textbf{Linear Dynamical System (LDS).} This model is also known as \textit{Kalman filter}. Different from deep latent variable based state space models, the transition function and emission probability are linear (see Table \ref{tab:overview}).
\end{itemize}

We further include some variants of our model, to study the contribution and behavior of each component in our model.

\begin{itemize}

    \item \textbf{SISVAE-p} Our recommended model, which uses the sliding windows of a time series matrix as input, similar to Donut. This model is more applicable for long sequences.
    
    \item \textbf{SISVAE-e.} Different from SISVAE-p, it takes absolute reconstruction error as anomaly score. We use this model to show the performance gap by different detection criteria.
    
    \item \textbf{SISVAE-0} To show the effect of our smoothness regularizer, we let SISVAE-0 to be SISVAE-p for $\lambda=0$ .
    
    \item \textbf{VAE-s.} To test the contribution of the proposed smoothness prior technique, we add the regularization to vanilla VAE as a comparison named VAE-s. The model can be viewed as our SISVAE model without latent temporal structure. We use this model to show the effectiveness of introducing latent temporal structure.
\end{itemize}
\begin{figure*}[t!]
    \centering
    \subfigure[sensitivity to anomaly ratio]
    {\includegraphics[width=0.245\textwidth]
    {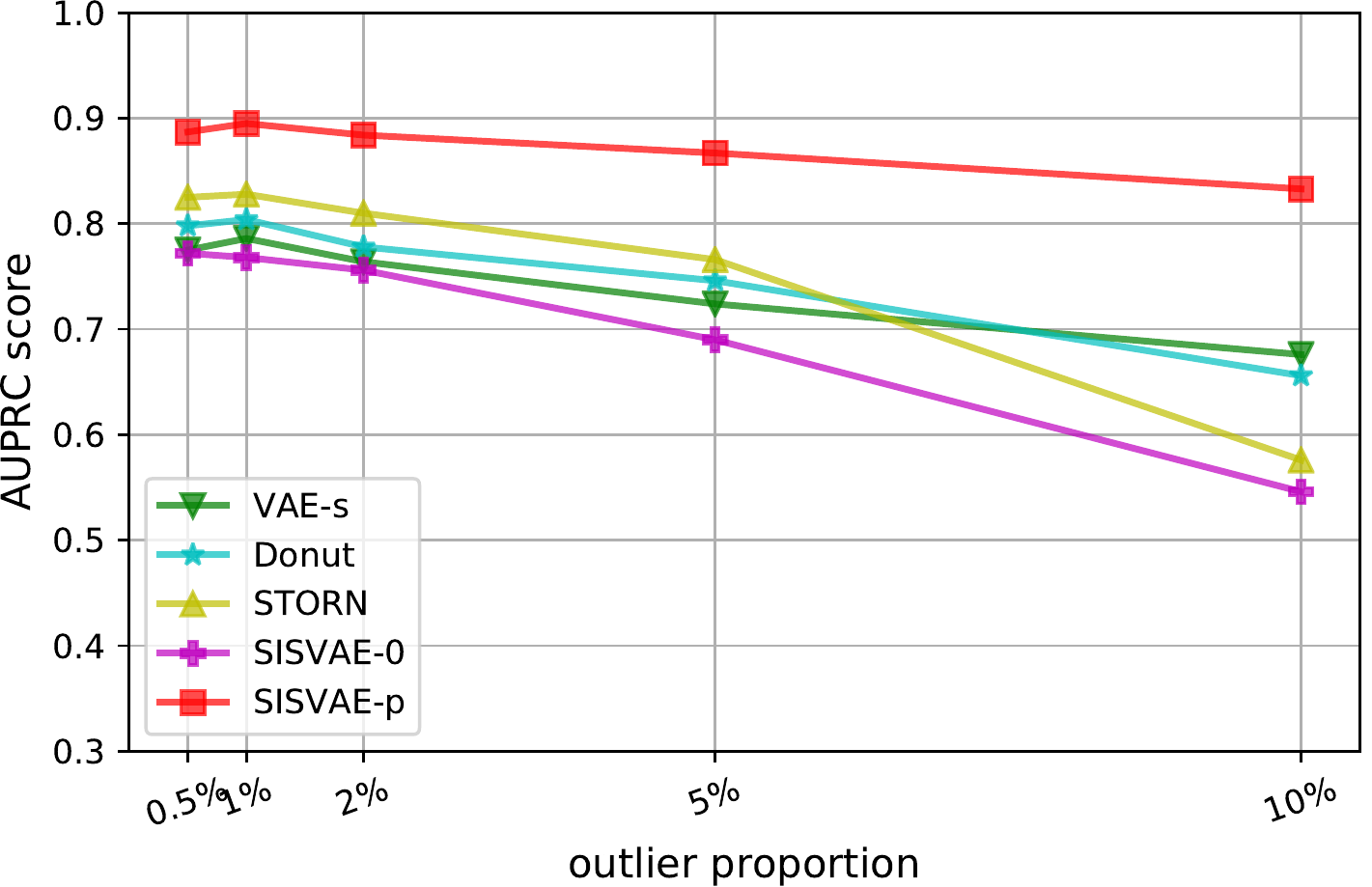}\label{fig:syn_frac2}}
    \subfigure[sensitivity to regularizer]
    {\includegraphics[width=0.245\textwidth]{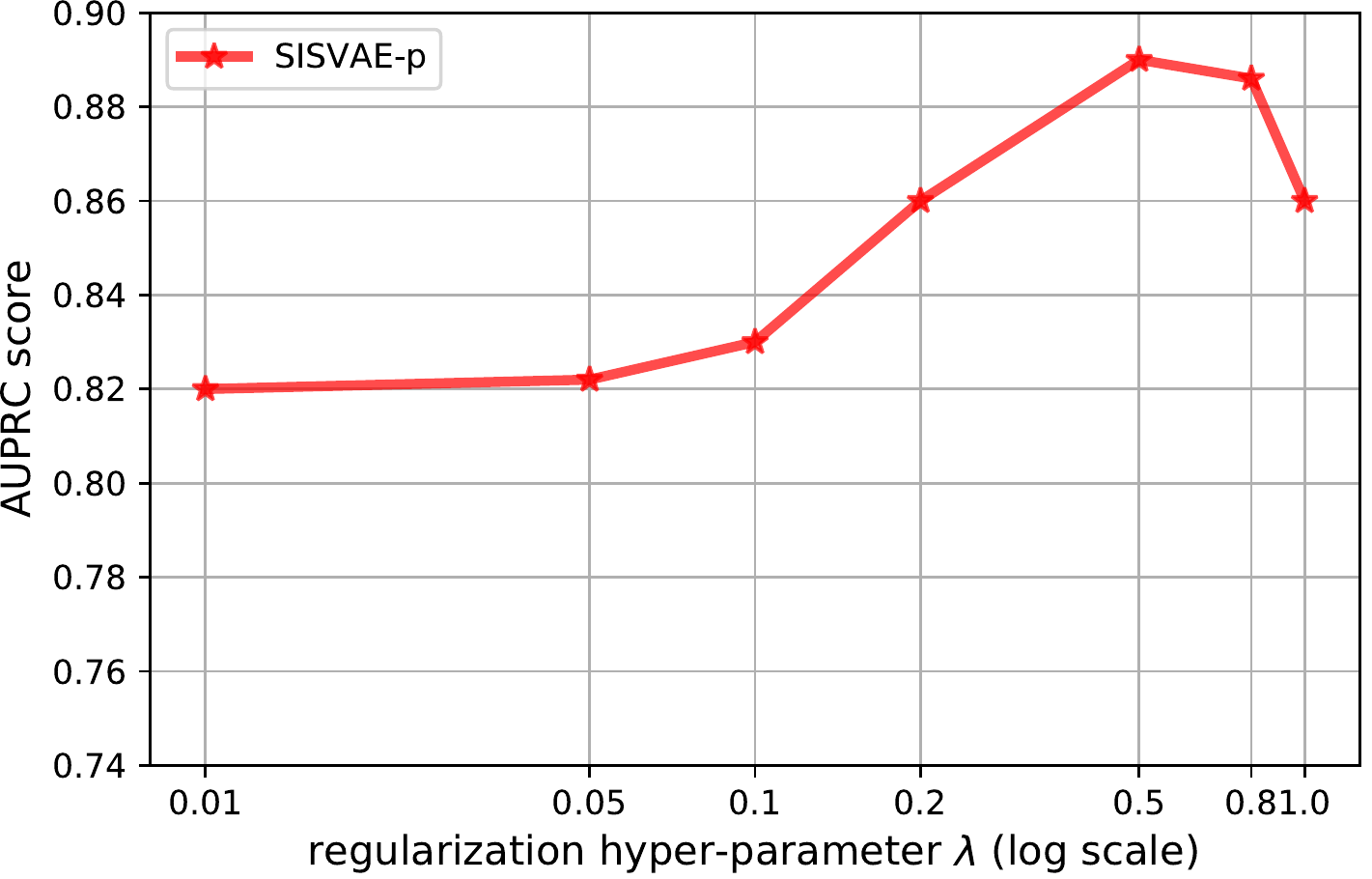}\label{fig:pr_regularization}}
    \subfigure[effects of regularizer]
    {\includegraphics[width=0.245\textwidth]{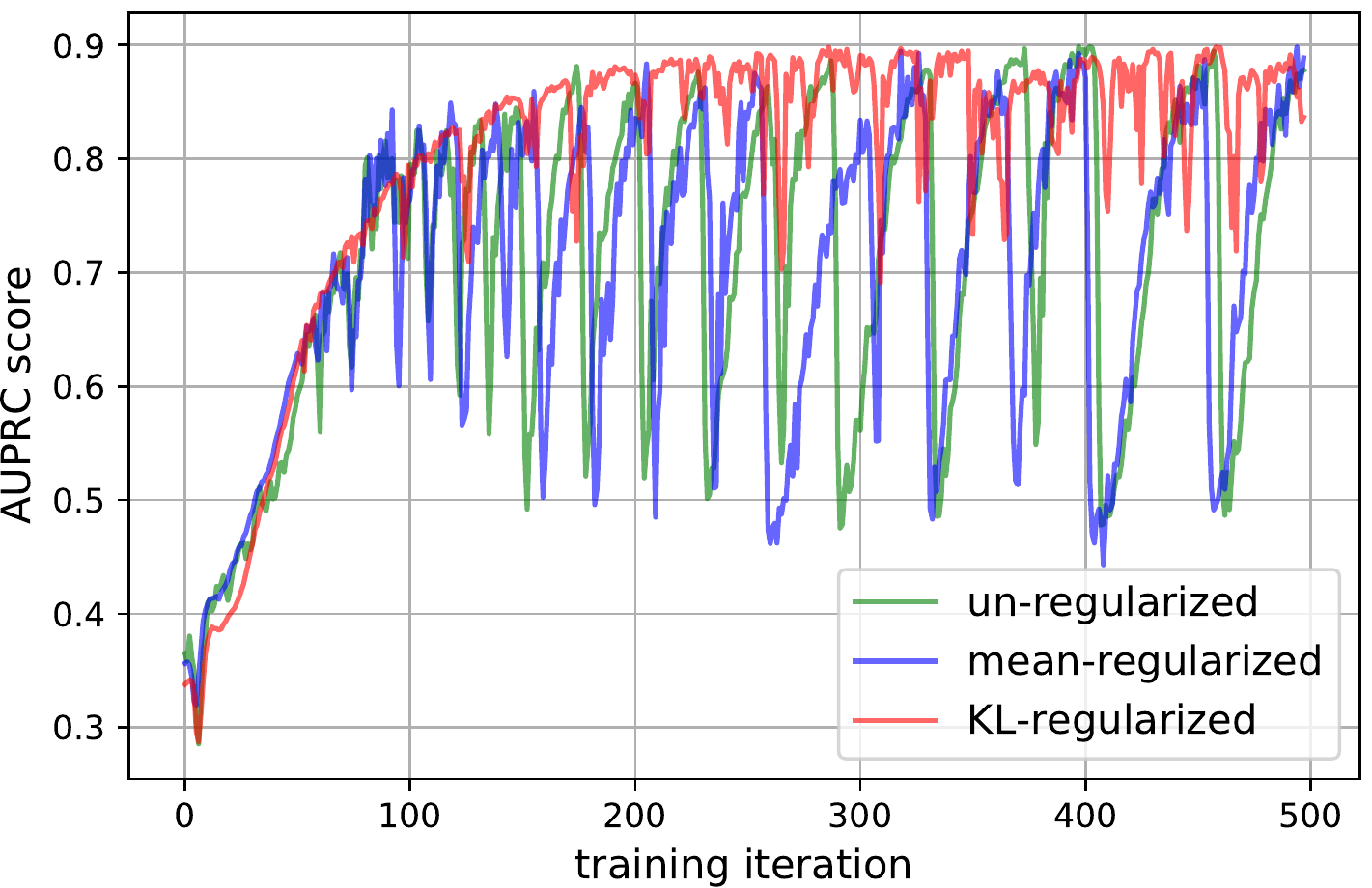}\label{fig:convergence}}
    \subfigure[convergence of losses]
    {\includegraphics[width=0.245\textwidth]{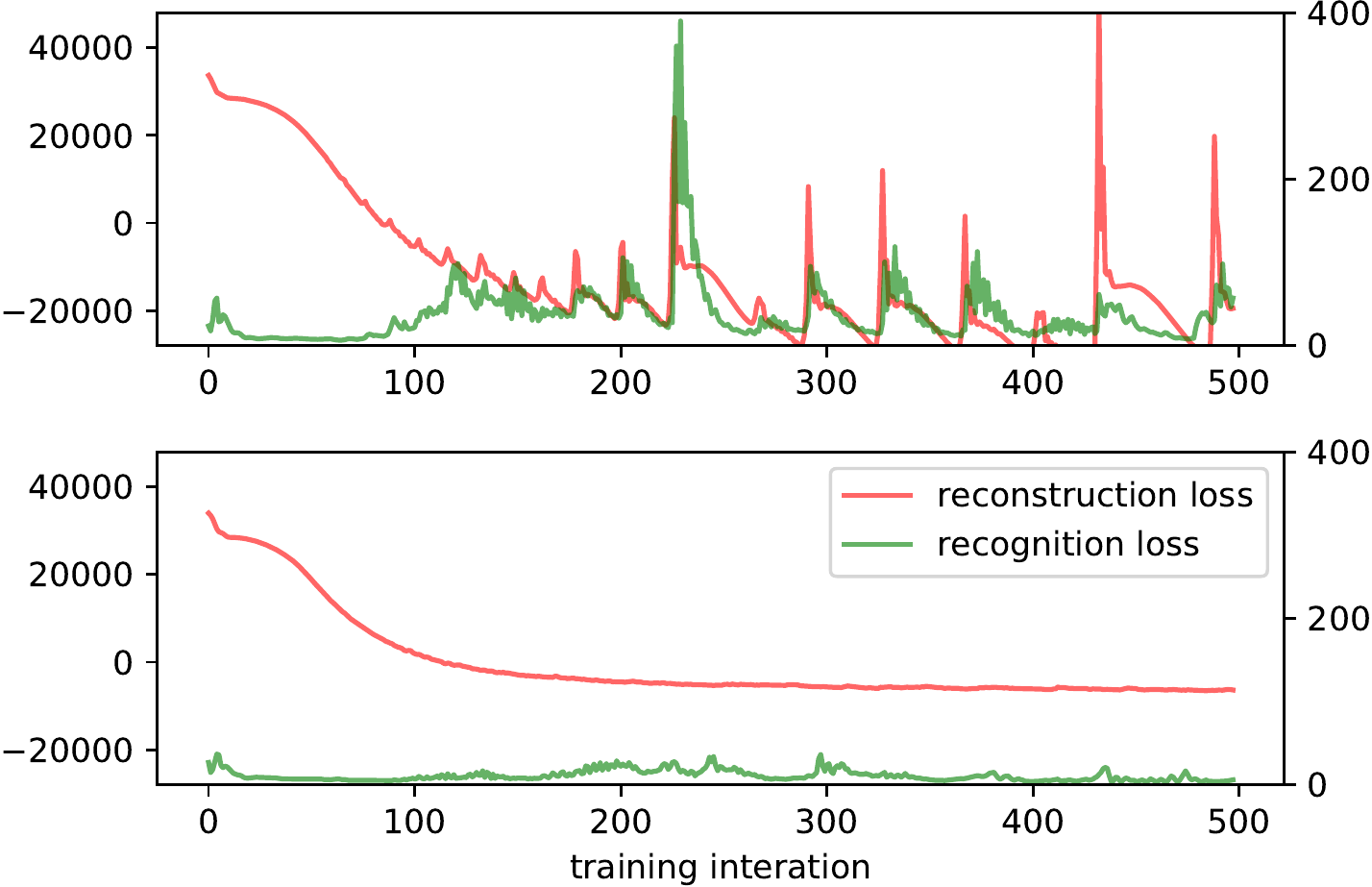}\label{fig:loss_iteration}}
    \vspace{-10pt}\caption{Results of controlled experiment on synthetic datasets. (a) Anomaly detection performance for various proportions of anomalies. (b) Anomaly detection performance of SISVAE with different values of regularization hyper-parameter $\lambda$. (c) AUPRC score over training iterations with different regularizers (KL-regularized is our proposed regularizer and the mean-regularized means only the mean is regularized as done in classical methods). (d) Convergence property of reconstruction term and recognition term of objective function, top:un-regularized model, bottom:KL-regularized model. Here SISVAE denotes the reconstruction probability based model, the result for reconstruction error based method performs worse hence is omitted here.}
    \label{fig:synthetic}
\end{figure*}

In particular, we compare the characteristics of peer methods in Table~\ref{tab:overview}, whereby both traditional methods i.e. History Average (HA), Autoregressive Moving Average (ARMA), Kalman Filter (KMF) and deep latent model i.e. Donut, STORN, SISVAE, SISVAE-mb are included. 

\textbf{Implementation details} For preprocessing, we set window size $W$ to be 120 for each dataset. The sequential VAE based models, STORN, and SISVAE have a similar network architecture, which is defined by two varying hyper-parameters, \textit{h\_dim} for dimension of hidden layer, and \textit{z\_dim} for dimension of latent variable $\bz$. We set \textit{h\_dim} to be 200, and \textit{z\_dim} to be 40 for all datasets. We empirically set the smoothness regularization hyper-parameter $\lambda=0.5$ for SISVAE during training process. The encoder, decoder, and RNN of models use the same hidden dimension. The three models are implemented using PyTorch~\cite{paszke2019pytorch} 0.4, and trained on a single Nvidia RTX 2080ti GPU. We train the three models 200 epochs using Adam \cite{kingma2014adam} with learning rate 0.001. Note for Donut, the M-ELBO technique proposed by Donut may require some anomaly labels to learn `normal' patterns. For fair comparison, we provide no anomaly label to the model, and the experiments are conducted under fully unsupervised setting. We use the code open sourced by the authors. We set \textit{h\_dim} to be 200, and \textit{z\_dim} to be 40, same as our model. The model is fundamentally designated for uni-variate time series and it is nontrivial to extend it to the multi-dimensional case. We test two settings: 1) train a model for each sequence individually, and compute anomaly score individually; 2) flatten the multi-dimensional time series data after preprocessing to accommodate the model, the anomaly score is computed together. We find the second protocol exhibits better performance. This may be because the normal patterns are common along different sequences. With more sequences as input, the model can be less influenced by anomalies.  We adopt the second protocol for Donut in the experiments. For all deep latent variable models, we use negative reconstruction probability as anomaly score output. For all three traditional statistical models, we use the same preprocessing method. For history average (HA), the anomaly score is absolute observation value since all series have zero mean. For ARMA and LDS, we use absolute reconstruction error as anomaly score.
\begin{table}[tb!]
  \begin{center}
    \caption{Statistics of the used datasets.}
    \label{tab:dataset}
    \begin{tabular}{l|c|c|c|c|c}
      \toprule
      \textbf{Dataset} & \textbf{A1} & \textbf{A2} & \textbf{A3} & \textbf{A4} & \textbf{$\mu$PMU}\\
      \hline
      \# sequences   & 65    & 100   & 100   & 100   & 36\\
      \# length       & 1420  & 1421  & 1680  & 1680  &   1080\\
      \# points          & 92300 & 142100& 168000& 168000&   38800\\
      \# anomalies         & 1462  & 426   & 943   & 1044  &   240\\
      \# proportion&1.58\% & 0.32\%& 0.56\%    & 0.62\%&   0.62\%\\
    \bottomrule
    \end{tabular}
  \end{center}
\end{table}
\begin{figure*}[t!]
    \centering
    \subfigure[AUROC]
    {\includegraphics[width=0.28\textwidth]
    {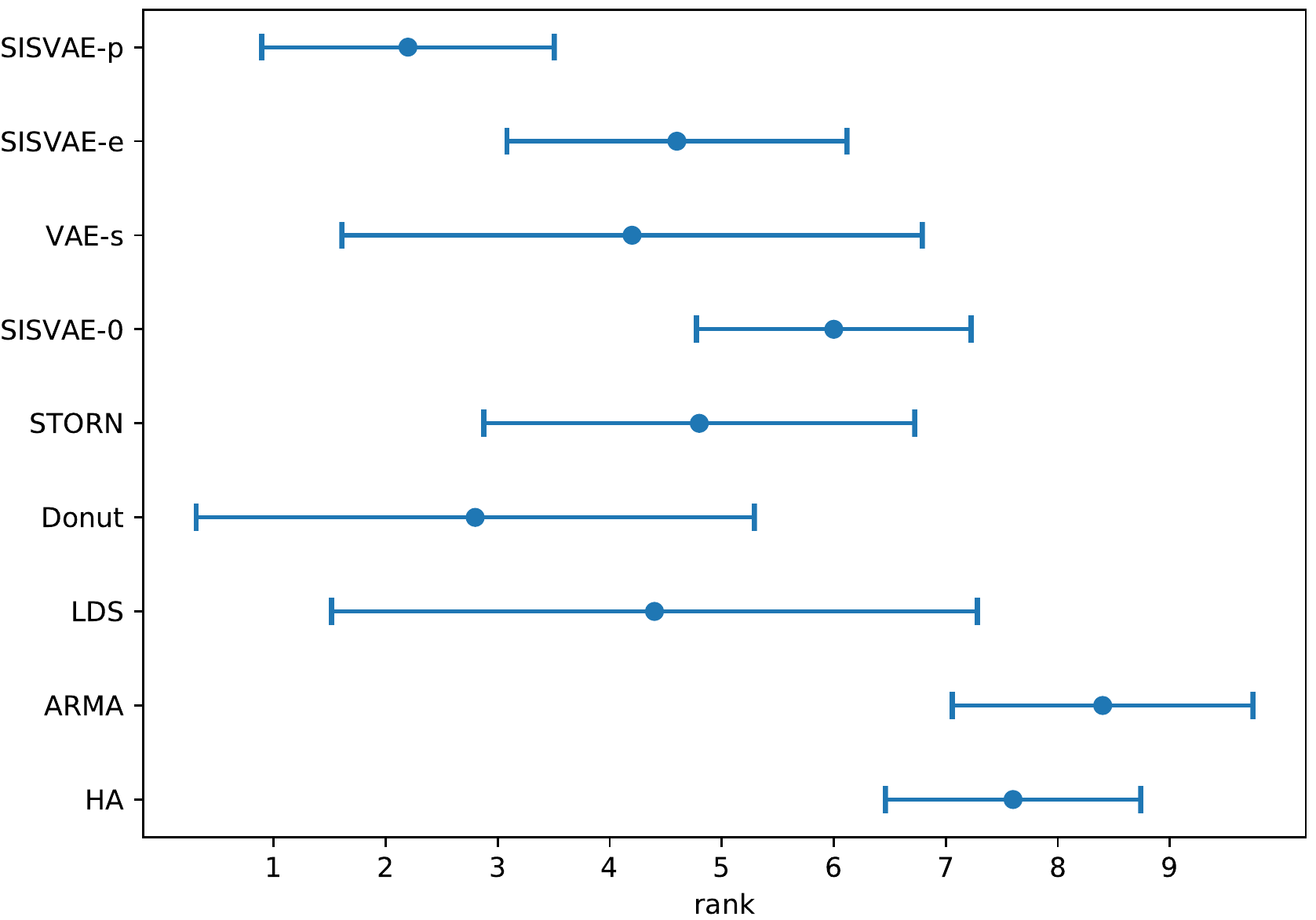}\label{fig:rank_auroc}}
    \subfigure[AUPRC]
    {\includegraphics[width=0.28\textwidth]{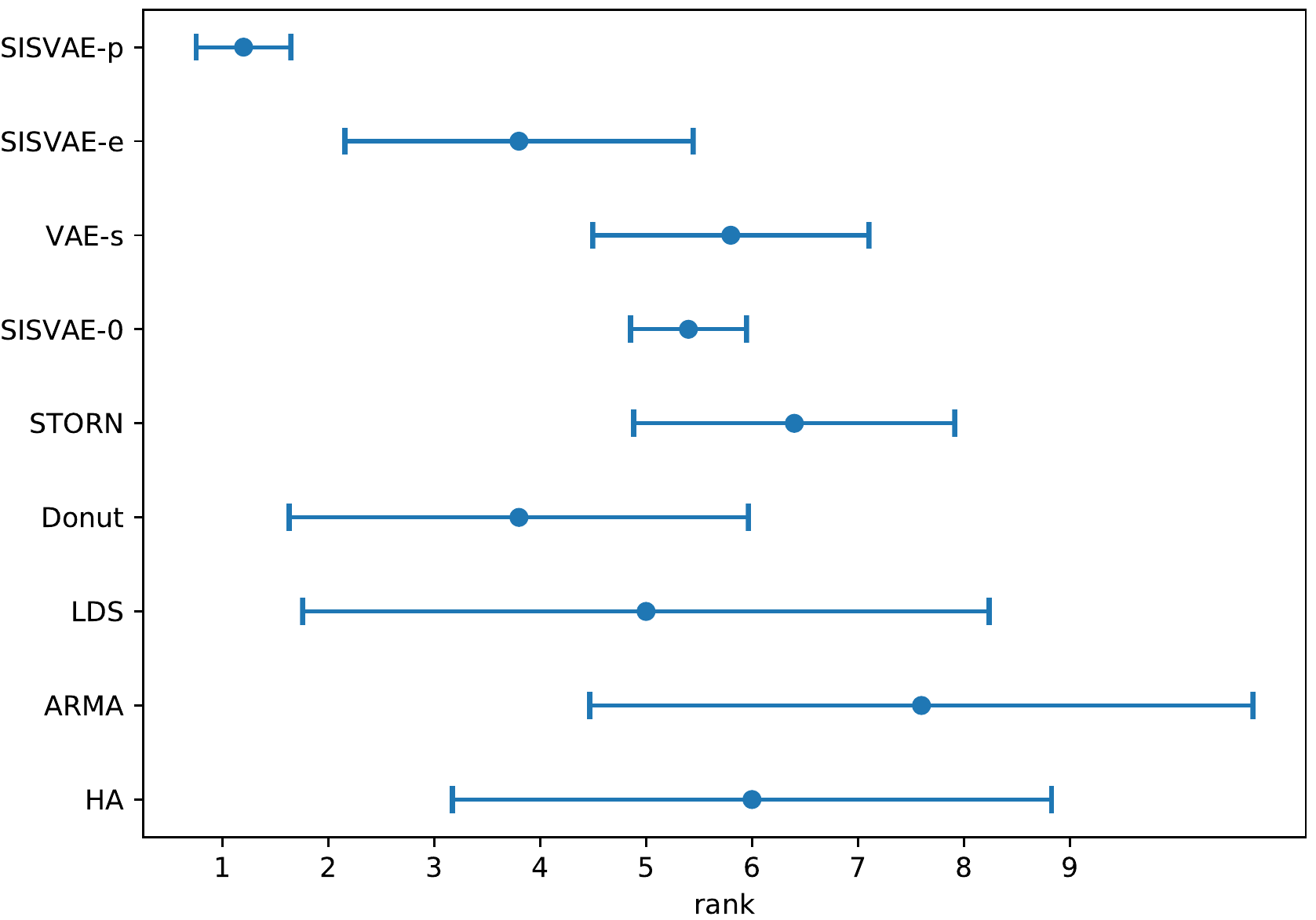}\label{fig:rank_auprc}}
    \subfigure[F1]
    {\includegraphics[width=0.28\textwidth]{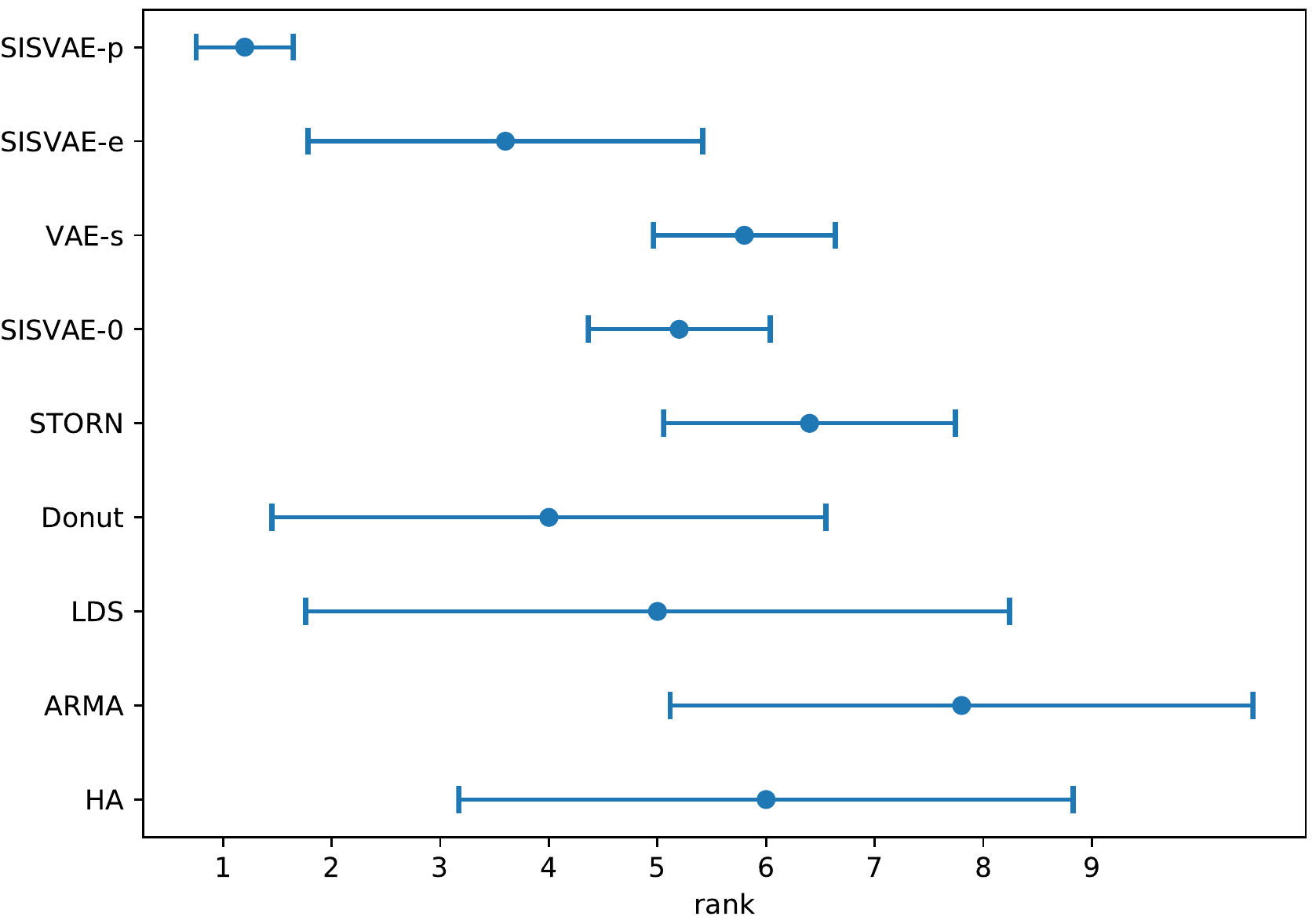}\label{fig:rank_f1}}
    \caption{Ranking of models with respect to three evaluation metrics. Bar ranges indicate the standard variations centered with their ranking mean.}\label{fig:rank_plot}.
\end{figure*}

\subsection{Experiments on Synthetic Data}
In this section, we aim to gain insights into the behavior of our proposed SISVAE model. We investigate the anomaly detection performance of our model for different proportions of anomalies. Furthermore, we are interested in the anomaly detection performance under different settings of smoothness regularization hyper-parameter $\lambda$.

Referring the synthetic data generation protocol in~\cite{gornitz2015hidden}, we generate 100 time series sequences of length 200 from a multi-output Gaussian process with a multiple output kernel \cite{alvarez2012kernels}, implemented by a python package named GPy \cite{gpy2014}. Sequences are correlated through a random sampled positive definite coregionalization matrix, and we add random non-stationary Gaussian noise on the 100-dimensional sequence of length 200. The location of anomalies is simulated by a binary mask with Bernoulli distribution of probability \{0.5\%, 1\%, 2\%, 5\%, 10\%\} for different fractions of anomalies. We insert Poisson noise where the binary mask is equal to one as anomalies, and the intensity $\lambda_t$ of each anomaly is proportional to the amplitude of inserted place $x_t$. The plus and minus sign is random sampled from Bernoulli distribution with $p=0.5$.

\textbf{Effects of anomaly proportion.}
Model performance variation with respect to the proportion of anomalies is shown in Fig. \ref{fig:syn_frac2}. Since all the models are unsupervised, they are inevitably influenced by the anomalies, because a full reconstruction of both normal and anomaly data is encouraged in the objective function. The proposed variational smoothness regularizer places penalties at consecutive outputs of the generative model that violate the temporal smoothness assumption. As observed from the result, as the anomaly proportion grows, the AUPRC scores of two un-regularized model SISVAE-0 and STORN drop significantly, while the regularized models' performance almost keeps maintained. SISVAE-0, the proposed SISVAE model shows superior performance over competing methods in all settings of anomaly proportion.

\textbf{Effects of regularization hyper-parameter.}
The variational smoothness hyper-parameter $\lambda$ in the objective by Eq. \ref{eq:objective} determines the penalty of non-smooth output of the generative model. We analyze the detection performance of SISVAE model under different settings of hyper-parameters. We train the model with $\lambda$ = \{0.01,0.05,0.1,0.2,0.5,0.8,1.0\} using synthetic dataset with 2\% anomaly proportion. Figure \ref{fig:pr_regularization} shows the change of AUPRC performance w.r.t $\lambda$, where the model achieves best performance when $\lambda$ is set to be 0.5.

\begin{table*}[t!]
\centering
\caption{Performance on real-world public datasets. SISVAE-e is omitted for its worse performance than SISVAE-p.}
\label{tab:real_perf}
\begin{tabular}{c|c|ccccccccc}
\toprule
Dataset                   & Metric & HA    & ARMA  & LDS & Donut      & STORN & SISVAE-0  & VAE-s      & SISVAE-e & SISVAE-p    \\\hline
\multirow{3}{*}{A1}       & AUROC  & 0.616 & 0.845 & 0.664  & 0.825      & 0.867 & 0.862 & \tb{0.886} & 0.865   & 0.876      \\
                          & AUPRC  & 0.276 & 0.305 & 0.099  & 0.178      & 0.166 & 0.184 & 0.185      & 0.300   & \tb{0.369} \\
                          & F1     & 0.337 & 0.351 & 0.140  & 0.215      & 0.240 & 0.254 & 0.254      & 0.358   & \tb{0.401} \\\hline
\multirow{3}{*}{A2}       & AUROC  & 0.956 & 0.658 & 0.952& \tb{0.999} & 0.926 & 0.974 & 0.993      & 0.989   & 0.981      \\
                          & AUPRC  & 0.745 & 0.233 & 0.401  & 0.623      & 0.435 & 0.608 & 0.536      & 0.731   & \tb{0.817} \\
                          & F1     & 0.774 & 0.457 & 0.480  & 0.691      & 0.515 & 0.588 & 0.611      & 0.709   & \tb{0.789} \\\hline
\multirow{3}{*}{A3}       & AUROC  & 0.724 & 0.486 & 0.989 & 0.990      & 0.917 & 0.912 & 0.907      & 0.951   & \tb{0.993} \\
                          & AUPRC  & 0.252 & 0.005 & 0.835 & 0.753      & 0.732 & 0.746 & 0.713      & 0.865   & \tb{0.954} \\
                          & F1     & 0.331 & 0.011 & 0.831 & 0.798      & 0.757 & 0.791 & 0.763      & 0.880   & \tb{0.951} \\\hline
\multirow{3}{*}{A4}       & AUROC  & 0.715 & 0.487 & 0.979 & \tb{0.991}  & 0.850 & 0.834 & 0.836      & 0.831   & 0.924 \\
                          & AUPRC  & 0.107 & 0.005 & 0.737 & 0.598      & 0.438 & 0.521 & 0.525      & 0.521   & \tb{0.774} \\
                          & F1     & 0.165 & 0.010 & 0.745 & 0.704      & 0.478 & 0.582 & 0.573      & 0.582   & \tb{0.820} \\\hline
\multirow{3}{*}{$\mu$PMU} & AUROC  & 0.629 & 0.587 & 0.805  & 0.714      & 0.684 & 0.628 & 0.629      & 0.630   & \tb{0.814} \\
                          & AUPRC  & 0.084 & 0.023 & 0.303  & \tb{0.312} & 0.185 & 0.174 & 0.089      & 0.153   & 0.310      \\
                          & F1     & 0.225 & 0.049 & 0.356  & \tb{0.389} & 0.279 & 0.247 & 0.228      & 0.244   & 0.367
\\\bottomrule
\end{tabular}
\end{table*}

\textbf{Convergence study.} We analyze the contribution of our regularizer to convergence property. On the synthetic 2\% anomaly proportion dataset, we train three models: i) SISVAE-p without regularization ($\lambda = 0$). ii) classical smoothness regularizer of deterministic model where only the mean is regularized which is formulated as: $\lambda\sum_{t=1}^{T}\sum_{m=1}^{M}(\nabla^2_t \hat{x}_{m,t})^2$. iii) proposed KL regularizer. We record the AUPRC score over training. As shown in Fig.~\ref{fig:convergence}, at the first 100 iterations, all the three models are trained very fast, and the detection performance increases stably. This suggests that the models are learning the major `normal' patterns of the data, while the anomalies are temporally neglected. However, after 100 iterations, the performance of the un-regularized and mean-regularized models fluctuate severely between 0.45 and 0.85, and our KL-regularized model (i.e. SISVAE-p) also fluctuates but in a relatively small interval. We think the fluctuation is caused by contradiction between generative (decoder) model and recognition (encoder) model: the generative model is trying hard to reconstruct every data point perfectly, while the inference model aims to map the observed anomaly-contaminated data points into a simpler low-dimensional latent space. The decoder and encoder are playing a seesaw battle in the training procedure, thus causing a sawtooth shape AUPRC score. Under the probabilistic framework, classical mean-regularizer which is in its nature deterministic, cannot work well. Its convergence curve is akin to the un-regularzied model. In contrast, the proposed variational smoothness regularizer weakens the reconstruction objective, and guides the decoder to generate sequences that obey temporal smoothness assumption, preventing the model from overfitting `anomaly' patterns. As shown in Fig.~\ref{fig:loss_iteration}, the reconstruction loss and inference loss of regularized model converge stably, while the training of un-regularized model is unstable. Hence the KL-regularized SISVAE model reaches a higher AUPRC score.

\begin{figure*}[t!]
    \centering
    \subfigure[Yahoo-A1]
    {\includegraphics[width=0.192\textwidth]
    {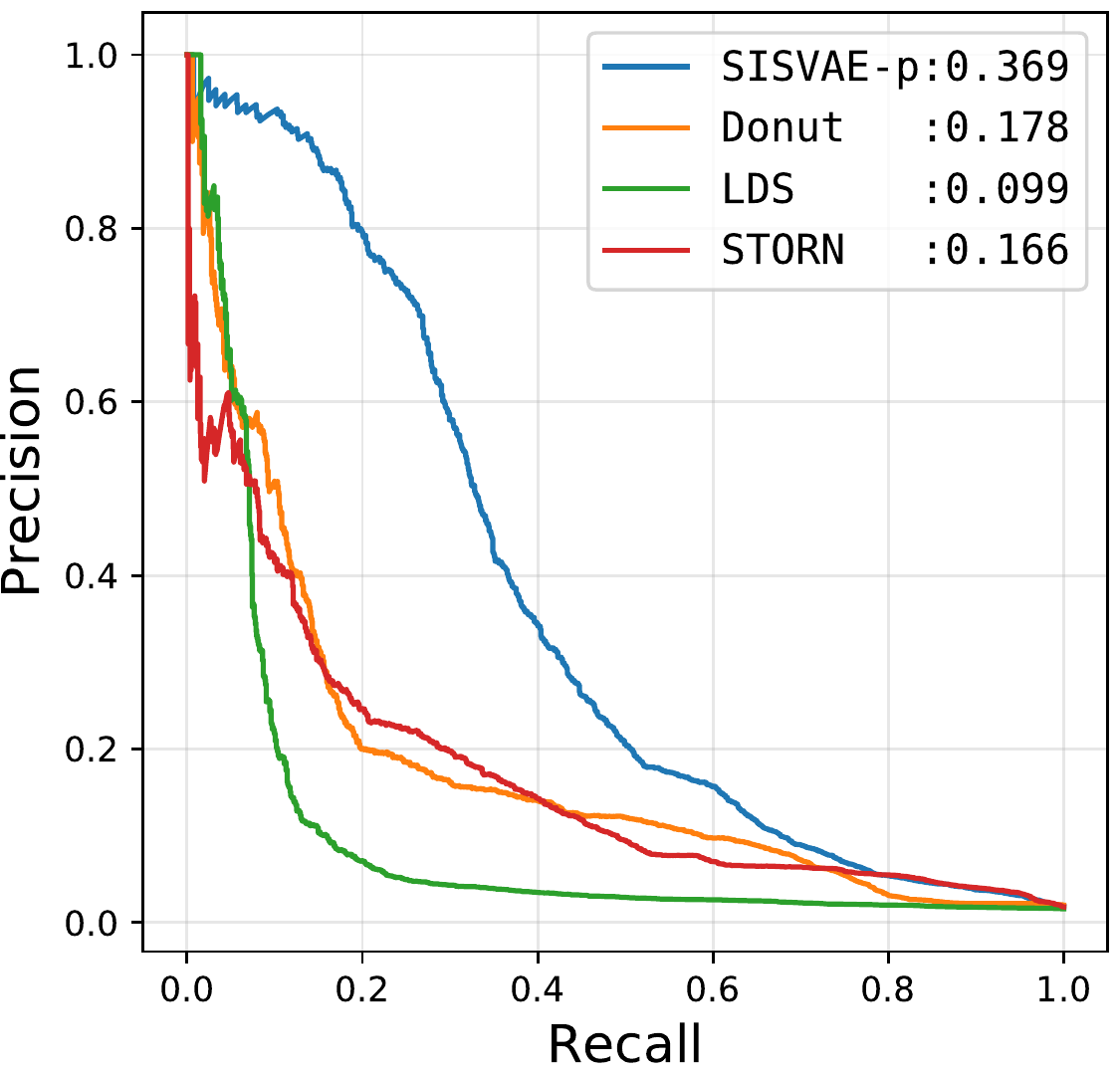}\label{fig:prc_a1}}
    \subfigure[Yahoo-A2]
    {\includegraphics[width=0.192\textwidth]{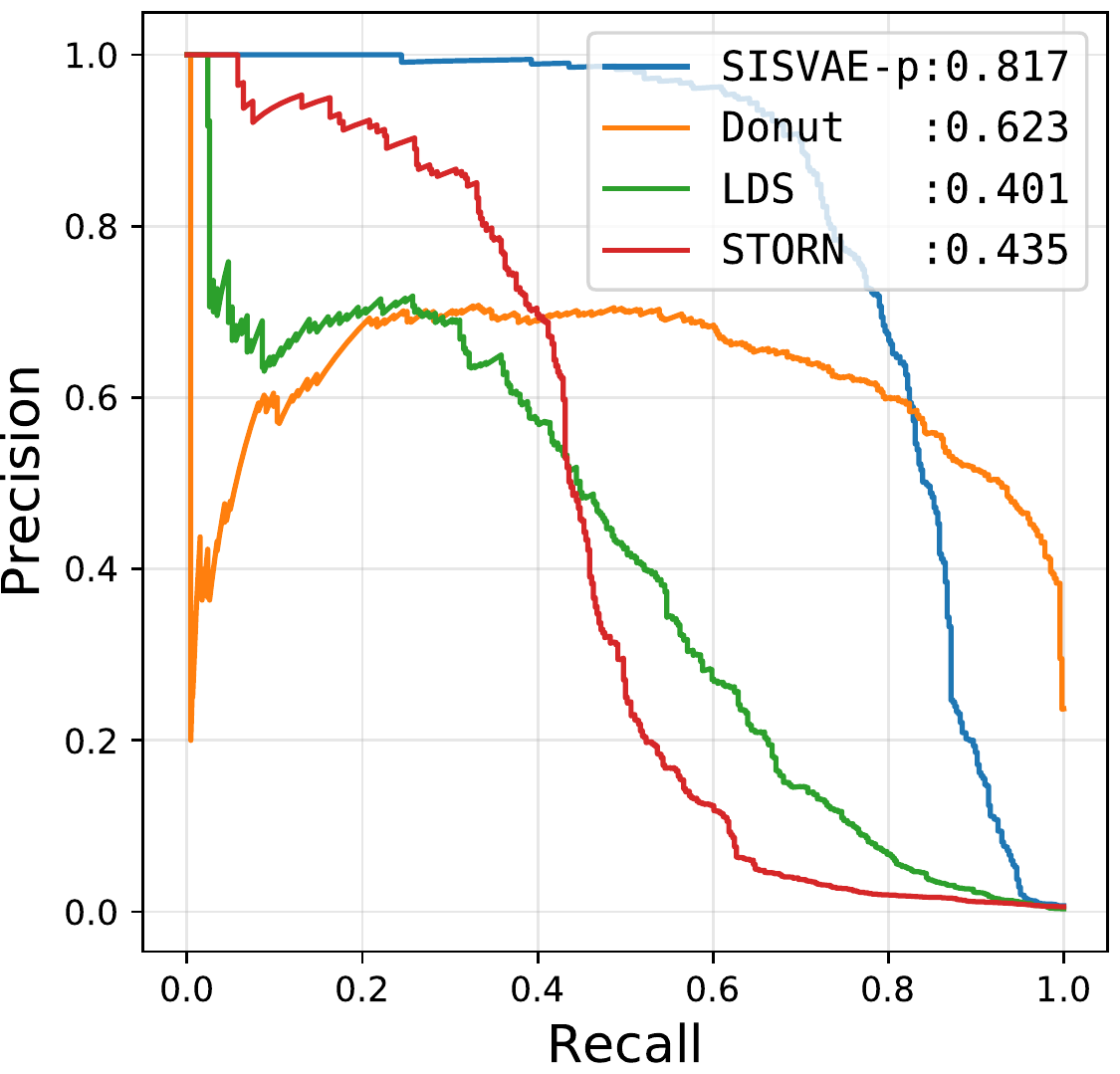}\label{fig:prc_a2}}
    \subfigure[Yahoo-A3]
    {\includegraphics[width=0.192\textwidth]{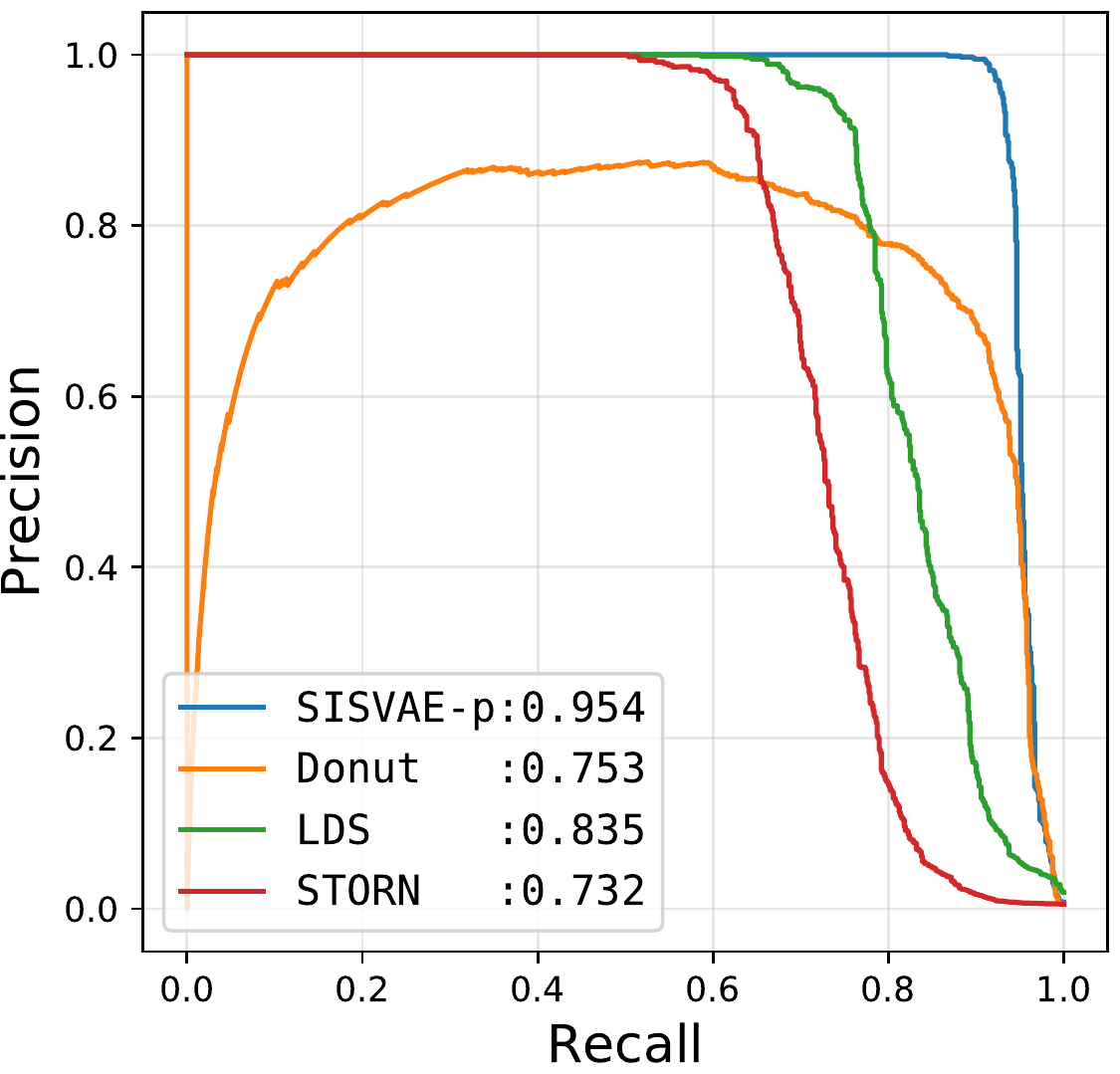}\label{fig:prc_a3}}
    \subfigure[Yahoo-A4]
    {\includegraphics[width=0.192\textwidth]{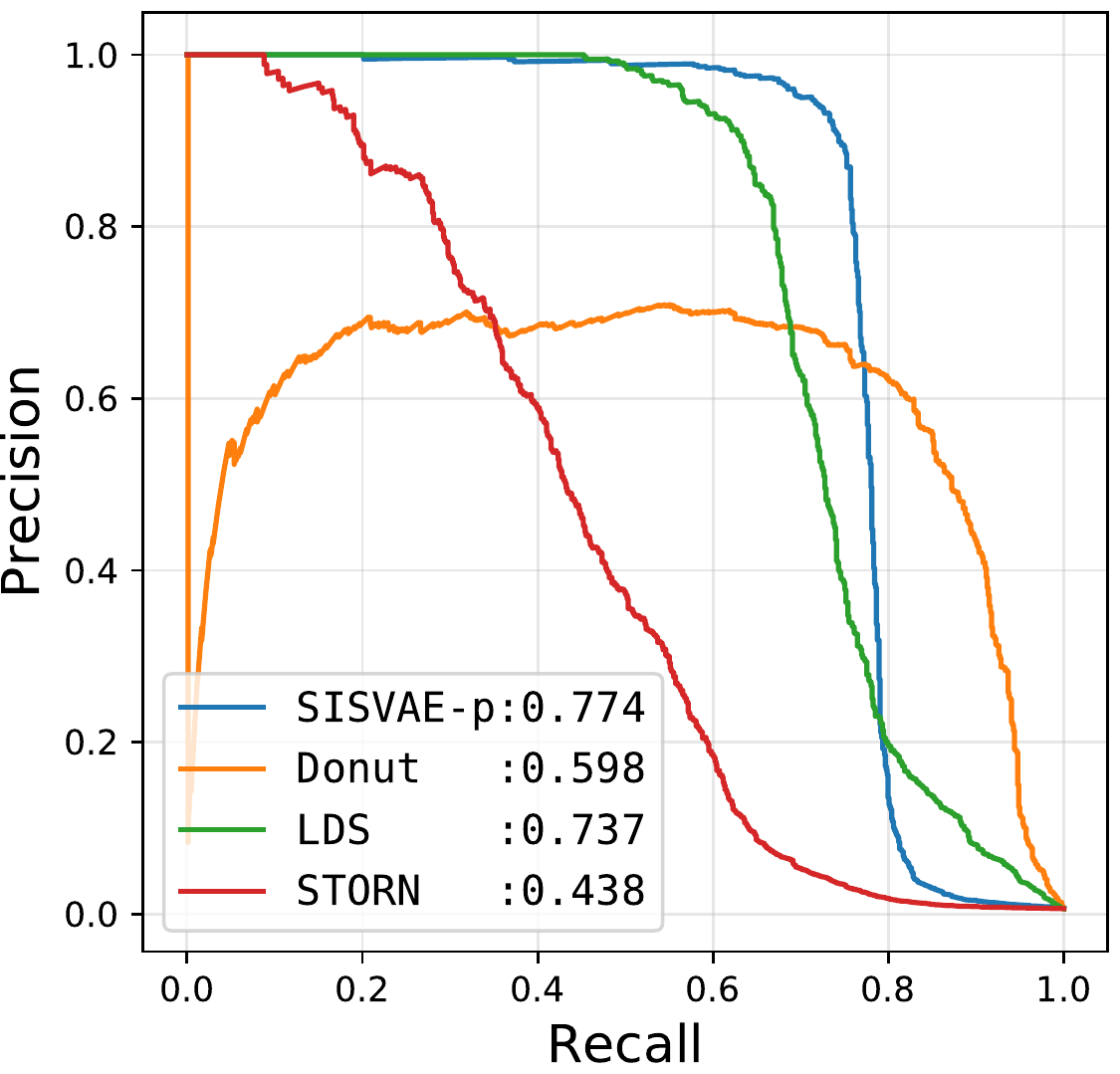}\label{fig:prc_a4}}
    \subfigure[$\mu$PMU]{
    \includegraphics[width=0.192\textwidth]{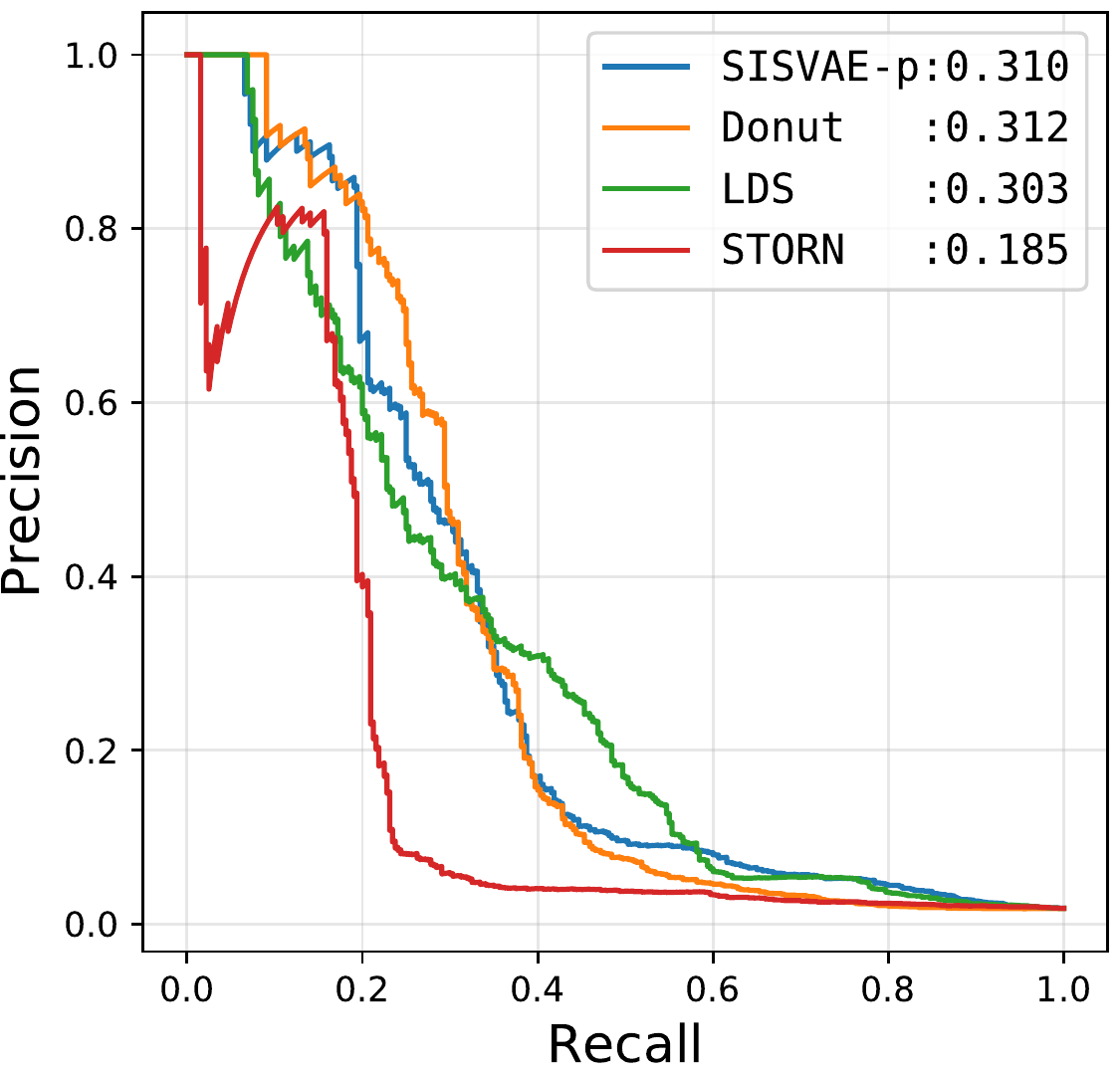}\label{fig:prc_a5}}
    \caption{Visualization of precision and recall curve of SISVAE-p, Donut, LDS, and STORN on real-world datasets.}
    \label{fig:auprc}
\end{figure*}
\begin{figure*}[t!]
    \centering
    \subfigure[Yahoo-A1]
    {\includegraphics[width=0.192\textwidth]
    {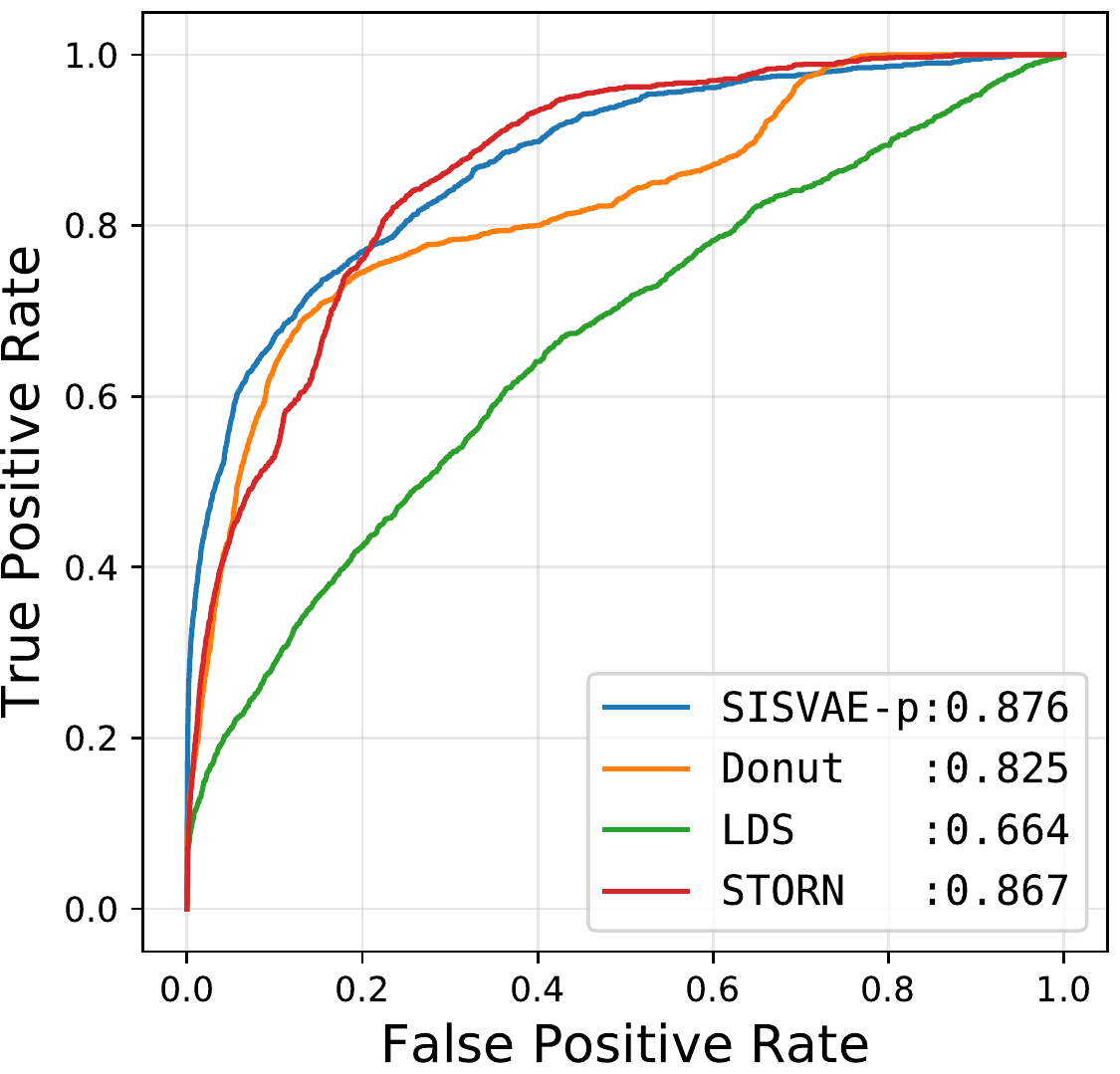}\label{fig:roc_a1}}
    \subfigure[Yahoo-A2]
    {\includegraphics[width=0.192\textwidth]{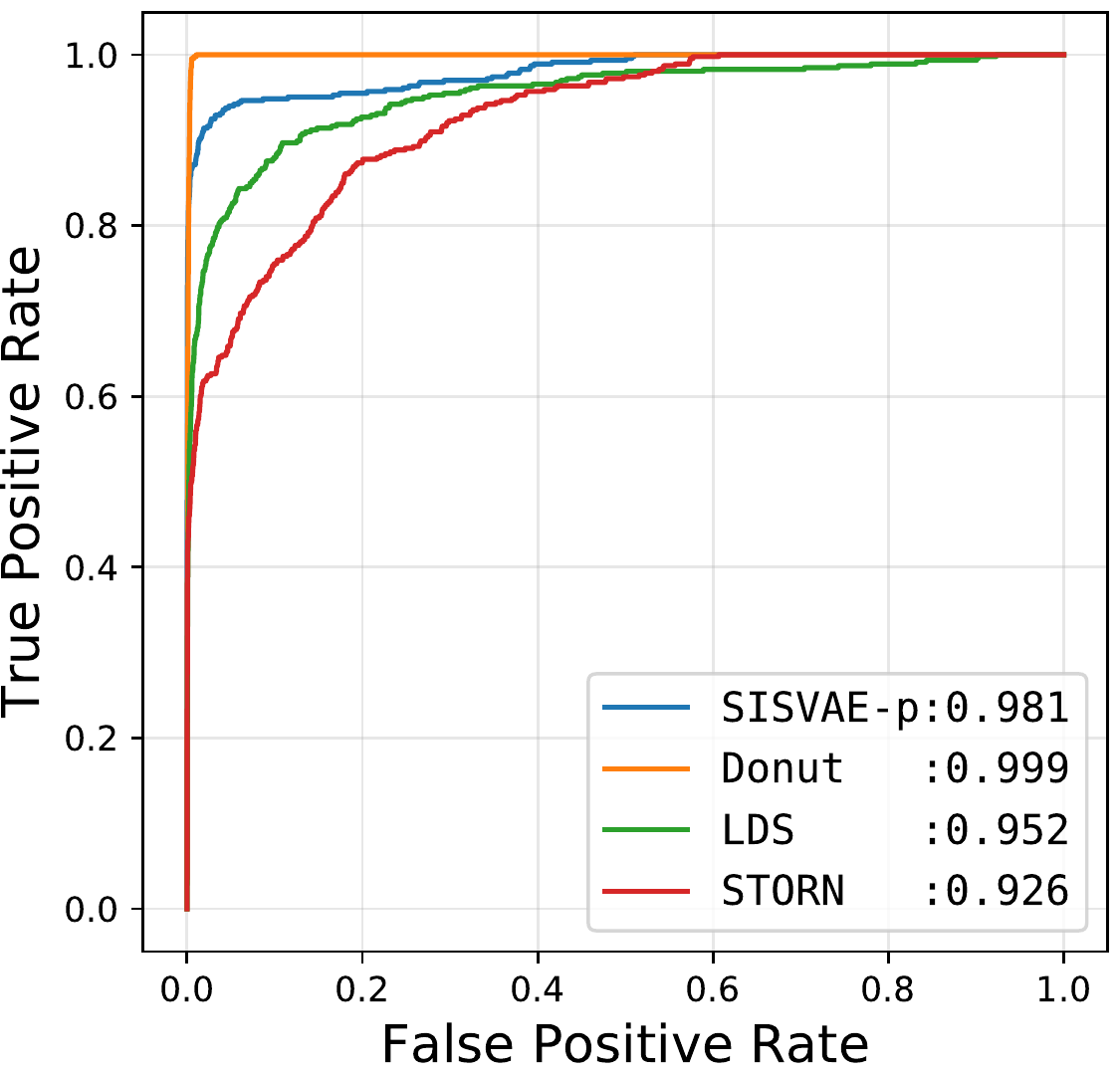}\label{fig:roc_a2}}
    \subfigure[Yahoo-A3]
    {\includegraphics[width=0.192\textwidth]{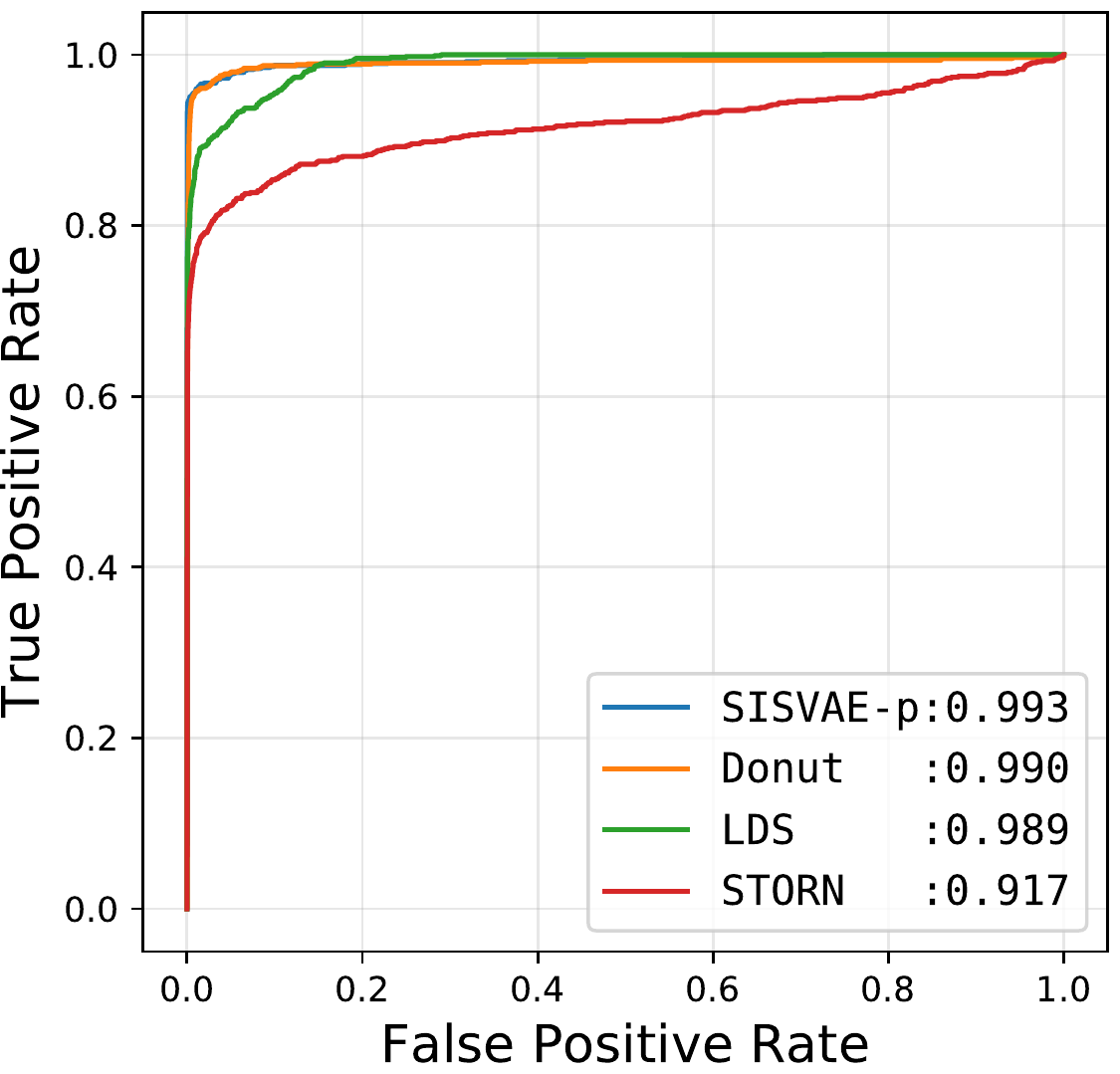}\label{fig:roc_a3}}
    \subfigure[Yahoo-A4]
    {\includegraphics[width=0.192\textwidth]{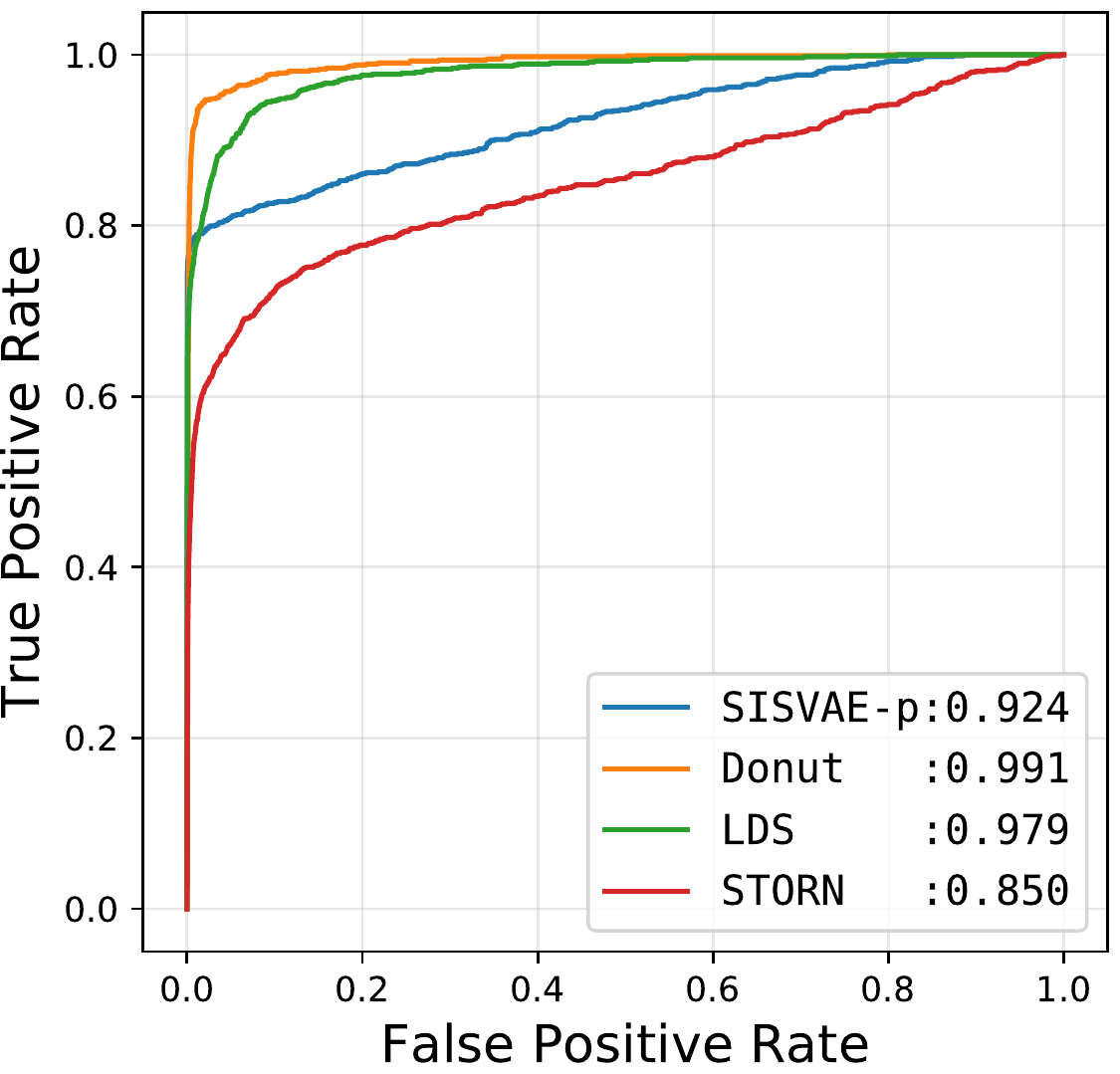}\label{fig:roc_a4}}
    \subfigure[$\mu$PMU]{
    \includegraphics[width=0.192\textwidth]{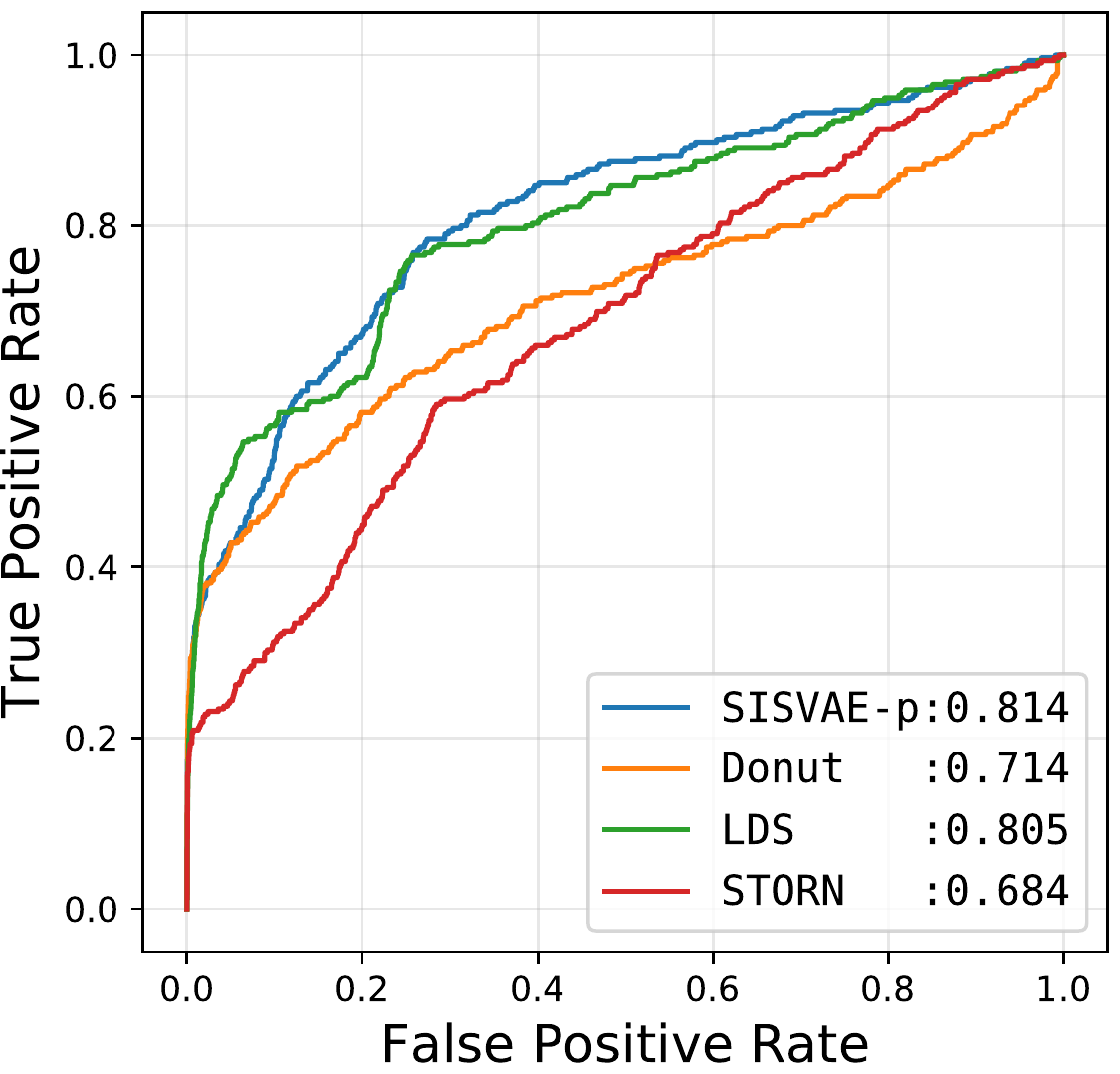}\label{fig:roc_a5}}
    \caption{Visualization of receiver operating characteristic (ROC) of SISVAE-p, Donut, LDS, and STORN on real-world datasets.}
    \label{fig:auroc}
\end{figure*}
\subsection{Experiments on Real-world Data}
\textbf{Datasets}
This experiment involves the following datasets, we conclude statistics of datasets in Table \ref{tab:dataset}.
\begin{itemize}
    \item \textbf{Yahoo's S5 Webscope Dataset:} This dataset is a part of Yahoo's Webscope program\footnote{https://webscope.sandbox.yahoo.com/}. It consists of 367 time series. This dataset consists of four different classes A1/A2/A3/A4 with 65/100/100/100 time series respectively. Class A1 has real data collected from computational systems, which records real production traffic of Yahoo's website, and the anomalies are labelled by human expert manually. Classes A2, A3 and A4 contain synthetic data, whereby anomalies are constructed and are inserted at random positions.

\item \textbf{Open $\mu$PMU power network dataset:  } This dataset is collected from a power distribution equipped in Alameda, CA, USA with advanced smart meters called phasor measurement units ($\mu$PMU)\footnote{https://powerdata.lbl.gov} \cite{stewart2016open}. There are three $\mu$PMU devices, each recording three-phase voltage and current magnitude and phase angle. Anomalies caused by voltage disturbance are labeled by human experts. We collect measurement data and anomaly labels during the whole day of June 3 with 20 seconds resolution. We use 36-dimensional time series of length 1080.
\end{itemize}

\textbf{Preprocessing} In real application, different time series may have different scales, e.g. temperature and air pressure sensors have different scales. So we follow the common practice to normalize each sequence to standard normal distribution. 

\begin{figure*}[ht!]
    \centering
    \subfigure[Yahoo-A1]
    {\includegraphics[width=0.192\textwidth]{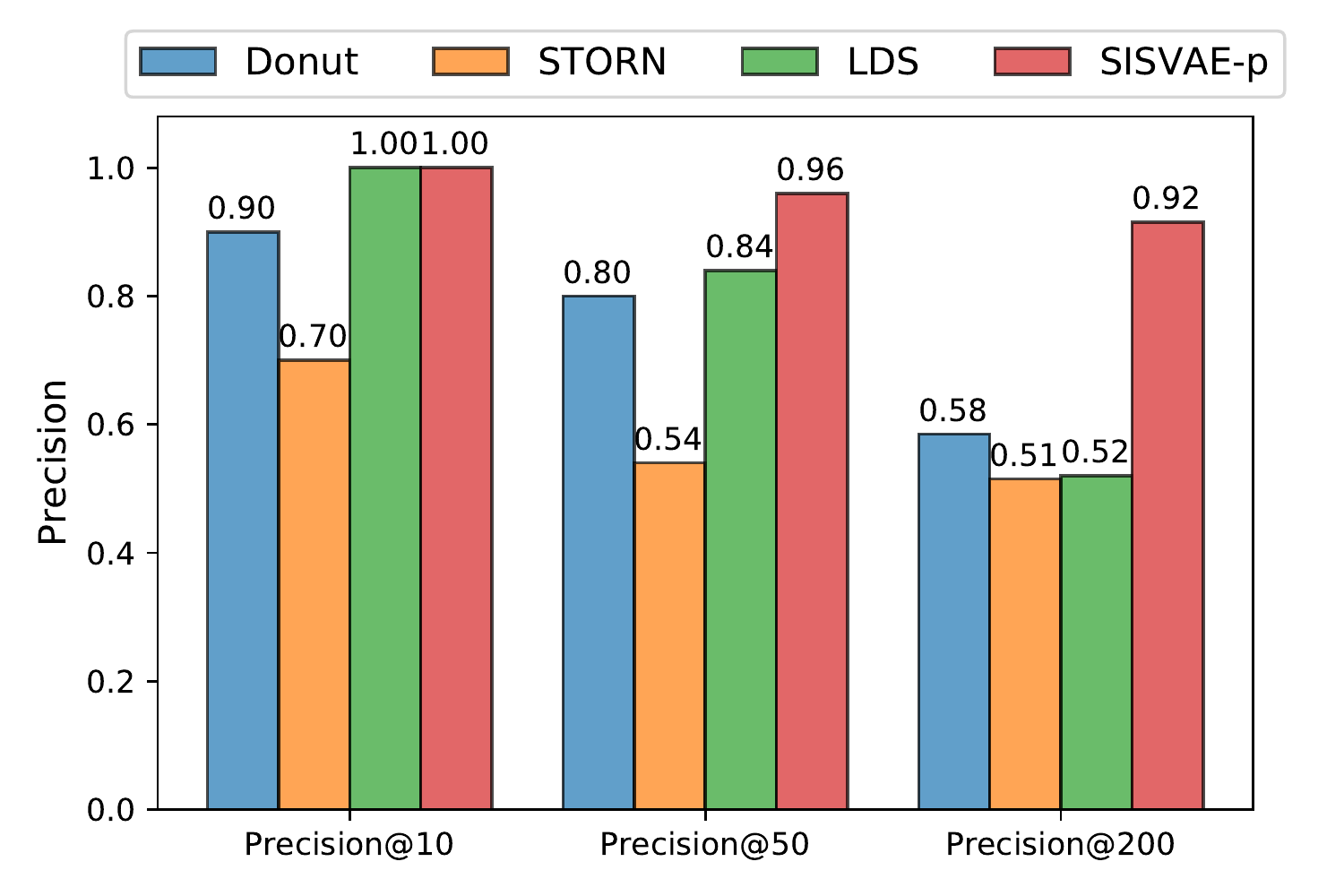}\label{fig:rank_A1}}
    \subfigure[Yahoo-A2]
    {\includegraphics[width=0.192\textwidth]{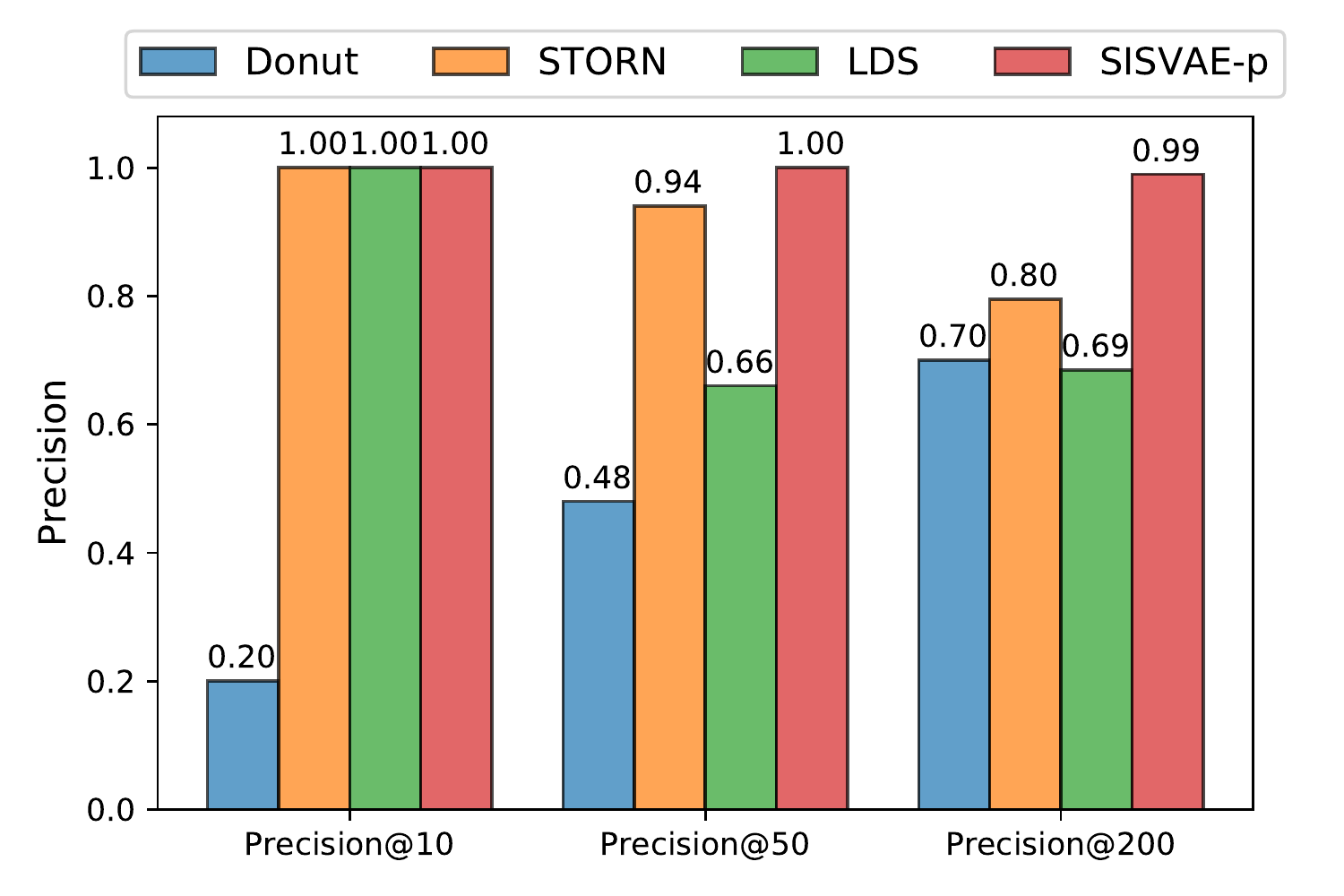}\label{fig:rank_A2}}
    \subfigure[Yahoo-A3]
    {\includegraphics[width=0.192\textwidth]{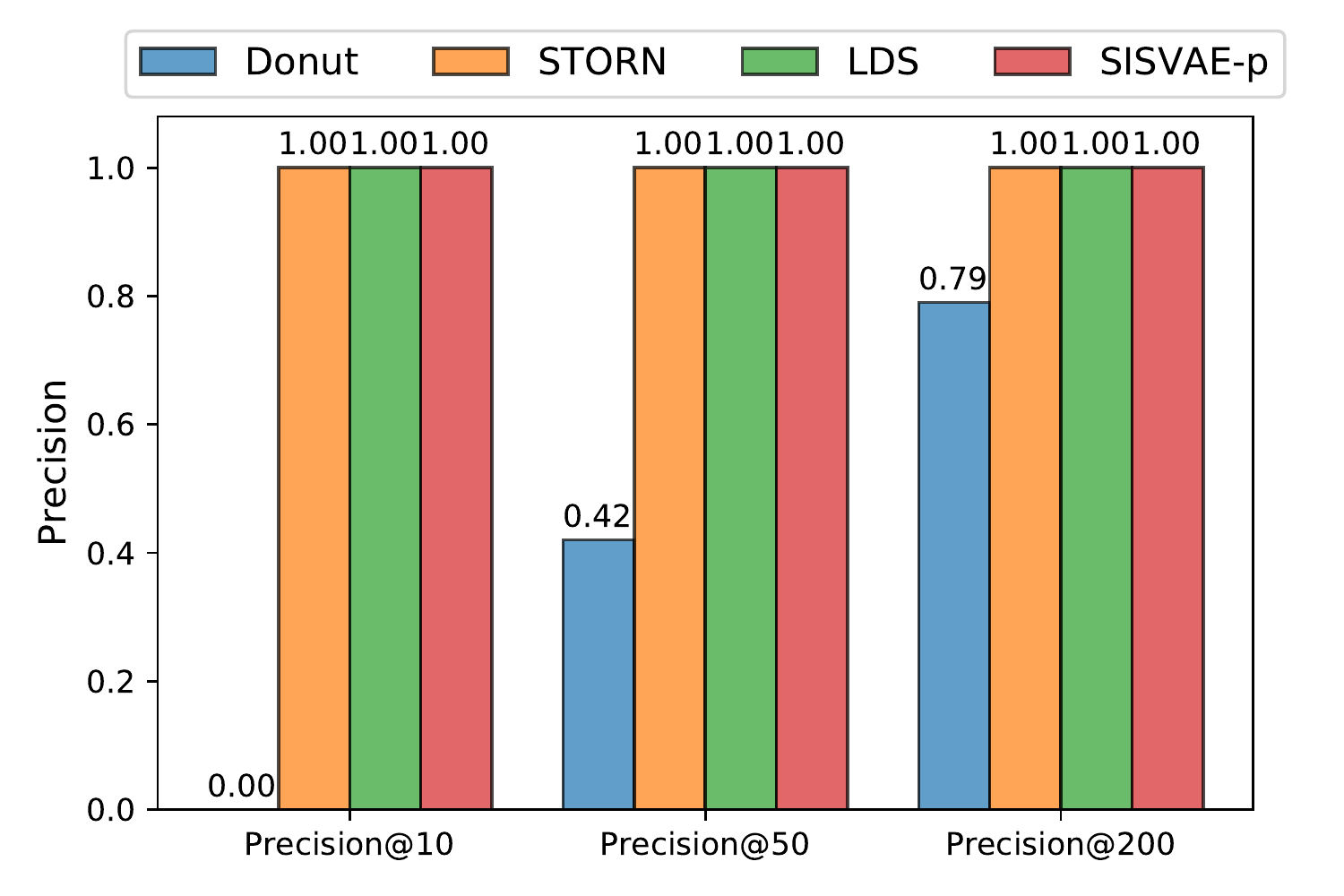}\label{fig:rank_A3}}
    \subfigure[Yahoo-A4]
    {\includegraphics[width=0.192\textwidth]{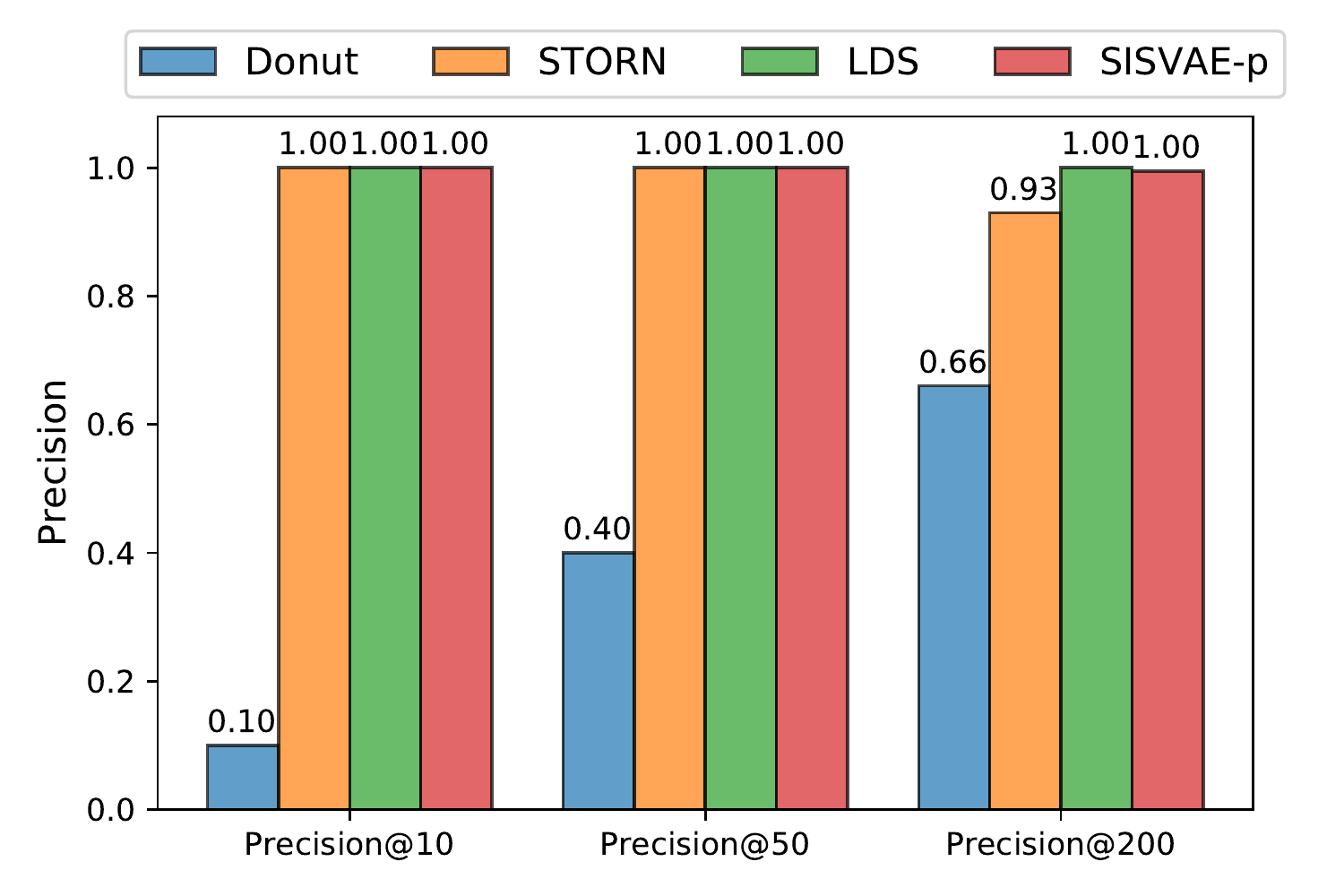}\label{fig:rank_A4}}
    \subfigure[$\mu$PMU]
    {\includegraphics[width=0.192\textwidth]{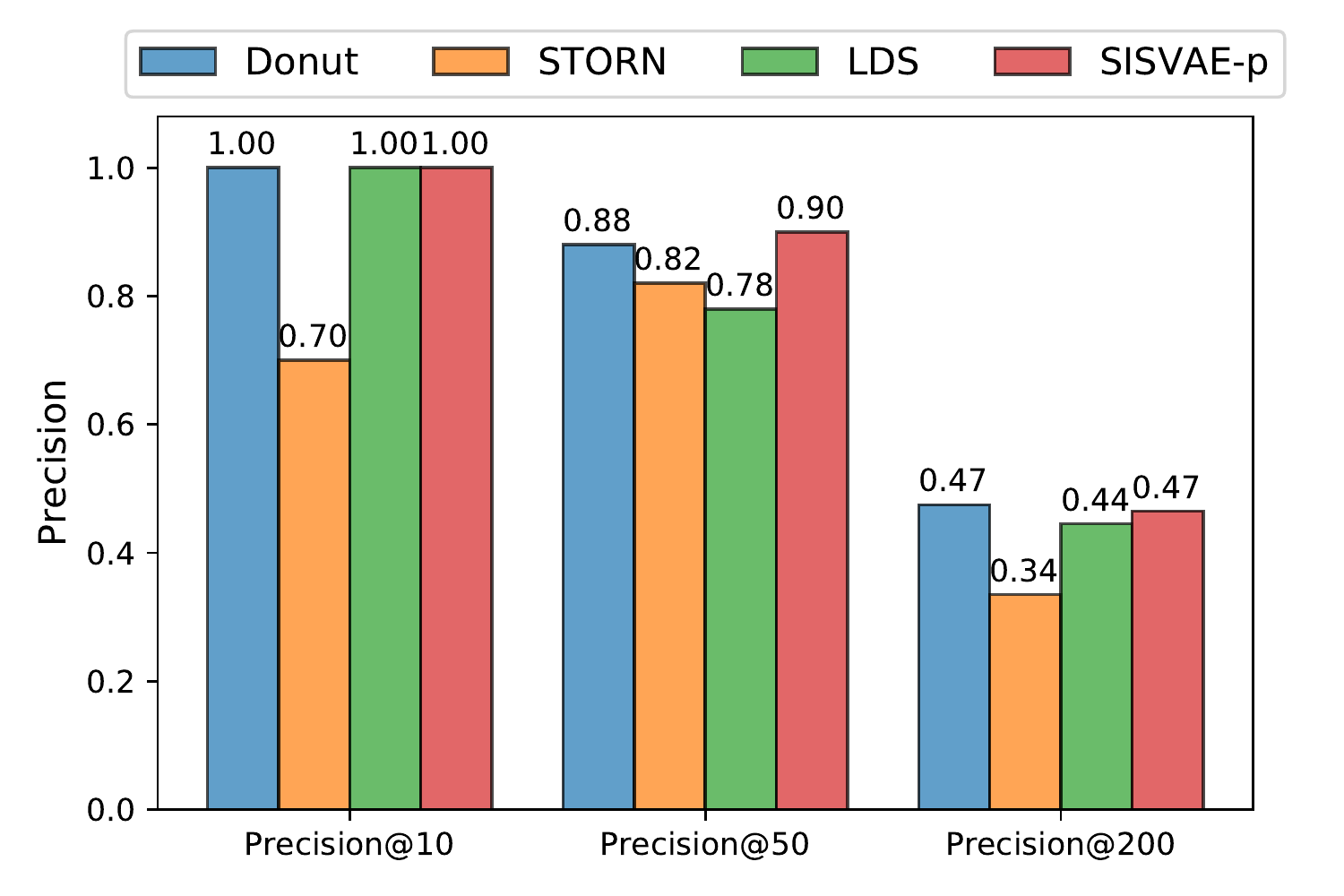}\label{fig:rank_uPMU}}
    \caption{Visualization of precision@K for ranking based detection tasks on real-world datasets.}
    \label{fig:precisionK}
\end{figure*}
\textbf{Threshold based anomaly detection.} First, we conduct threshold based anomaly detection experiment. For each model, we collect anomaly score matrix $\bm{A}$ for each model, then we compute AUROC, AUPRC and F1-score for each possible threshold $\alpha$. We select the highest F1-score the model could achieve as the final F1-score. We compare all the peer methods in this experiment, whose results are shown in Table \ref{tab:real_perf}. SISVAE-p outperforms other models in most datasets and metrics. For SISVAE-p and SISVAE-e, we find that our proposed SMC based reconstruction probability for computing anomaly score notably improves the performance. Among all the five datasets, we find SISVAE-p and SISVAE-e perform similar on A2 dataset. We further find that the A2 dataset is in general stationary regarding with variance, which means that modeling the time-varying variance is important for anomaly detection. For traditional statistical models, we find the simple method HA shows decent performance in some datasets, however, the potential of HA is very limited. We are surprised that ARMA failed on A3, A4 and $\mu$PMU dataset. Our further study uncovers that these three datasets are highly non-stationary with both mean and variance, which violate the basic stationary assumption of ARMA. In contrast, we find LDS performs competitively on A3, A4 and $\mu$PMU comparing to the best model, however, it fails on dataset A1. This suggests that linear transition and linear emission probability of LDS may not be able to capture complex non-linear relationship. For VAE-s, the proposed smoothness prior regularizer works even without a temporal structure model. In our analysis, the regularizer constructs inductive bias of the true signal, leading to robustness to unknown anomaly. 

As discussed, we find models' anomaly detection performance varies across datasets, such as LDS and ARMA. To make the overall performance comparison more intuitionistic, we compute the rank of models for each dataset and performance metric, and compute the mean and standard deviation of ranks for each models. The plots of rankings are shown in Fig. \ref{fig:rank_plot} showing our model outperforms other models notably. 

\textbf{Ranking based anomaly detection.} Due to limited budget, sometimes we can only deal with a limited number of detected anomalies hence the precision need be high. In this context, precision@K is a useful metric, which is widely used in information retrieval \cite{baeza2011modern}. For each model, we select data points that are assigned with top $K$ highest anomaly score, and precision@K is computed by ${V}/{K}$, where $V$ is the number of true anomalies in $K$ predictions. We examine models with $K = {10, 50, 200}$, and the results are shown in Fig. \ref{fig:precisionK}. We can see that, when $K$ is 10, STORN, LDS and SISVAE-p perform well. However, Donut fails on A2, A3 and A4 dataset. The reason is that Donut places very high anomaly score on some normal data points. Overall, SISVAE-p's precision keeps stable as $K$ increases. This means SISVAE-p makes more precise detection when the number of detection trials is limited.

\textbf{Model characteristics.} The detection threshold $\alpha$ is often set according to specific needs. In order to investigate the detailed detection characteristic, we plot the precision-recall curve (PRC), and receiver operating characteristic curve (ROC), as shown in Fig. \ref{fig:auprc} and Fig. \ref{fig:auroc}. Comparing to LDS and STORN, SISVAE-p outperforms at most thresholds. For precision and recall curve in Yahoo A2, A3 and A4 dataset, note that Donuts performs strangely, where the precision and recall curves do not monotonically decrease as expected, but increase first and then decrease. We find this is because of Donut places very large anomaly scores to some normal points, while the other three models do not have this problem. This means modeling latent temporal dependencies is important for time series anomaly detection tasks. 

\textbf{The trade-off between recall and precision.} Despite the fact that SISVAE-p is better than baselines in most cases, however, the preference of detection metric also affects the choice of model. It is important to know when to use which model. The cost of validating a detected anomaly can be expensive which calls for model's high precision. For large-scale power grid system, it is required the precision need be strictly above 0.95. In this case, SISVAE-p outperforms other methods in recall. In contrast, for fraud detection, recall can be more important. If we set a hard bar for recall to 0.95, we find Donut is also competitive, it outperforms SISVAE-p in A2 and A4 dataset w.r.t. precision. 


\subsection{Case study}
We compare the three deep generative models, SISVAE-p, STORN, and Donut with real-world examples from Yahoo-A1 dataset. For each model, we choose to set threshold $\alpha$ when F1-score is maximized. The dataset consists of 65 sequences each with length 1420. In what follows we show some representative time series fragments for discussion.
\begin{figure}[tb!]
    \centering
    \subfigure[studied dimension of the 65-dim time series labeled with ground truth anomaly windows (stretched  twice compared to (b) and (c))]
    {\includegraphics[width=0.48\textwidth]
    {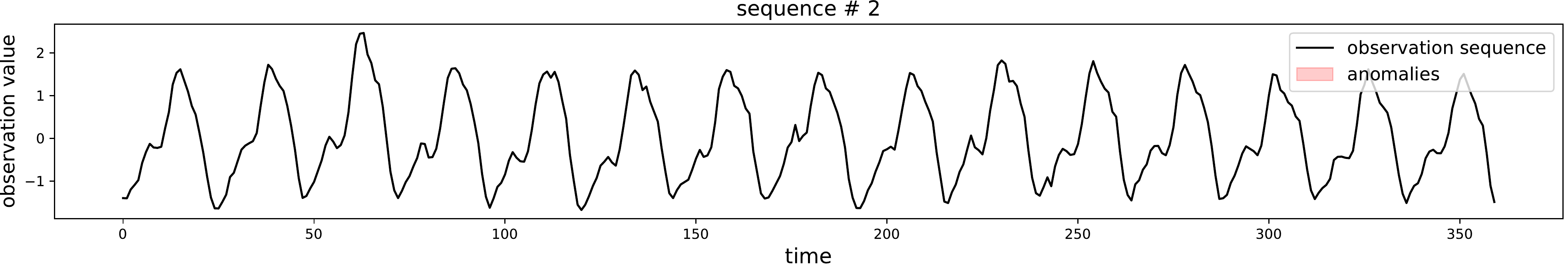}\label{fig:seq1_true}}
    \subfigure[density estimation]
    {\includegraphics[width=0.23\textwidth]{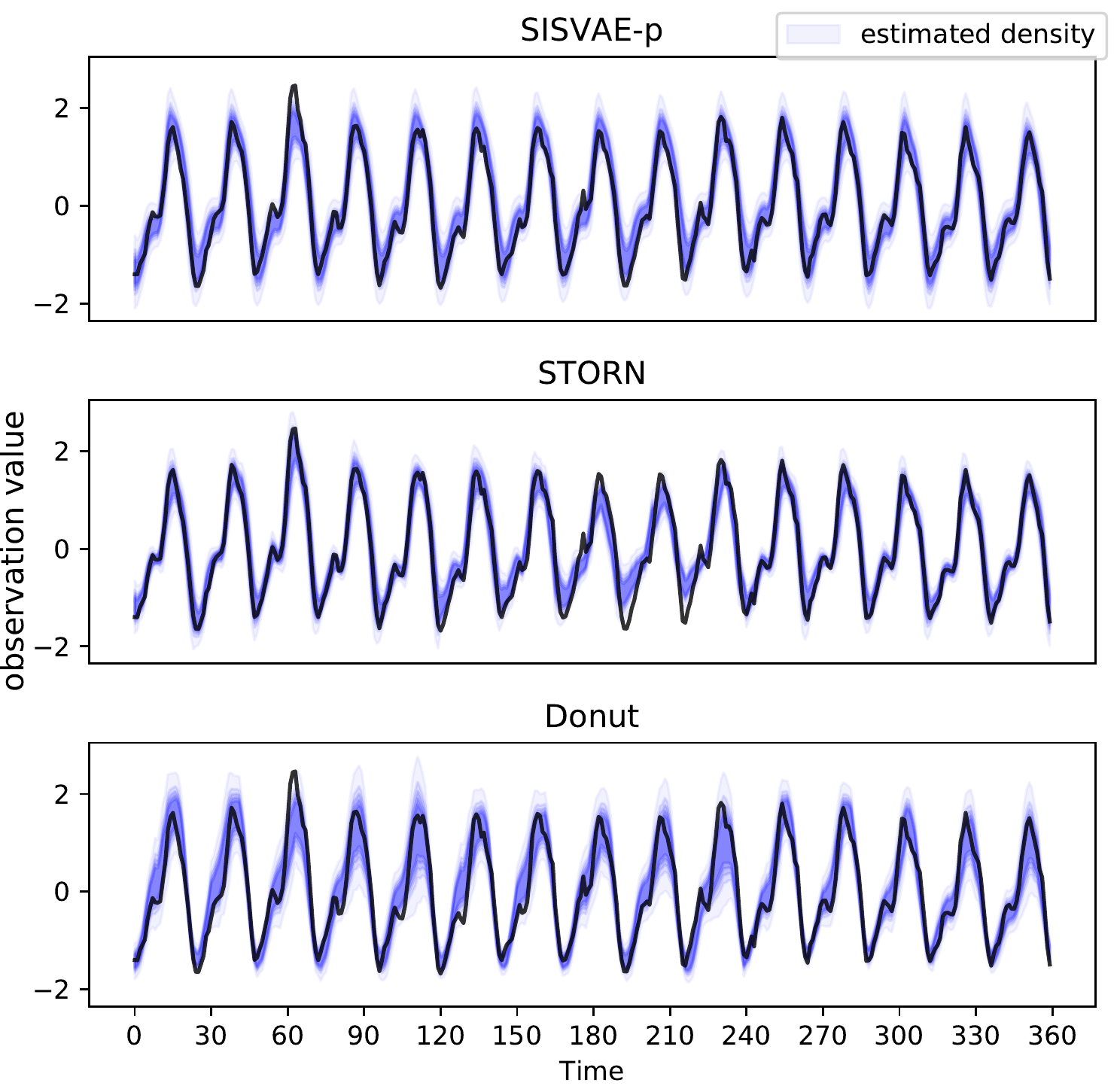}\label{fig:seq1_density}}
    \subfigure[anomaly detection score]
    {\includegraphics[width=0.23\textwidth]{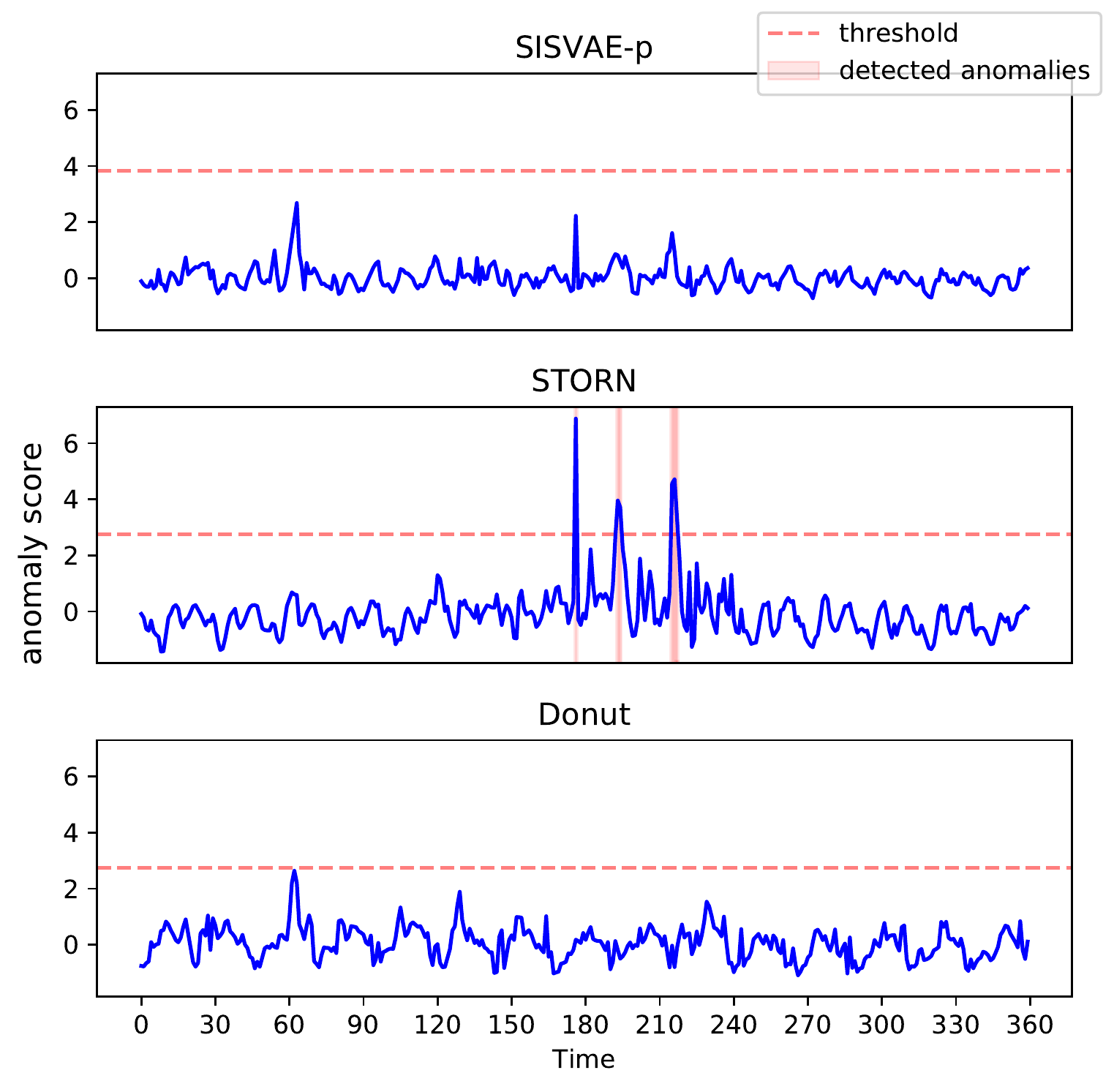}\label{fig:seq1_score}}
    \label{fig:seq1_all}
    \caption{Example for time series without anomaly. Note false alarm by STORN.}
\end{figure}

\textbf{False alarms on normal time series.}
Figure \ref{fig:seq1_true} shows a normal sequence without any anomaly. Since the training data is contaminated with anomalies, the model can be influenced by anomalies and fails to capture normal patterns such that false alarms can be be reported. Figure \ref{fig:seq1_density} illustrates the density estimation of the three models, and Fig. \ref{fig:seq1_score} shows the anomaly score generated by three models. Note STORN reports three false alarms, while SISVAE-p and Donut do not. Moreover, we find the anomaly score generated by SISVAE-p is stably low, while Donut is more fluctuating given noisy data. The results demonstrate that SISVAE-p is robust to contaminated data for unsupervised training.
\begin{figure}[tb!]
    \centering
    \subfigure[studied dimension of the 65-dim time series labeled with ground truth anomaly windows (stretched  twice compared to (b) and (c))]
    {\includegraphics[width=0.48\textwidth]
    {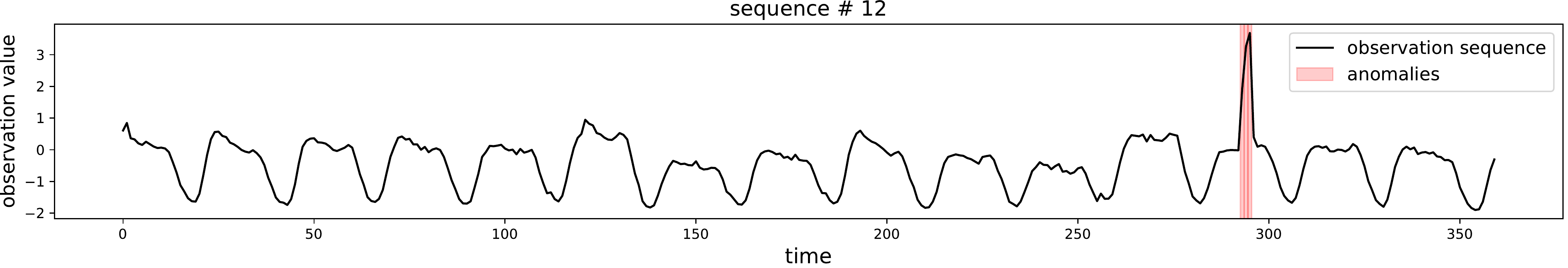}\label{fig:seq3_true}}
    \subfigure[density estimation]
    {\includegraphics[width=0.23\textwidth]{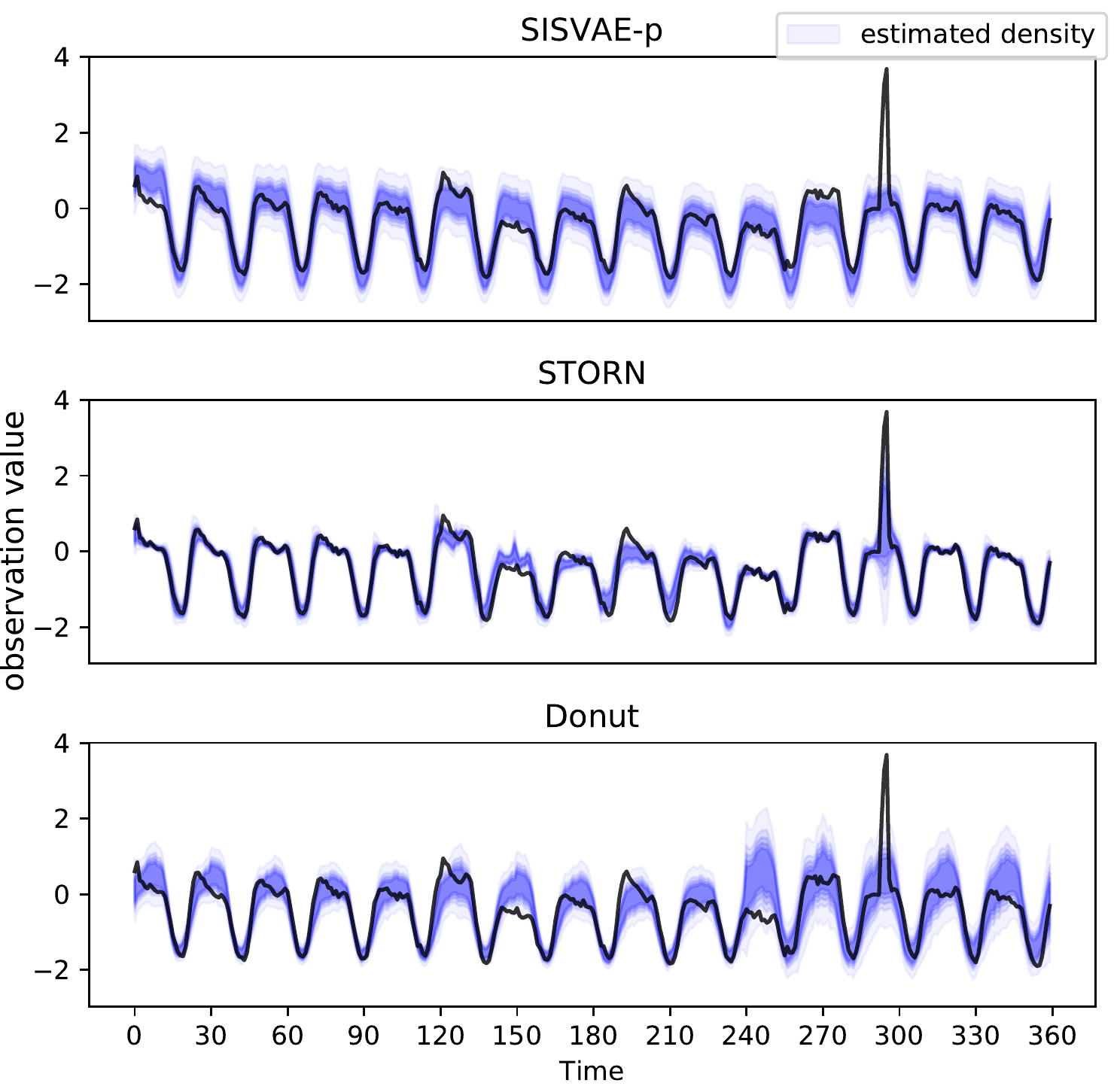}\label{fig:seq3_density}}
    \subfigure[anomaly detection score]
    {\includegraphics[width=0.23\textwidth]{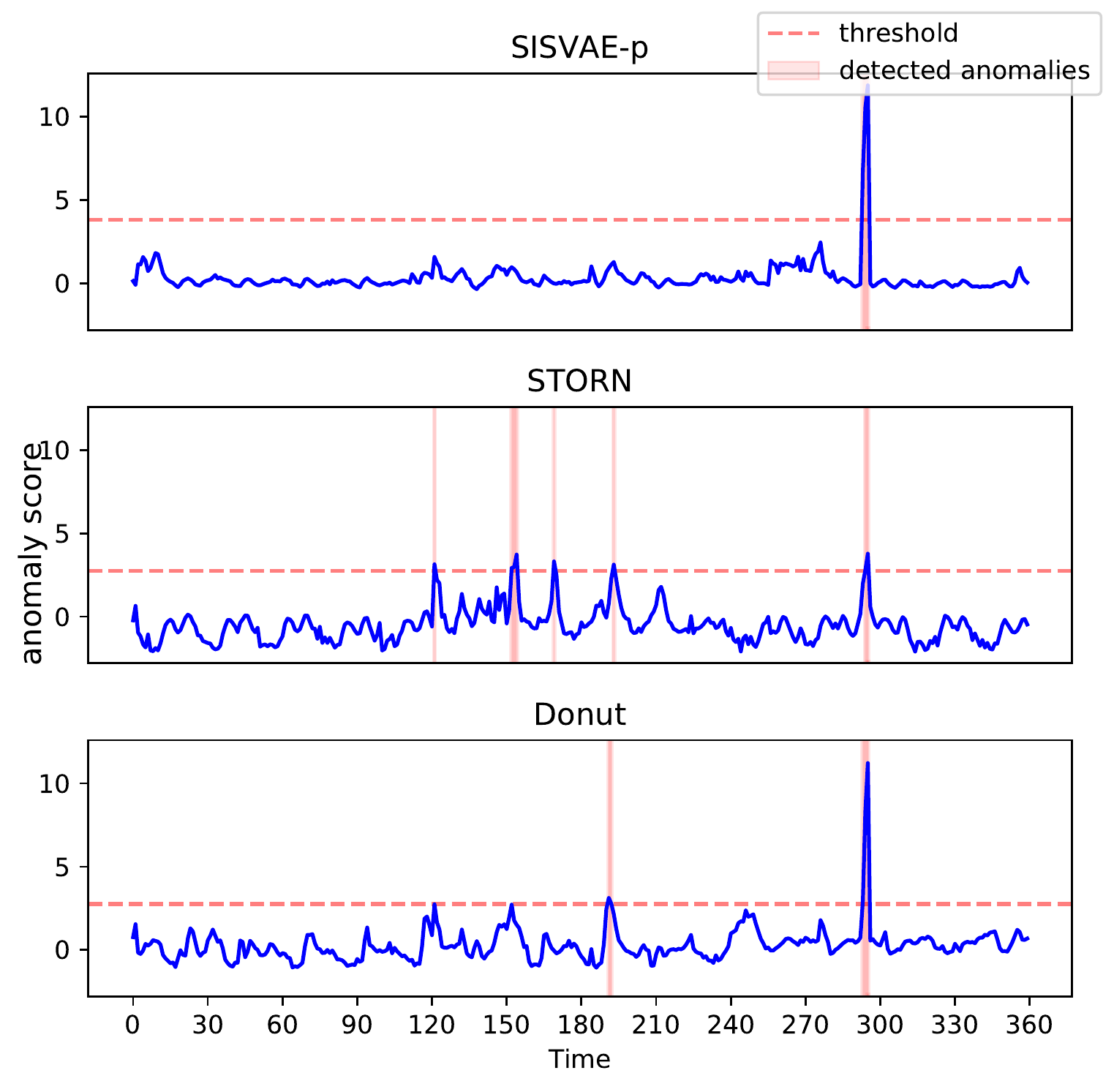}\label{fig:seq3_score}}
    \label{fig:seq3_all}
    \caption{Example for time series sample with point-level anomaly. Note the false alarms by STORN and Donut.}
\end{figure}

\textbf{Detection of point-level anomalies.} Figure \ref{fig:seq3_true} includes a sequence from Yahoo-A1 dataset with anomaly at around time point 300. This type of anomaly is often called \textit{additive outlier}~\cite{burridge2006additive} or \textit{extreme value}~\cite{siffer2017anomaly}. As shown in Fig. \ref{fig:seq3_density}, density estimation of STORN and Donut are corrupted by the point anomaly, such that they place much probability density to the anomaly point. In contrast, SISVAE-p successfully recovers the true underlying density. Figure \ref{fig:seq3_score} shows the anomaly score generated by the three models, and we can see that all the models identify the anomaly point. However, STORN and Donut report some false alarms. Moreover, the anomaly score of SISVAE-p is more smooth than those of other models, suggesting that SISVAE-p is more robust to the moderate fluctuation of normal data. Overall, SISVAE is capable of detecting point-level anomalies robustly.

\begin{figure}[tb!]
    \centering
    \subfigure[studied dimension of the 65-dim time series labeled with ground truth anomaly windows (stretched  twice compared to (b) and (c))]
    {\includegraphics[width=0.48\textwidth]
    {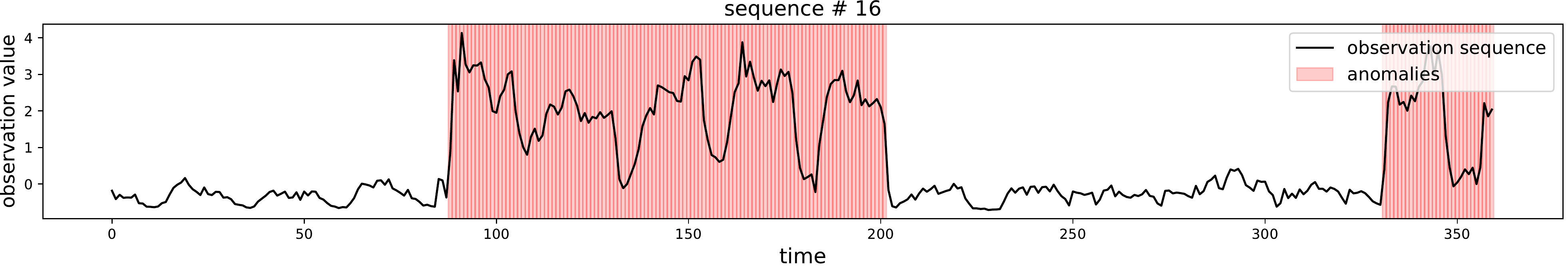}\label{fig:seq2_true}}
    \subfigure[density estimation]
    {\includegraphics[width=0.23\textwidth]{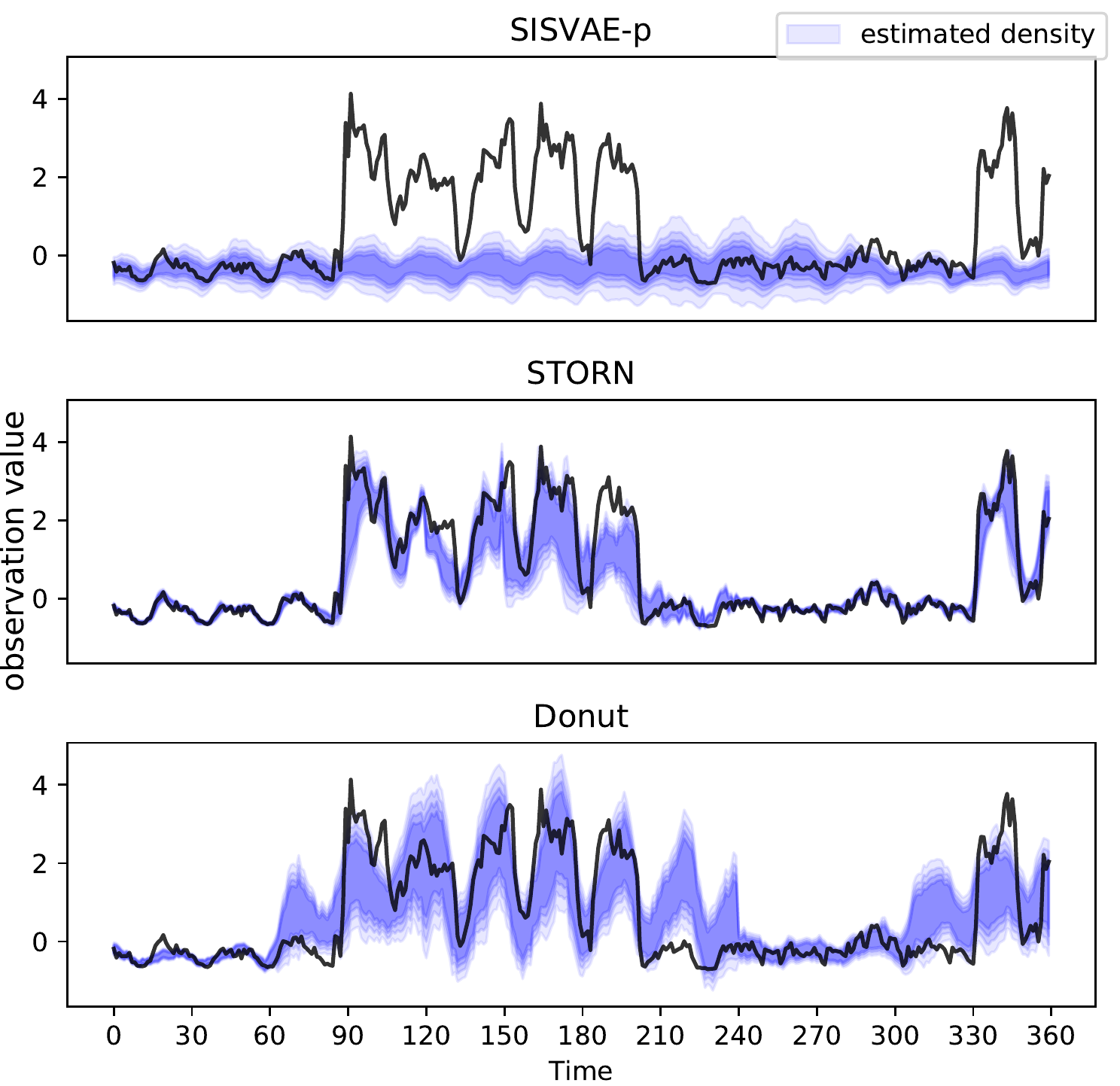}\label{fig:seq2_density}}
    \subfigure[anomaly detection score]
    {\includegraphics[width=0.23\textwidth]{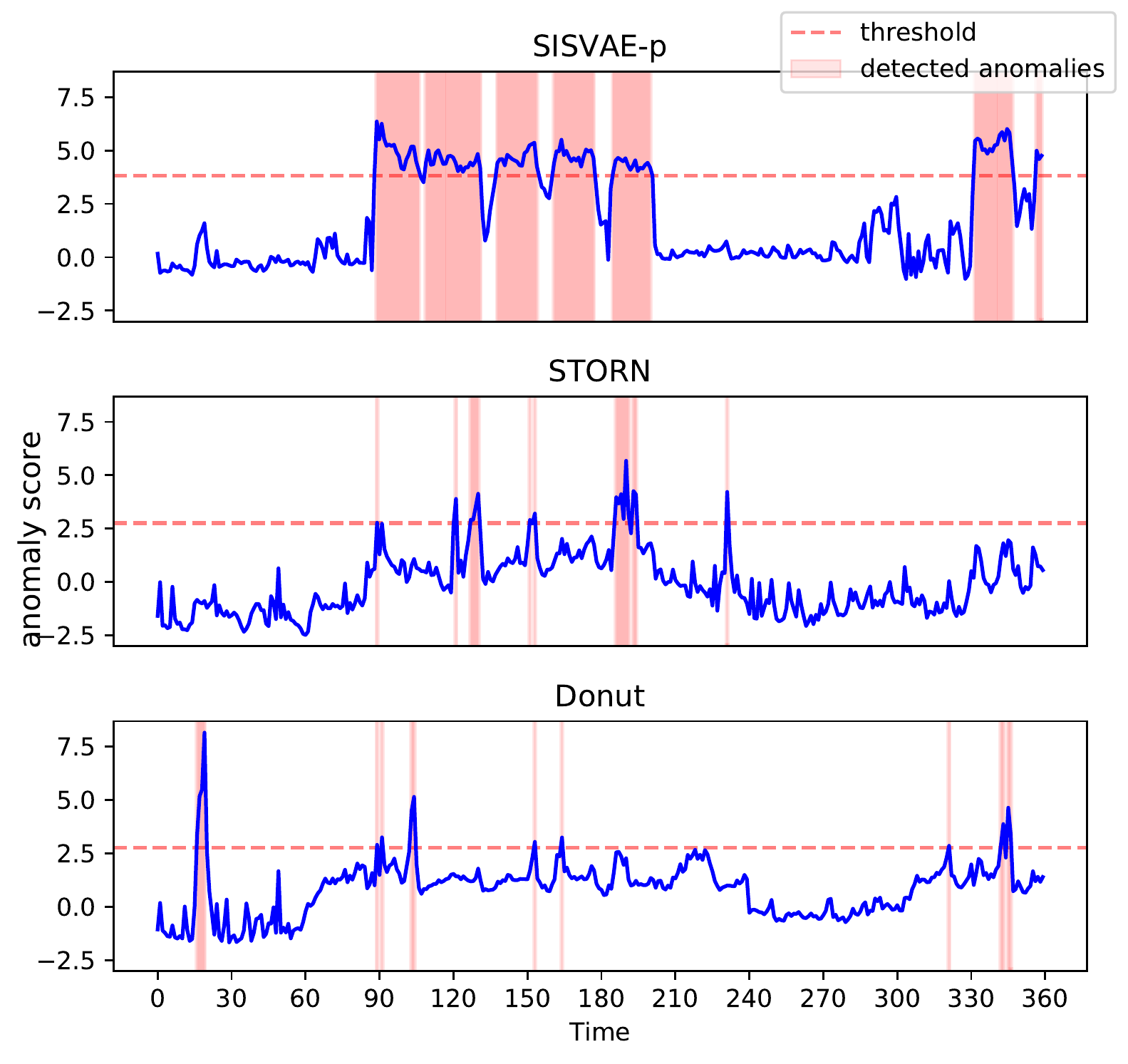}\label{fig:seq2_score}}
    \label{fig:seq2_all}
    \caption{Example for time series sample with window-sized anomaly. Note the missed anomaly regions by STORN and Donut.}
\end{figure}

\textbf{Detection of sub-sequence level anomalies.}
In some cases, the anomalies is in the presence of sub-sequences, such as anomaly cardiac cycle in ECG data \cite{chuah2007ecg}. Figure \ref{fig:seq2_true} shows a sequence with two anomaly sub-sequences, whereby the red shaded region covers the true anomaly. Figure \ref{fig:seq2_density} shows the density estimation, whereby Donut and STORN are notably influenced by the anomaly sub-sequence. In contrast,  SISVAE-p estimates a smooth probability density over time, which is more likely to be the true density. Figure \ref{fig:seq2_score} shows SISVAE-p successfully detects most of the anomaly sub-sequences, while STORN and Donut suffer from false negative and false positives. The results demonstrate the effectiveness of the proposed smoothness prior regularizer, helping recover the true density and improve detection performance.

\begin{figure}[tb!]
    \centering
    \includegraphics[width=0.48\textwidth]{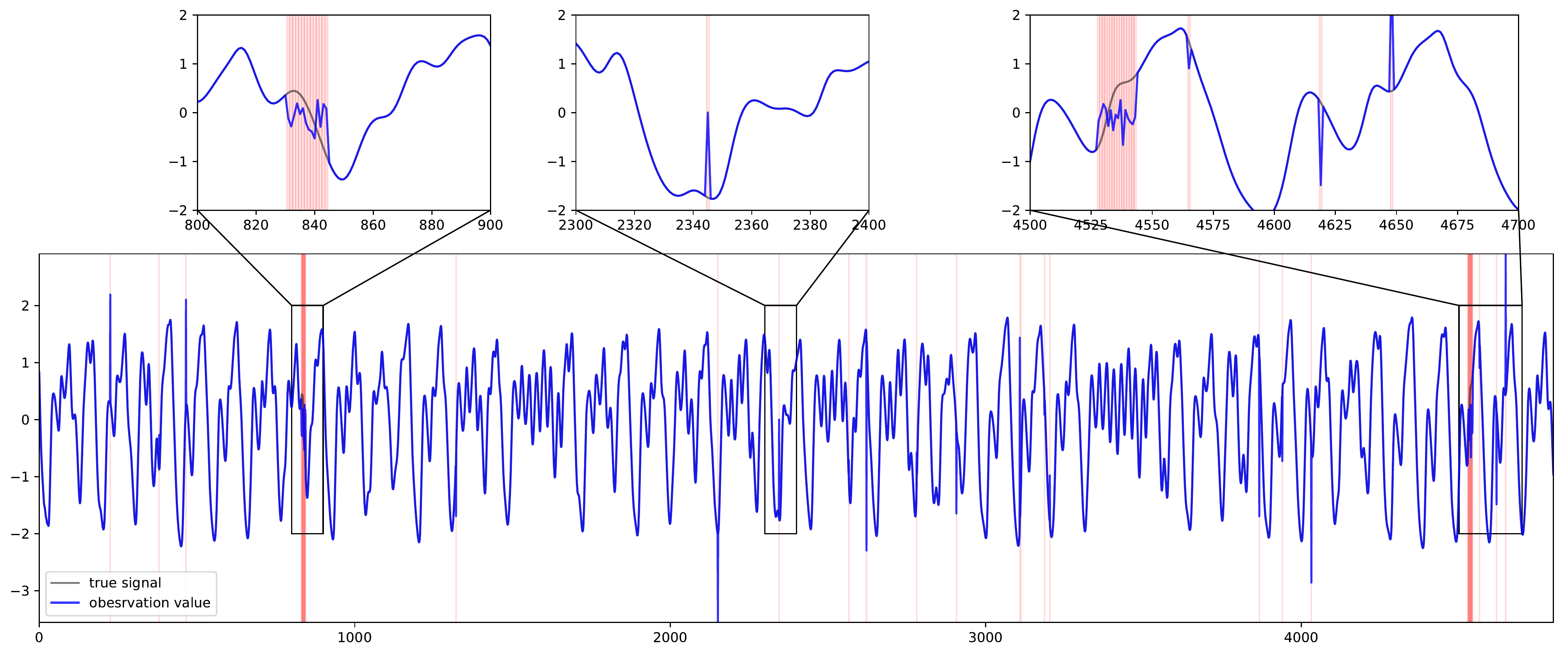}
    \caption{Generated Mackey-Glass time series data, and inserted anomalies.}
    \label{fig:mackey_generate}
\end{figure}
\begin{figure}[tb!]
    \centering
    \subfigure[Precision and recall curve]
    {\includegraphics[width=0.24\textwidth]{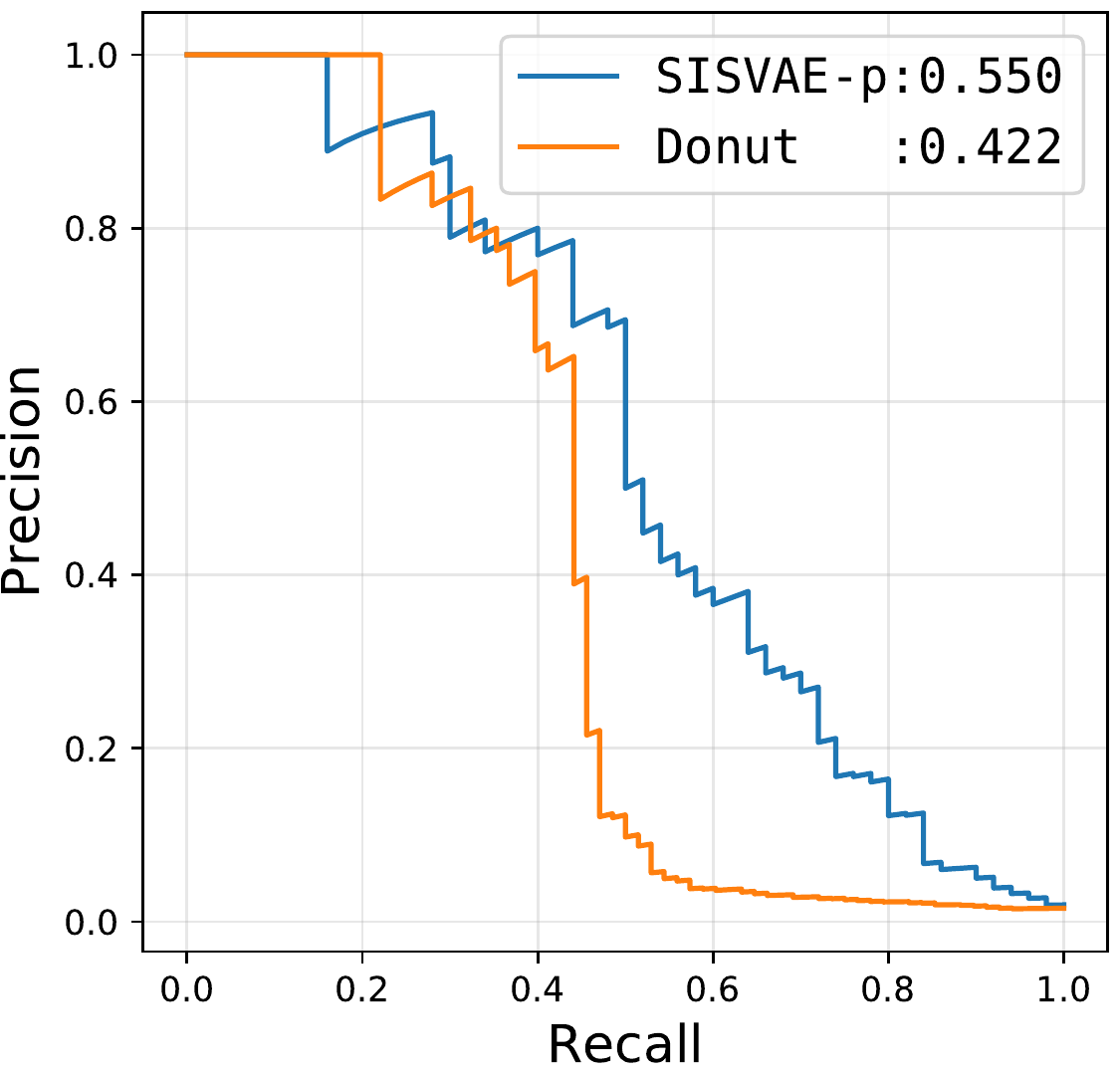}\label{fig:prc_mackey}}
    \subfigure[Receiver operating characteristic]
    {\includegraphics[width=0.24\textwidth]{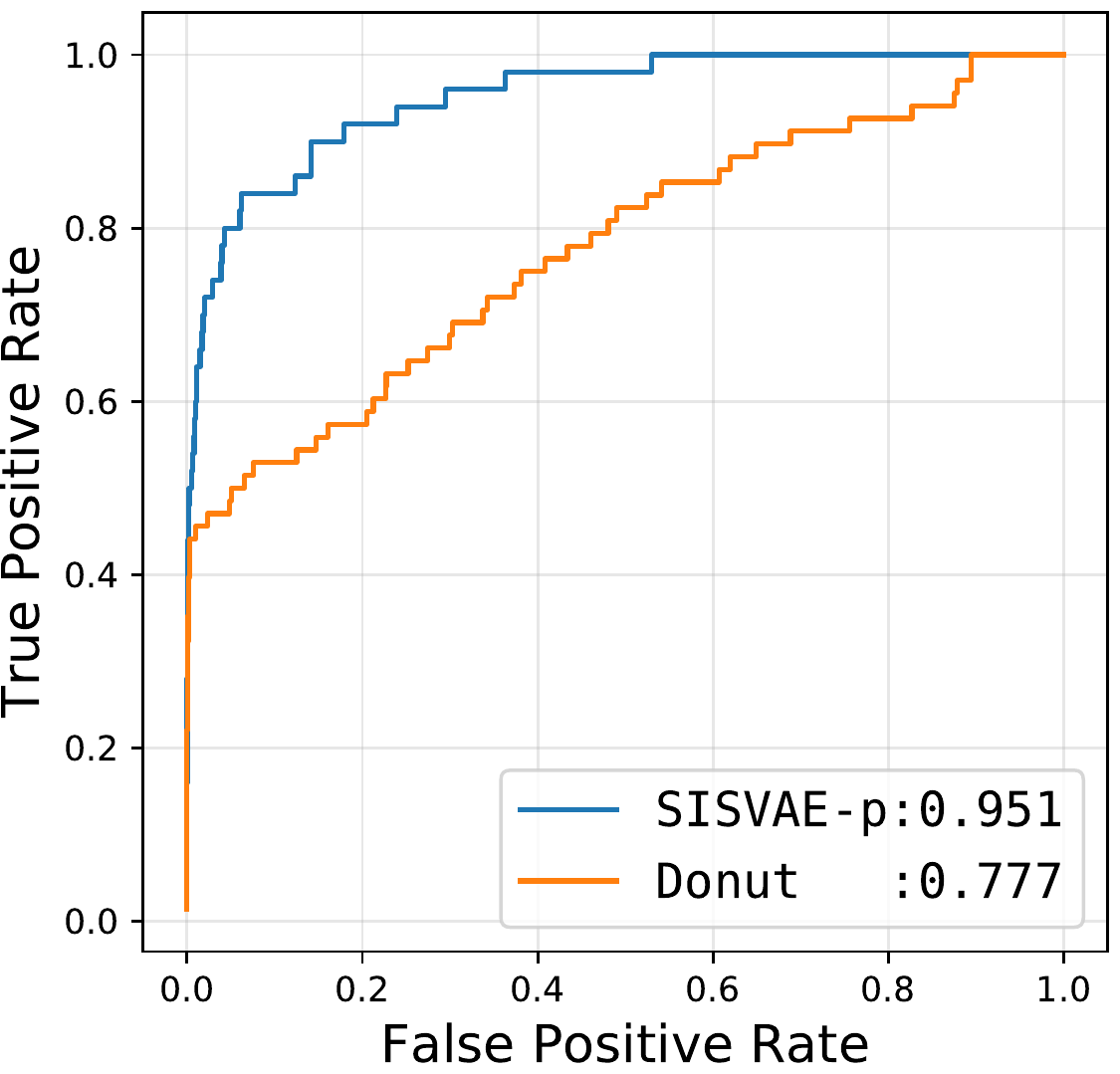}\label{fig:prc_mackey}}
    \caption{Comparison of anomaly detection performance between SISVAE-p and Donut on Mackey-Glass time series dataset.}
    \label{fig:curve_mackey}
\end{figure}
\subsection{Further Comparison with Donut}
\textbf{Simulation study on uni-variate time series.}  
As discussed in Section \ref{sec:setting}, Donut is designated for uni-variate time series. In order to investigate the detection performance in one-dimensional case, we conduct simulation study on data generated from Mackey-Glass equation~\cite{mackey1977oscillation}, which serves a benchmark for nonlinear models since the true underlying function is nonlinear~\cite{kantz2004nonlinear}. The time series is generated as:
\begin{equation}
    \frac{dx(t)}{dt} =  \frac{\alpha x(t-\tau) }{1+x^\beta(t-\tau)}-\gamma x, \quad \gamma,\beta,n > 0,
\end{equation}
with $\gamma=0.1, \beta=10, \alpha=0.2$.
We run the Mackey-Glass equation 5000 iterations, then we collect an uni-variate time series with length 5000. Two types of anomalies are added:
\begin{itemize}
    \item \textbf{Point level anomaly:} we insert point anomalies at $0.3\%$ time step randomly, each point anomaly is a combination of: i) Poisson noise with $\lambda=1$; ii) Gaussian noise with zero mean and unit variance; iii) 0.2 fixed bias. 
    \item \textbf{Sub-sequence level anomaly:} two sub-sequences anomalies are inserted, starting at random time step, with random length sampled from even distribution of range $10\sim 28$. For each anomaly sub-sequence, true signal is replaced by a sequence sampled from a Gaussian process with RBF kernel of unit variance, and 0.3 length-scale.
\end{itemize}

The generated Mackey-Glass time series is shown in Fig. \ref{fig:mackey_generate}. To illustrate the generated anomalies, we zoom in three fragments out of the whole time series.

In this experiment, we find Donut shows better performance when $\text{z\_dim}$ is lower. We enumerate Donut's performance on $\text{z\_dim}=\{2\dots 10\}$, and we find Donut performs best when $\text{z\_dim}=5$. For SISVAE-p, we find the performance is relatively stable when we change $\text{z\_dim}$. Hence we use the same setup as discussed in Section \ref{sec:setting}. The ROC and PRC are shown in Fig. \ref{fig:curve_mackey}. In conclusion, benefiting from the sequential network structure of SISVAE, and the proposed smoothness prior regularizer, SISVAE-p exhibits better anomaly detection performance for uni-variate time series.
\section{Conclusion}
\label{sec:conclusion}
We have presented a deep generative model based method i.e. SISVAE (and the variants) for robust estimation and anomaly detection for multi-dimensional correlated time series. To capture `normal' patterns from anomaly-contaminated time series, we propose a novel variational smoothness regularizer, which provides robustness by placing penalty on non-smooth output of the generative model. Furthermore, we discuss two decision criteria for anomaly detection: reconstruction probability and reconstruction error. Systemic experiments are conducted on both synthetic and real-world datasets, which demonstrate the proposed method outperforms state-of-the-art unsupervised anomaly detection models for multi-dimensional time series. For future work, we are interested in exploring heavy-tailed distribution such as Student-$t$ distribution as the output of the generative model.

\section*{Acknowledgments}
This research was partially supported by China Major State Research Development Program (2018YFC0830400, 2018AAA0100704) and NSFC (61972250, U19B2035).

\ifCLASSOPTIONcompsoc
   \section*{Acknowledgments}
   This research was partially supported by NSFC (61602176, 61672231), NSFC-Zhejiang Joint Fund for the Integration of Industrialization and Informatization U1609220.
\else

\fi
\ifCLASSOPTIONcaptionsoff
  \newpage
\fi
\bibliographystyle{IEEEtran}
\bibliography{main}

\begin{thebibliography}{10}
\providecommand{\url}[1]{#1}
\csname url@samestyle\endcsname
\providecommand{\newblock}{\relax}
\providecommand{\bibinfo}[2]{#2}
\providecommand{\BIBentrySTDinterwordspacing}{\spaceskip=0pt\relax}
\providecommand{\BIBentryALTinterwordstretchfactor}{4}
\providecommand{\BIBentryALTinterwordspacing}{\spaceskip=\fontdimen2\font plus
\BIBentryALTinterwordstretchfactor\fontdimen3\font minus
  \fontdimen4\font\relax}
\providecommand{\BIBforeignlanguage}[2]{{%
\expandafter\ifx\csname l@#1\endcsname\relax
\typeout{** WARNING: IEEEtran.bst: No hyphenation pattern has been}%
\typeout{** loaded for the language `#1'. Using the pattern for}%
\typeout{** the default language instead.}%
\else
\language=\csname l@#1\endcsname
\fi
#2}}
\providecommand{\BIBdecl}{\relax}
\BIBdecl

\bibitem{li2009temporal}
X.~Li, Z.~Li, J.~Han, and J.-G. Lee, ``Temporal outlier detection in vehicle
  traffic data,'' in \emph{IEEE International Conference on Data
  Engineering}.\hskip 1em plus 0.5em minus 0.4em\relax IEEE, 2009, pp.
  1319--1322.

\bibitem{hauskrecht2013outlier}
M.~Hauskrecht, I.~Batal, M.~Valko, S.~Visweswaran, G.~F. Cooper, and
  G.~Clermont, ``Outlier detection for patient monitoring and alerting,''
  \emph{Journal of biomedical informatics}, vol.~46, no.~1, pp. 47--55, 2013.

\bibitem{xu2018unsupervised}
H.~Xu, W.~Chen, N.~Zhao, Z.~Li, J.~Bu, Z.~Li, Y.~Liu, Y.~Zhao, D.~Pei, Y.~Feng
  \emph{et~al.}, ``Unsupervised anomaly detection via variational auto-encoder
  for seasonal kpis in web applications,'' in \emph{Proceedings of the 2018
  World Wide Web Conference on World Wide Web}.\hskip 1em plus 0.5em minus
  0.4em\relax International World Wide Web Conferences Steering Committee,
  2018, pp. 187--196.

\bibitem{jyothsna2011review}
V.~Jyothsna, V.~R. Prasad, and K.~M. Prasad, ``A review of anomaly based
  intrusion detection systems,'' \emph{International Journal of Computer
  Applications}, vol.~28, no.~7, pp. 26--35, 2011.

\bibitem{hill2010anomaly}
D.~J. Hill and B.~S. Minsker, ``Anomaly detection in streaming environmental
  sensor data: A data-driven modeling approach,'' \emph{Environmental Modelling
  \& Software}, vol.~25, no.~9, pp. 1014--1022, 2010.

\bibitem{YanIETEL11}
J.~Yan, C.~Tian, J.~Huang, and F.~Albertao, ``Incremental dictionary learning
  for fault detection with applications to oil pipeline leakage detection,''
  \emph{Electronics Letters}, vol.~47, no.~21, pp. 1198--1199, 2011.

\bibitem{kim2018zero}
J.-Y. Kim, S.-J. Bu, and S.-B. Cho, ``Zero-day malware detection using
  transferred generative adversarial networks based on deep autoencoders,''
  \emph{Information Sciences}, vol. 460, pp. 83--102, 2018.

\bibitem{chandola2009anomaly}
V.~Chandola, A.~Banerjee, and V.~Kumar, ``Anomaly detection: A survey,''
  \emph{ACM computing surveys (CSUR)}, vol.~41, no.~3, p.~15, 2009.

\bibitem{aggarwal2015outlier}
C.~C. Aggarwal, ``Outlier analysis,'' in \emph{Data mining}.\hskip 1em plus
  0.5em minus 0.4em\relax Springer, 2015, pp. 237--263.

\bibitem{gupta2014outlier}
M.~Gupta, J.~Gao, C.~C. Aggarwal, and J.~Han, ``Outlier detection for temporal
  data: A survey,'' \emph{IEEE Transactions on Knowledge and Data Engineering},
  vol.~26, no.~9, pp. 2250--2267, 2014.

\bibitem{chandola2012anomaly}
V.~Chandola, A.~Banerjee, and V.~Kumar, ``Anomaly detection for discrete
  sequences: A survey,'' \emph{IEEE Transactions on Knowledge and Data
  Engineering}, vol.~24, no.~5, pp. 823--839, 2012.

\bibitem{tsay2000outliers}
R.~S. Tsay, D.~Pe{\~n}a, and A.~E. Pankratz, ``Outliers in multivariate time
  series,'' \emph{Biometrika}, vol.~87, no.~4, pp. 789--804, 2000.

\bibitem{basu2007automatic}
S.~Basu and M.~Meckesheimer, ``Automatic outlier detection for time series: an
  application to sensor data,'' \emph{Knowledge and Information Systems},
  vol.~11, no.~2, pp. 137--154, 2007.

\bibitem{erfani2016high}
S.~M. Erfani, S.~Rajasegarar, S.~Karunasekera, and C.~Leckie,
  ``High-dimensional and large-scale anomaly detection using a linear one-class
  svm with deep learning,'' \emph{Pattern Recognition}, vol.~58, pp. 121--134,
  2016.

\bibitem{laxhammar2009anomaly}
R.~Laxhammar, G.~Falkman, and E.~Sviestins, ``Anomaly detection in sea
  traffic-a comparison of the gaussian mixture model and the kernel density
  estimator,'' in \emph{Information Fusion, 2009. FUSION'09. 12th International
  Conference on}.\hskip 1em plus 0.5em minus 0.4em\relax IEEE, 2009, pp.
  756--763.

\bibitem{chung2015recurrent}
J.~Chung, K.~Kastner, L.~Dinh, K.~Goel, A.~C. Courville, and Y.~Bengio, ``A
  recurrent latent variable model for sequential data,'' in \emph{Advances in
  neural information processing systems}, 2015, pp. 2980--2988.

\bibitem{locatello2019challenging}
F.~Locatello, S.~Bauer, M.~Lucic, G.~Raetsch, S.~Gelly, B.~Sch{\"o}lkopf, and
  O.~Bachem, ``Challenging common assumptions in the unsupervised learning of
  disentangled representations,'' in \emph{International Conference on Machine
  Learning}, 2019, pp. 4114--4124.

\bibitem{zhao2018bias}
S.~Zhao, H.~Ren, A.~Yuan, J.~Song, N.~Goodman, and S.~Ermon, ``Bias and
  generalization in deep generative models: An empirical study,'' in
  \emph{Advances in Neural Information Processing Systems}, 2018, pp.
  10\,792--10\,801.

\bibitem{shang2015enhancing}
C.~Shang, X.~Huang, J.~A. Suykens, and D.~Huang, ``Enhancing dynamic soft
  sensors based on dpls: A temporal smoothness regularization approach,''
  \emph{Journal of Process Control}, vol.~28, pp. 17--26, 2015.

\bibitem{cai2015fast}
Y.~Cai, H.~Tong, W.~Fan, and P.~Ji, ``Fast mining of a network of coevolving
  time series,'' in \emph{Proceedings of the 2015 SIAM International Conference
  on Data Mining}.\hskip 1em plus 0.5em minus 0.4em\relax SIAM, 2015, pp.
  298--306.

\bibitem{enders2008applied}
W.~Enders, \emph{Applied econometric time series}.\hskip 1em plus 0.5em minus
  0.4em\relax John Wiley \& Sons, 2008.

\bibitem{zhou2018non}
Y.~Zhou, H.~Zou, R.~Arghandeh, W.~Gu, and C.~J. Spanos, ``Non-parametric
  outliers detection in multiple time series a case study: Power grid data
  analysis.'' in \emph{AAAI}, 2018.

\bibitem{cai2015facets}
Y.~Cai, H.~Tong, W.~Fan, P.~Ji, and Q.~He, ``Facets: Fast comprehensive mining
  of coevolving high-order time series,'' in \emph{Proceedings of the 21th ACM
  SIGKDD International Conference on Knowledge Discovery and Data
  Mining}.\hskip 1em plus 0.5em minus 0.4em\relax ACM, 2015, pp. 79--88.

\bibitem{gornitz2015hidden}
N.~G{\"o}rnitz, M.~Braun, and M.~Kloft, ``Hidden markov anomaly detection,'' in
  \emph{International Conference on Machine Learning}, 2015, pp. 1833--1842.

\bibitem{li2009dynammo}
L.~Li, J.~McCann, N.~S. Pollard, and C.~Faloutsos, ``Dynammo: Mining and
  summarization of coevolving sequences with missing values,'' in
  \emph{Proceedings of the 15th ACM SIGKDD international conference on
  Knowledge discovery and data mining}.\hskip 1em plus 0.5em minus 0.4em\relax
  ACM, 2009, pp. 507--516.

\bibitem{xiong2011direct}
L.~Xiong, X.~Chen, and J.~Schneider, ``Direct robust matrix factorizatoin for
  anomaly detection,'' in \emph{Data Mining (ICDM), 2011 IEEE 11th
  International Conference on}.\hskip 1em plus 0.5em minus 0.4em\relax IEEE,
  2011, pp. 844--853.

\bibitem{kingma2013auto}
D.~P. Kingma and M.~Welling, ``Auto-encoding variational bayes,'' \emph{arXiv
  preprint arXiv:1312.6114}, 2013.

\bibitem{an2015variational}
J.~An and S.~Cho, ``Variational autoencoder based anomaly detection using
  reconstruction probability,'' \emph{Special Lecture on IE}, vol.~2, pp.
  1--18, 2015.

\bibitem{Elman1990RNN}
J.~L. Elman, ``Finding structure in time,'' \emph{Cognitive Science}, vol.~14,
  pp. 179--211, 1990.

\bibitem{solch2016variational}
M.~S{\"o}lch, J.~Bayer, M.~Ludersdorfer, and P.~van~der Smagt, ``Variational
  inference for on-line anomaly detection in high-dimensional time series,''
  \emph{arXiv preprint arXiv:1602.07109}, 2016.

\bibitem{kitagawa2012smoothness}
G.~Kitagawa and W.~Gersch, \emph{Smoothness priors analysis of time
  series}.\hskip 1em plus 0.5em minus 0.4em\relax Springer Science \& Business
  Media, 2012, vol. 116.

\bibitem{kingma2014stochastic}
D.~P. Kingma and M.~Welling, ``Stochastic gradient vb and the variational
  auto-encoder,'' in \emph{Second International Conference on Learning
  Representations, ICLR}, 2014.

\bibitem{fabius2014variational}
O.~Fabius and J.~R. van Amersfoort, ``Variational recurrent auto-encoders,''
  \emph{arXiv preprint arXiv:1412.6581}, 2014.

\bibitem{bayer2014learning}
J.~Bayer and C.~Osendorfer, ``Learning stochastic recurrent networks,''
  \emph{arXiv preprint arXiv:1411.7610}, 2014.

\bibitem{fraccaro2016sequential}
M.~Fraccaro, S.~K. S{\o}nderby, U.~Paquet, and O.~Winther, ``Sequential neural
  models with stochastic layers,'' in \emph{Advances in neural information
  processing systems}, 2016, pp. 2199--2207.

\bibitem{cho2014learning}
K.~Cho, B.~Van~Merri{\"e}nboer, C.~Gulcehre, D.~Bahdanau, F.~Bougares,
  H.~Schwenk, and Y.~Bengio, ``Learning phrase representations using rnn
  encoder-decoder for statistical machine translation,'' \emph{arXiv preprint
  arXiv:1406.1078}, 2014.

\bibitem{higgins2017beta}
I.~Higgins, L.~Matthey, A.~Pal, C.~Burgess, X.~Glorot, M.~Botvinick,
  S.~Mohamed, and A.~Lerchner, ``beta-vae: Learning basic visual concepts with
  a constrained variational framework,'' in \emph{International Conference on
  Learning Representations}, 2017.

\bibitem{werbos1990backpropagation}
P.~J. Werbos, ``Backpropagation through time: what it does and how to do it,''
  \emph{Proceedings of the IEEE}, vol.~78, no.~10, pp. 1550--1560, 1990.

\bibitem{pascanu2013difficulty}
R.~Pascanu, T.~Mikolov, and Y.~Bengio, ``On the difficulty of training
  recurrent neural networks,'' in \emph{International Conference on Machine
  Learning}, 2013, pp. 1310--1318.

\bibitem{zhou2017anomaly}
C.~Zhou and R.~C. Paffenroth, ``Anomaly detection with robust deep
  autoencoders,'' in \emph{Proceedings of the 23rd ACM SIGKDD International
  Conference on Knowledge Discovery and Data Mining}.\hskip 1em plus 0.5em
  minus 0.4em\relax ACM, 2017, pp. 665--674.

\bibitem{doucet2001introduction}
A.~Doucet, N.~De~Freitas, and N.~Gordon, ``An introduction to sequential monte
  carlo methods,'' in \emph{Sequential Monte Carlo methods in practice}.\hskip
  1em plus 0.5em minus 0.4em\relax Springer, 2001, pp. 3--14.

\bibitem{doucet2000sequential}
A.~Doucet, S.~Godsill, and C.~Andrieu, ``On sequential monte carlo sampling
  methods for bayesian filtering,'' \emph{Statistics and computing}, vol.~10,
  no.~3, pp. 197--208, 2000.

\bibitem{liu1998sequential}
J.~S. Liu and R.~Chen, ``Sequential monte carlo methods for dynamic systems,''
  \emph{Journal of the American statistical association}, vol.~93, no. 443, pp.
  1032--1044, 1998.

\bibitem{davis2006relationship}
J.~Davis and M.~Goadrich, ``The relationship between precision-recall and roc
  curves,'' in \emph{Proceedings of the 23rd international conference on
  Machine learning}.\hskip 1em plus 0.5em minus 0.4em\relax ACM, 2006, pp.
  233--240.

\bibitem{saito2015precision}
T.~Saito and M.~Rehmsmeier, ``The precision-recall plot is more informative
  than the roc plot when evaluating binary classifiers on imbalanced
  datasets,'' \emph{PloS one}, vol.~10, no.~3, 2015.

\bibitem{galeano2006outlier}
P.~Galeano, D.~Pe{\~n}a, and R.~S. Tsay, ``Outlier detection in multivariate
  time series by projection pursuit,'' \emph{Journal of the American
  Statistical Association}, vol. 101, no. 474, pp. 654--669, 2006.

\bibitem{hamilton1994time}
J.~D. Hamilton, \emph{Time series analysis}.\hskip 1em plus 0.5em minus
  0.4em\relax Princeton university press Princeton, NJ, 1994, vol.~2.

\bibitem{paszke2019pytorch}
A.~Paszke, S.~Gross, F.~Massa, A.~Lerer, J.~Bradbury, G.~Chanan, T.~Killeen,
  Z.~Lin, N.~Gimelshein, L.~Antiga \emph{et~al.}, ``Pytorch: An imperative
  style, high-performance deep learning library,'' in \emph{Advances in Neural
  Information Processing Systems}, 2019, pp. 8024--8035.

\bibitem{kingma2014adam}
D.~P. Kingma and J.~Ba, ``Adam: A method for stochastic optimization,''
  \emph{arXiv preprint arXiv:1412.6980}, 2014.

\bibitem{alvarez2012kernels}
M.~A. Alvarez, L.~Rosasco, N.~D. Lawrence \emph{et~al.}, ``Kernels for
  vector-valued functions: A review,'' \emph{Foundations and
  Trends{\textregistered} in Machine Learning}, vol.~4, no.~3, pp. 195--266,
  2012.

\bibitem{gpy2014}
{GPy}, ``{GPy}: A gaussian process framework in python,''
  \url{http://github.com/SheffieldML/GPy}, since 2012.

\bibitem{stewart2016open}
E.~Stewart, A.~Liao, and C.~Roberts, ``Open $\mu$pmu: A real world reference
  distribution micro-phasor measurement unit data set for research and
  application development,'' 2016.

\bibitem{baeza2011modern}
R.~Baeza-Yates, B.~d. A.~N. Ribeiro \emph{et~al.}, \emph{Modern information
  retrieval}.\hskip 1em plus 0.5em minus 0.4em\relax New York: ACM Press;
  Harlow, England: Addison-Wesley,, 2011.

\bibitem{burridge2006additive}
P.~Burridge and A.~Robert~Taylor, ``Additive outlier detection via
  extreme-value theory,'' \emph{Journal of Time Series Analysis}, vol.~27,
  no.~5, pp. 685--701, 2006.

\bibitem{siffer2017anomaly}
A.~Siffer, P.-A. Fouque, A.~Termier, and C.~Largouet, ``Anomaly detection in
  streams with extreme value theory,'' in \emph{Proceedings of the 23rd ACM
  SIGKDD International Conference on Knowledge Discovery and Data
  Mining}.\hskip 1em plus 0.5em minus 0.4em\relax ACM, 2017, pp. 1067--1075.

\bibitem{chuah2007ecg}
M.~C. Chuah and F.~Fu, ``Ecg anomaly detection via time series analysis,'' in
  \emph{International Symposium on Parallel and Distributed Processing and
  Applications}.\hskip 1em plus 0.5em minus 0.4em\relax Springer, 2007, pp.
  123--135.

\bibitem{mackey1977oscillation}
M.~C. Mackey and L.~Glass, ``Oscillation and chaos in physiological control
  systems,'' \emph{Science}, vol. 197, no. 4300, pp. 287--289, 1977.

\bibitem{kantz2004nonlinear}
H.~Kantz and T.~Schreiber, \emph{Nonlinear time series analysis}.\hskip 1em
  plus 0.5em minus 0.4em\relax Cambridge university press, 2004, vol.~7.

\end{thebibliography}
\iftrue
\begin{IEEEbiography}[{\includegraphics[width=1in,height=1.25in,clip,keepaspectratio]{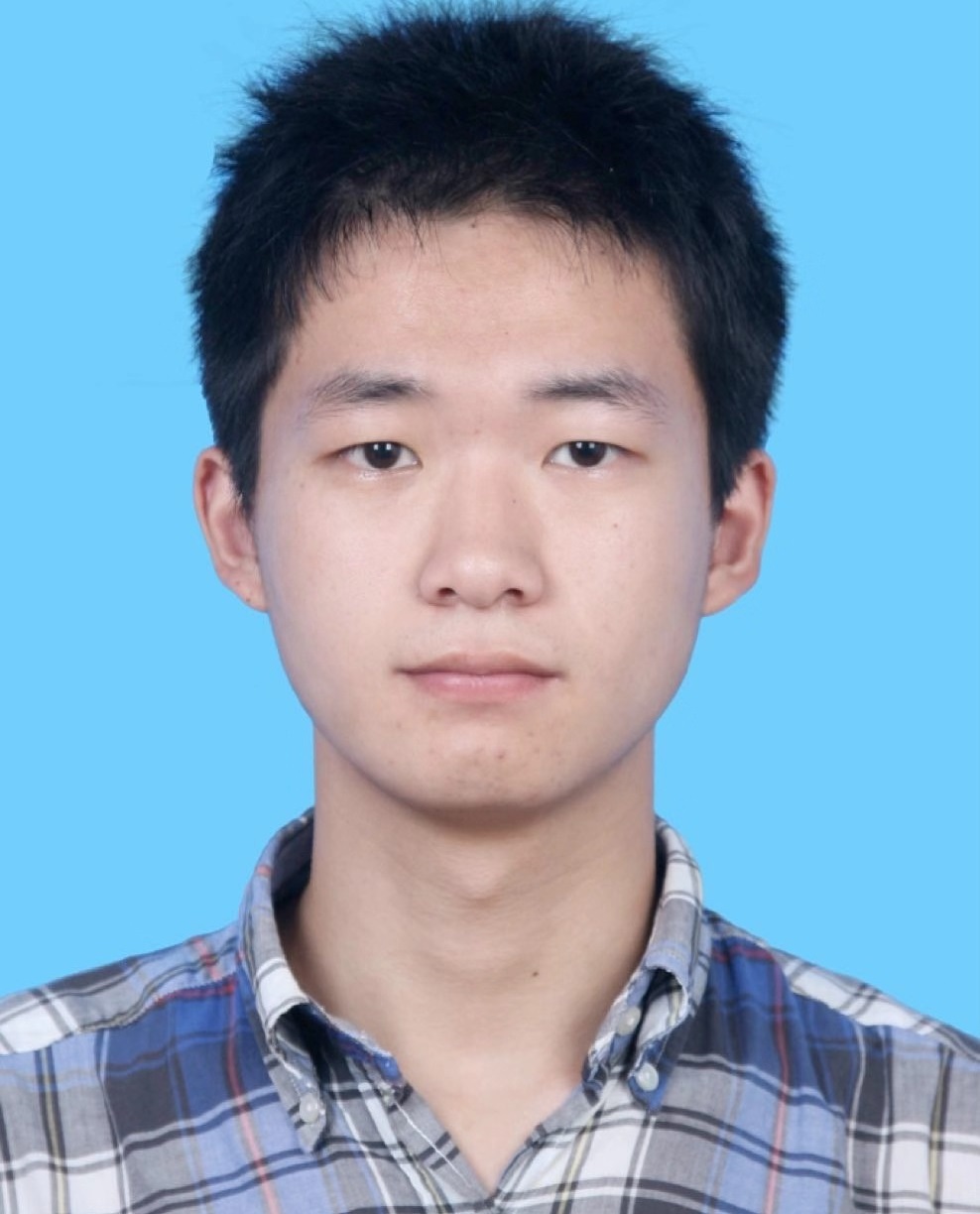}}]{Longyuan Li} received the B.E. degree in Electronic Engineering from Huazhong University of Science and Technology in 2013. He is currently working toward the Ph.D. degree in the Department of Electronic Engineering, Shanghai Jiao Tong University, Shanghai, China. His research interests include machine learning and data mining. He focuses on probabilistic models and Bayesian non-parametric models, for sequential data such as multi-dimensional time series. He is also interested in deep probabilistic graphical models and uncertainty estimation.
\end{IEEEbiography}
\begin{IEEEbiography}[{\includegraphics[width=1in,height=1.25in,clip,keepaspectratio]{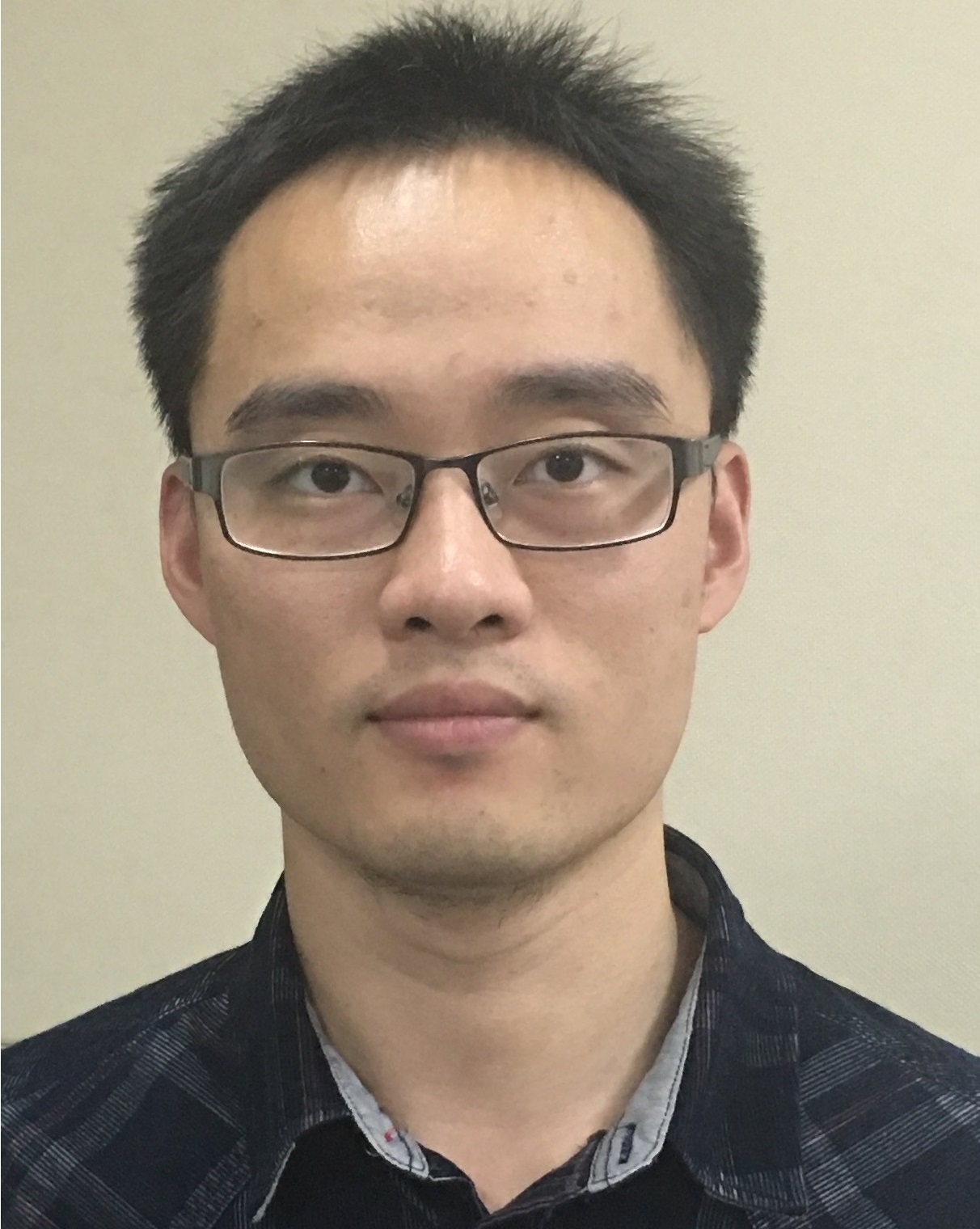}}]{Junchi Yan} (M'10) is currently an Associate Professor with Department of Computer Science and Engineering, Shanghai Jiao Tong University, Shanghai, China. Before that, he was a Research Staff Member and Principal Scientist with IBM Research -- China, where he started his career since April 2011. He obtained the Ph.D. degree in Electrical Engineering from Shanghai Jiao Tong University. His research interests are machine learning and computer vision. He serves as Area Chair for ICPR 2020, CVPR 2021, Senior PC for CIKM 2019, Associate Editor for IEEE ACCESS.
\end{IEEEbiography}
\begin{IEEEbiography}[{\includegraphics[width=1in,height=1.25in,clip,keepaspectratio]{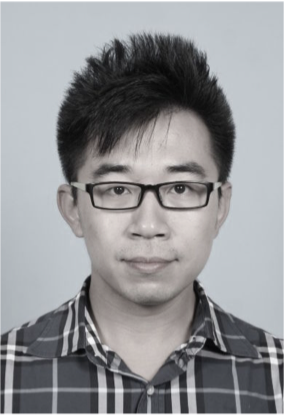}}]{Haiyang Wang} obtained B.E. degree and Ph.D. degree both in Electrical Engineering from Shanghai Jiao Tong University, Shanghai, China, in year 2013 and 2019, respectively. His research area include machine learning and data mining, especially massive spatial-temporal data mining and intelligent transportation. 
\end{IEEEbiography}
\begin{IEEEbiography}[{\includegraphics[width=1in,height=1.25in,clip,keepaspectratio]{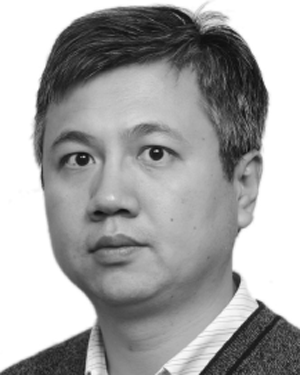}}]{Yaohui Jin} was a Technical Staff Member with Bell Labs Research China. He joined Shanghai Jiao Tong University in 2002, where he is a Professor with the State Key Laboratory of Advanced Optical Communication Systems and Networks and the Deputy Director of Network and Information Center. His research interests include civic engagement and open innovation, cloud computing network architecture, and streaming data analysis. He is the Founder of OMNILab, which is an open innovation lab focusing on data analysis. He has published over 100 technical papers in leading conferences and journals and is the owner of over 10 patents. In 2014, OMNILab won the champion of CCF national big data challenge among nearly 1000 teams, and was the champion of the Shanghai open data innovation and creation competition. He has served over 10 technical committees. He enthuses public service and science popularization, actively promotes crowd engaged innovation, and interdisciplinary collaboration.
\end{IEEEbiography}
\fi
\end{document}